\documentclass[Afour,sageh,times]{sagej}

\usepackage{moreverb,url}

\usepackage{graphics} 
\usepackage{subfigure}
\usepackage{epsfig} 
\usepackage{amsmath} 
\usepackage[T1]{fontenc}
\usepackage{bm}
\usepackage{cite}
\usepackage{booktabs}
\usepackage{multirow}
\usepackage{diagbox}
\usepackage{threeparttable}
\usepackage{colortbl}
\usepackage{algorithm}
\usepackage{algpseudocode}
\usepackage{stfloats}
\usepackage[colorlinks,bookmarksopen,bookmarksnumbered,citecolor=red,urlcolor=red]{hyperref}
\hypersetup{
    colorlinks=false, 
    }
\usepackage{xcolor}

\newcommand \transpose {\mathsf{T}}

\algrenewcommand{\algorithmiccomment}[1]{$//$ #1}
\definecolor{myblue}{RGB}{0, 0, 255}

\newcommand\BibTeX{{\rmfamily B\kern-.05em \textsc{i\kern-.025em b}\kern-.08em
T\kern-.1667em\lower.7ex\hbox{E}\kern-.125emX}}

\def\volumeyear{2023}

\begin{document}

\runninghead{Yu et al.}

\title{Generalizable whole-body global manipulation of deformable linear objects by dual-arm robot in 3-D constrained environments}

\author{Mingrui Yu\affilnum{1}, Kangchen Lv\affilnum{1}, Changhao Wang\affilnum{2}, 
Yongpeng Jiang\affilnum{1},  Masayoshi Tomizuka\affilnum{2} and Xiang Li\affilnum{1}}

\affiliation{\affilnum{1}Department of Automation, Tsinghua University, Beijing, China\\
\affilnum{2}Department of Mechanical Engineering, University of California, Berkeley, CA, USA}

\corrauth{Xiang Li, Department of Automation, Tsinghua University, Beijing, China}

\email{xiangli@tsinghua.edu.cn}

\begin{abstract}
Constrained environments, compared with open spaces without other objects, are more common in practical applications of manipulating deformable linear objects (DLOs) by robots, where movements of both DLOs and robot manipulators should be constrained and unintended collision should be avoided. Such a task is high-dimensional and highly constrained owing to the highly deformable DLOs, dual-arm robots with high degrees of freedom, and 3-D complex environments, which render global planning extremely challenging. Furthermore, accurate DLO models needed by planning are often unavailable owing to their strong nonlinearity and diversity, resulting in unreliable planned paths.
This article focuses on the global moving and shaping of DLOs in constrained environments by dual-arm robots. The main objectives are 1) to efficiently and accurately accomplish this task, and 2) to achieve generalizable and robust manipulation of various DLOs.
To this end, we propose a complementary framework with whole-body planning and control using appropriate DLO model representations. First, a global planner is proposed to efficiently find feasible solutions based on a simplified DLO energy model, which considers the full system states and all constraints to plan more reliable paths.
Then, a closed-loop manipulation scheme is proposed to compensate for the modeling errors and enhance the robustness and accuracy, which incorporates a constrained model predictive controller to track the planned path as guidance while real-time adjusting the robot motion based on an adaptive DLO motion model. 
This framework systematically considers multiple constraints for this problem, including stable deformation, overstretch prevention, closed-chain movements, and collision avoidance. The key novelty is that it can efficiently solve the high-dimensional problem subject to all those constraints and generalize to various DLOs without elaborate model identifications.
Experiments demonstrate that our framework can accomplish considerably more complicated tasks than existing works. It achieves a 100\% planning success rate among thousands of trials with an average time cost of less than 15 second, and a 100\% manipulation success rate among 135 real-world tests on five different DLOs.
\end{abstract}


\maketitle

\section{Introduction}

Deformable linear objects (DLOs), such as cables, wires, ropes, and rods, are prevalent in various everyday scenarios \citep{jose2018robotic}.
The inherent deformable nature of DLOs presents many new challenges when applying classical manipulation methods primarily developed for rigid objects to DLO manipulation \citep{zhu2022challenges}.
Many research works have been devoted to robotic manipulation of DLOs, but most of them are designed for unobstructed environments without other objects. Such settings are impractical for real-world applications, such as
cable assembly in industries or suturing in medical surgeries, where the movements of both DLOs and robots are constrained by the workspace and obstacles.

\begin{figure*} [tb]
  \centering 
    \includegraphics[width=\textwidth]{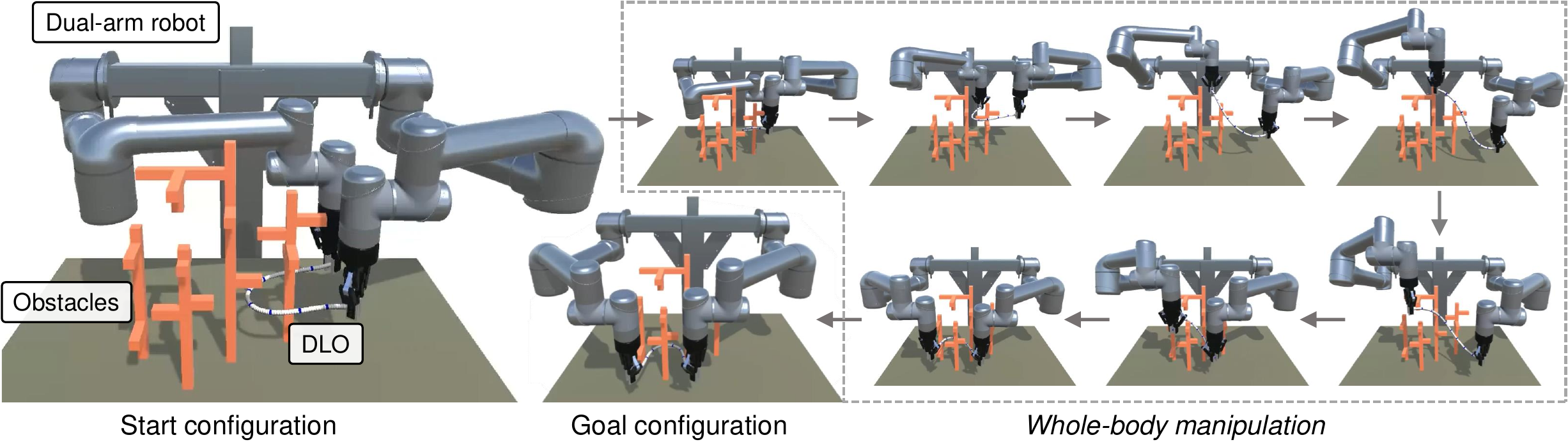} 
  \caption{Illustration of the task: whole-body global manipulation of DLOs by a dual-arm robot in 3-D constrained environments. Given the environment and goal configuration, the proposed approach achieves collision-free moving and shaping of the DLO from start to goal configuration, during which the whole body of the DLO and robot is considered.}
  \label{fig:fig1}
\end{figure*}

This article focuses on DLO manipulation in constrained environments.
Figure \ref{fig:fig1} illustrates an example of the manipulation task considered in this article, which constitutes a general problem in DLO manipulation, i.e., the task of a dual-arm robot manipulating a DLO from a start configuration to a desired (goal) configuration in a complex constrained environment with non-convex obstacles.
This task involves both moving and shaping of the DLO by the robot, necessitating both accurate final manipulation results and collision-free moving paths for the DLO and robot body.

This task is high-dimensional and highly constrained owing to the deformable nature of DLOs, high degrees of freedom (DoFs) of dual-arm robots, 3-D complex environments, under-actuated nature of the system, and requirement of long-distance movements. Consequently, global planning becomes indispensable but also extremely challenging. 
Some previous works have tried this task by offline planning and open-loop executions.
The most critical issue is that it is difficult to obtain accurate DLO models needed by planning in practical applications, given the significant diversity and nonlinearity of DLOs. Thus, directly executing planned paths may fail owing to the inevitable DLO modeling errors. However, most works assumed the acquisition of sufficiently accurate DLO models either through analytical modeling \citep{wakamatsu2004static,moll2006path,bretl2014quasi,roussel2020motion,sintov2020motion} or learning from pre-collected data \citep{mitrano2021learning}, taking no account of adaptability to new DLOs. 
Furthermore, previous works usually simplified the challenging planning problem or only partially addressed it, for example, by regarding the end-effectors as floating grippers without considering arm bodies \citep{wakamatsu2004static,moll2006path,bretl2014quasi,roussel2020motion}, relying on time-consuming physical engines \citep{roussel2020motion} or pre-built roadmaps \citep{sintov2020motion}, or over-simplifying the representations of DLOs \citep{mcconachie2020manipulating}.

In contrast, humans can easily pick up an arbitrary DLO and move it to the desired configurations without exhaustive training. Think about how a human achieve this task: first, the human imagines a collision-free path for the DLO body from the start to the goal configuration; then, the human manipulates the DLO approximately along this path while monitoring the DLO state in real time and adjusting the hand movements accordingly.

Inspired by this human approach, this article proposes a novel framework for global whole-body collision-free manipulation of DLOs in constrained environments. 
We aim to address two key questions: 
\begin{enumerate}
    \item How to both efficiently and accurately accomplish this high-dimensional task subject to multiple constraints, such as the stable deformation, overstretch prevention, closed-chain movements, and collision avoidance?
    \item How to enhance the robustness and generalizability of the proposed approach to various real-world DLOs for which accurate models are difficult to obtain?
\end{enumerate}
Our solution is a complementary framework that integrates whole-body global planning and local control: the former efficiently finds feasible (but imperfect) solutions and the latter improves the robustness and accuracy of manipulations.
Specifically, the global planner relies on a moderately simplified DLO energy model and efficiently computes a collision-free global path. We use a rapidly-exploring random tree (RRT) framework for this high-dimensional planning problem, in which we use projection and rejection methods to ensure all constraints are satisfied, including the stable DLO configuration constraint, closed-chain constraint, and collision-free constraint. 
Then, the local controller tracks the planned path as guidance while utilizing real-time feedback to compensate for planning errors. We formulate it as a model predictive controller (MPC) incorporating necessary constraints, such as collision avoidance and overstretch prevention, and we employ our previously proposed adaptive DLO Jacobian model \citep{yu2023global} for locally predicting DLO states. 
We additionally use a replanning strategy to deal with extreme situations in which the local controller fails.
In this framework, the robustness and generalizability on various DLOs are improved by the closed-loop manipulation scheme, coarse identification of the DLO energy model for planning, and online updating of the DLO Jacobian model for control.

The key contributions are highlighted as follows: 
\begin{enumerate}
    \item We establish an efficient path planning algorithm for global collision-free manipulation of DLOs by dual-arm robots, which considers the full state space of both the DLO and arms and guarantees the satisfaction of all necessary constraints.
    \item We implement an MPC for smoothly tracking DLO and robot paths, which includes hard constraints for local obstacle avoidance and overstretch prevention in general 3-D  environments.
    \item We propose a complementary framework combining global planning and local control, in which the planner relies on a simplified DLO model to efficiently find feasible solutions, and the controller uses real-time feedback to closed-loop compensate for planning errors during tracking. 
    It achieves robust, accurate, and collision-free manipulation of various DLOs in complex constrained environments.
\end{enumerate}

We carry out exhaustive simulations and real-world experiments to demonstrate that our framework can effectively address the open challenges in dual-arm manipulation of DLOs in constrained environments, such as those pertaining to high dimensionality, multiple constraints, long-distance movements, and generalization on various DLOs. 
To the best of our knowledge, this study is the first successful attempt at achieving whole-body collision-free manipulation of various types of DLOs in real-world 3-D constrained environments.
The proposed approach achieves a 100\% planning success rate among thousands of trials with an average time cost of less than 15 second (such as the simulated task in Fig. \ref{fig:fig1}), and a 100\% manipulation success rate among 135 real-world 3-D tests on five DLOs of different properties with an average execution time of less than 1 minute. 
The supplementary video and source code are available on the project website\footnote{Project website: \url{https://mingrui-yu.github.io/DLO_planning_2}}.

This work is an extension of our previous work \citep{yu2023acoarse}. The improvements include: 
1) employing a new DLO energy model, the discrete elastic rod model, to account for twist energy and gravity effects;
2) further optimizing the planning algorithm to improve the success rate, efficiency, and path quality; 
3) extending the controller to a long-horizon MPC with hard constraints for obstacle avoidance;
and 4) conducting more comprehensive simulation studies and 3-D real-world experiments.

\section{Related work}

\subsection{DLO manipulation tasks}

Many different tasks involving the robotic manipulation of DLOs have been independently explored, such as sorting USB cables \citep{gao2022ahierarchical,gao2023development}, soldering flexible PCBs \citep{li2018vision}, assembling belt drive units \citep{jin2021trajectory}, packing into boxes \citep{ma2022action}, tangling \citep{wakamatsu2006knotting,saha2007manipulation}, untangling \citep{grannen2020untangling,viswanath2021disentangling,huang2023untangling}, picking wire harnesses \citep{zhang2023learning}, cable following \citep{she2021cable,yu2024hand}, cable insertion \citep{weifu2015anonline,de2019integration}, and medical suturing \citep{cao2020sewing}.
Notably, those approaches were designed for specific task requirements, and aspects irrelevant to the tasks, such as obstacles or deformations, were simplified. 
In contrast, this work focuses on a fundamental problem in DLO manipulation: dual-arm manipulating a DLO from a start configuration to a goal configuration, which is an essential procedure in many high-level tasks in real-world applications. We call this task \textit{general DLO shaping}, which involves global moving and shaping of DLOs.

We classify the \textit{general DLO shaping} tasks into two categories. In the first category, the DLO is placed on a 2-D plane, and the robot can access any point along the DLO. The DLO shapes are preserved by frictions with the table surface when the DLO is not grasped by the robot. Consequently, the robots can move the DLO to the desired configuration through a series of pick-and-place actions. 
Various learning-based approaches have been proposed to manipulate soft ropes to arbitrary shapes on tables, such as learning from human demonstrations \citep{nair2017combining,tang2018aframework}, applying reinforcement learning \citep{lin2020softgym}, and learning forward predictive models and applying MPCs \citep{yan_self-supervised_2020, lee2022sample}. 
In addition, some works have studied cable routing tasks, in which the DLO shapes are preserved by external fixtures \citep{jin2022robotic,waltersson2022planning,keipour2022efficient}. 

This work focuses on the second category of the tasks, in which the robot grasps only the ends of the DLO but aims to control the entire DLO shape in 3-D space. The deformation of the DLO under external forces is mainly elastic, i.e., the DLO returns to its natural configuration after being released by the robot. 
In such tasks, the manipulation of DLOs is under-actuated and difficult to model, making planning and control challenging.
In the following sections, we introduce the existing approaches for such tasks, including modeling, control, and planning.
In addition, we outline the popular methods about planning and control for robot arms (Appendix B), as our task involves robot arm bodies.

\subsection{DLO modeling}

Modeling of DLOs is a prerequisite for model-based planning and control, which can be classified into analytical and data-driven modeling. The mass-spring model is the simplest analytical model for deformable objects, but it exhibits low accuracy \citep{gibson1997survey,zhong2023regressor}. The finite-element method is a general and accurate method for modeling deformable objects with physical fidelity and interpretability, but the computation is usually too expensive for robotic applications \citep{reddy2019introduction, yin2021modeling}. In addition, many works have modeled and simulated DLOs using rod theories \citep{rubin2002cosserat, wakamatsu2004static, bretl2014quasi}. \citet{bergou2008discrete} derived a discrete form of the Kirchhoff rod theory and 
proposed a model called discrete elastic rod (DER) model, which has been widely applied in computer graphics, physical analysis, and robotics \citep{bergou2010discrete, jawed2014coiling, goldberg2019planar, lv2022dynamic}.
However, the problem of applying analytical models to DLO manipulation is that the accuracy of these models are highly dependent on the accuracy of DLO parameters, e.g., Young modulus and shear modulus, which are difficult to acquire in practical applications. 

Instead, some works have leveraged neural-network models trained on pre-collected data \citep{yan_self-supervised_2020,yang2021learning, mitrano2021learning, wang2022offline}, which predict the next DLO state given the current state and robot motion. However, these methods require to collect large amounts of data on the manipulated DLO, and may not be effective when applied to a different untrained DLO as they take no account of the adaptiveness to various DLOs.

\subsection{DLO shape control}

DLO shaping has been widely studied from the perspective of control. These studies have designed local controllers which use the real-time error between the current and desired configuration as feedback and locally minimize the error. Most of these controllers are based on local Jacobian models, which establish linear relationships between the DLO motion and robot motion in local regions. The Jacobian matrices are approximately estimated using online data \citep{navarro2013modelfree, navarro2016automatic, jin2019robust, lagneau_automatic_2020, zhu2021vision,yang2023model} or analytical models \citep{berenson2013manipulation, moha2022asrigid, aghajanzadeh2022anoffline}. Some approaches first train data-driven nonlinear forward predictive models offline and then use MPCs for online control \citep{yang2021learning, wang2022offline}. In our previous work \citep{yu2022shape, yu2023global}, we proposed an efficient and adaptive approach to learning a DLO Jacobian model and achieved stable and smooth large deformation control.
However, these control-only methods are typically designed for open spaces without any other objects.

Some studies have considered obstacles in control tasks, addressing issues such as avoiding collisions of grippers \citep{berenson2013manipulation,ruan2018accounting} or utilizing external contact for DLO shaping \citep{huang2023learning}. However, these approaches treat the robot end-effectors as floating grippers, ignoring the constraints imposed by the arm bodies. More importantly, they can only deal with local tasks without global long-distance movements and large deformations.

\subsection{DLO path planning}

Several works have designed planning strategies for specific tasks, such as cable routing \citep{zhu2020robotic} or packing \citep{ma2022action}. In these tasks, the DLO planning problems were simplified according to the task characteristics. For example, the manipulation of a long belt was simplified to the motion planning for the belt tail in \citet{qin2023dual}. In contrast, this work focuses on the global moving and shaping of DLOs in general constrained environments.

Sampling-based planning methods have been proven to be suitable for high dimensional problems.
Generally, the dimension of the stable DLO configuration space is lower than that of the raw configuration space, rendering the generation of stable DLO configurations non-trivial.
\citet{bretl2014quasi} formulated the DLO static equilibrium as an optimal control solution and theoretically derived that the equilibrium configuration space of a one-end-fixed Kirchhoff elastic rod is a six-dimensional manifold. 
Quasi-static DLO planning could be achieved by sampling and searching on this six-dimensional manifold.
One problem is that this model is not easy to incorporate the gravity effect, and \citet{mishani2022realtime} showed that it may only be appropriate for DLOs of very high stiffness, such as nitinol rods.
\citet{roussel2015manipulation,roussel2020motion} used this model for global planning and supplemented it with a physical simulator for local planning to accomplish complex tasks at the cost of significant computation time.
\citet{sintov2020motion} investigated the motion planning of both the DLO and dual arms, for which they pre-computed a roadmap of stable DLO configurations to reduce the online time cost but could not ensure efficient adaptability to new DLOs. We provide more discussion for this series of work in the \textit{Results} section.

An alternative way is to determine DLO stable configurations by locally minimizing the deformation energy using general optimization approaches; then, the path planning is achieved by optimization \citep{wakamatsu2004static} or probabilistic roadmap methods (PRMs) \citep{moll2006path}. In this work, we incorporate a similar approach into a single-query RRT framework, which is used not only for random sampling but also for constraining extended nodes. In addition, our planner takes into account the full configurations of both the DLO and robot arm bodies, as well as their corresponding constraints.

Recently, some works have also tried learning non-physical DLO models from data using neural networks for path planning \citep{zhou2021lasesom, mitrano2021learning}. However, these methods require a large amount of training data, and their generalizability to untrained DLOs and scenarios cannot be guaranteed.

Furthermore, all of the aforementioned works attempted to accomplish the tasks solely through offline planning, without considering the effects of potential differences between the modeled DLO and the manipulated DLO.

\subsection{Combination of planning and control}

\citet{mcconachie2020manipulating} proposed another framework for interleaving planning and control for manipulating deformable objects. 
Their controller first attempts to perform tasks directly. If their deadlock predictor predicts that the controller will get stuck, their planner will be invoked to move the object to a new region. 
Note that their aim and combination approach are different from ours. First, their planner uses a more simplified model named ``virtual elastic band'' focusing on only the overstretching of deformable objects by grippers or obstacles. In this way, the DLO may be over-compressed or hooked by obstacles, so they trained a network to predict the likely hooked states in their later work \citep{mcconachie2020learning}. In contrast, our planner plans full DLO configurations, avoids collisions, and reduces unnecessary deformations.
Second, their planned gripper path is executed in an open-loop manner without local adjustment based on real-time feedback, as their controller is activated when the deformable object is close to the final desired configuration. In contrast, our approach employs feedback control during the whole manipulation process to both robustly track the planned path and accurately reach the final desired configuration.
Third, collisions between the DLO and obstacles are allowed in their approach, whereas our method ensures collision-free planning that reduces unnecessary, unexpected, and uncontrollable deformations.

\section{Problem statement}

The article considers the task where a dual-arm robot rigidly grasps the two ends of a DLO and quasi-statically manipulates the DLO to the desired configuration in 3-D constrained environments, as illustrated in Fig. \ref{fig:fig1}. Collisions between DLO, robot arms, and obstacles are not allowed during manipulations. The properties of the DLO are not known in advance. No assumptions are made about the convexity of the obstacles. 

\textit{Assumptions}: 
The following assumptions are made:
\begin{enumerate}
    \item The motion of the DLO and robot is slow, allowing the manipulation process to be considered as quasi-static, during which the inertial effects are ignored.
    \item The deformation of the DLO during manipulation is elastic and determined by its potential energy (elastic deformation energy and gravity energy).
    \item The ends of the DLO are rigidly grasped by the robot end-effectors.
    \item The DLO states can be estimated in real time using RGB-D cameras.
    \item The joint velocities of the robot arms can be kinematically controlled.
\end{enumerate}

\textit{Notations}:
We introduce some frequently used notations:
\begin{itemize}
    \item {\spaceskip=0.14em\relax $[\bm a; \bm b]$: vertical concatenation of column vectors $\bm a$ and $\bm b$.}
    \item $\bm a \cdot \bm b$: dot product of vectors $\bm a$ and $\bm b$.
    \item $\bm a \times \bm b$: cross product of vectors $\bm a$ and $\bm b$.
    \item $\|\bm a\|_2$: Euclidean norm of vector $\bm a$.
\end{itemize}

The environment of obstacles is denoted as $\mathcal{O}$.
The joint position vector of the left and right arm is represented as $\bm q_l \in \Re^{n_l}$ and $\bm q_r \in \Re^{n_r}$, where $n_l$ and $n_r$ is the DoFs of the left and right arm, respectively. For convenience, we define $\bm q = [\bm q_l; \bm q_r]$.
The pose vector of the two robot end-effectors is denoted as $\bm r$.
The representation of the DLO is described in the next section. In addition, several distance metrics are defined in Appendix C.
The control input $\bm u$ to the system is the joint velocities of the dual-arm robot:
\begin{equation}
    \dot{\bm q} = [\dot{\bm q}_l; \dot{\bm q}_r] = \bm u .
\end{equation}

\section{Representing and modeling DLOs}

We use a common discrete formulation \citep{bergou2008discrete,bergou2010discrete,jawed2014coiling} to represent the DLO configuration. As shown in Fig. \ref{fig:der_vertices_edges}, the DLO is discretized into $(m+2)$ vertices $\bm x_0, \cdots, \bm x_{m+1}$ connected by $(m+1)$ straight edges $\bm e^0, \cdots, \bm e^{m}$, where $\bm e^{k} = \bm x_{k+1} - \bm x_{k}$. As shown in Fig. \ref{fig:der_frames}, each edge $\bm e^{k}$ is assigned a \textit{material frame} $\bm M^k = \{ \bm t^k, \bm m_1^k, \bm m_2^k \}$ to represent its actual orientation, where $\bm t^k = \bm e^k / \| \bm e^k \|_2$ is the normalized tangent vector of the edge.
We define $\bm \Gamma$ as the DLO configuration consisting of all vertices $\bm x_0, \cdots, \bm x_{m+1}$ and material frames $\bm M^0, \cdots, \bm M^m$.

In terms of the boundary conditions, we consider vertices $\bm x_1$ and $\bm x_m$ as the positions of the two ends, instead of $\bm x_0$ and $\bm x_{m+1}$. Additionally, frames $\bm M^0$ and $\bm M^{m}$ represent the orientations of the two ends. This is because when the poses of the two ends of an inextensible DLO are fixed, its $\bm x_0, \bm x_1, \bm M^0$ and $\bm x_m, \bm x_{m+1}, \bm M^{m}$ are all constrained, so $ \bm x_1$ and $\bm x_m$ cannot be regarded as internal non-grasped free vertices. Our formulation considers $\bm x_0$ and $\bm x_{m+1}$ to be virtual variables only for the convenience of calculation, which are actually redundant, as they can be fully determined by $\bm x_1, \bm M^0$ and $ \bm x_m, \bm M^{m}$.

Regarding the perception of DLO configurations in the real world, the positions of the centerline can be visually perceived, and the orientations of the ends can be measured by the robot forward kinematics. However, the internal orientations cannot be directly perceived by sensors.
Thus, we define a perceptible DLO configuration $\check{\bm \Gamma} = \{\bm x, \bm M^0, \bm M^{m} \}$, where $\bm x = [\bm x_1; \cdots; \bm x_m]$. We also call $(\bm x_1, \cdots, \bm x_m)$ the \textit{feature points} along the DLO, consistent with our previous work \citep{yu2023global}.

\subsection{Discrete elastic rod (DER) model}

Here we briefly introduce the DLO energy model used in the planning, which is called the discrete elastic rod (DER) model \citep{bergou2008discrete}.

The classical Kirchhoff theory of elastic rods assigns an elastic energy to a continuous DLO configuration, which is a linear combination of bend and twist energies \citep{dill1992kirchhoff}. Stretching energy can also be considered, but since most DLOs used in daily life are inextensible, the stretch energy is not included and replaced by inextensible constraints. 

\begin{figure} [tb]
  \centering 
  \subfigure[]{ 
    \label{fig:der_vertices_edges}
    \includegraphics[width=0.40\textwidth]{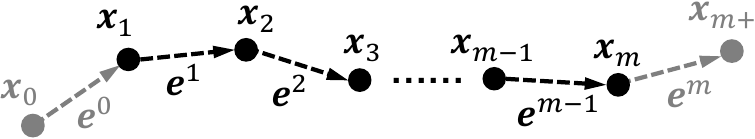} 
  } 
  \subfigure[]{ 
    \label{fig:der_frames}
    \includegraphics[width=0.42\textwidth]{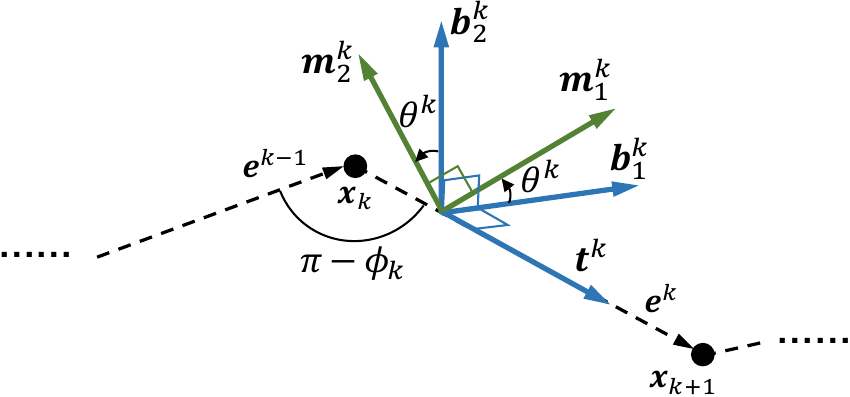} 
  }
  \caption{Discrete representation of DLO configuration. (a) The vertices and edges for discretizing the DLO; (b) The Bishop frame and material frame attached to each edge.}
  \label{fig:der_structure}
\end{figure}

The DER model derives a discrete form of the Kirchhoff rod model.
To measure how much the DLO is twisted, the DER model incorporates the \textit{Bishop frame}, an adapted frame with zero twist for a given centerline. As shown in Fig. \ref{fig:der_frames}, a \textit{Bishop frame} $\bm B^k = \{ \bm t^k, \bm b_1^k, \bm b_2^k \}$ is assigned to each edge $\bm e^k$, and it can be defined by the \textit{parallel transport} throughout the centerline; that is
\begin{equation} \label{eq:parallel_transport_1}
    \bm b_1^k = R_k (\bm b_1^{k-1}), 
    \quad
    \bm b_2^k = \bm t^k \times \bm b_1^k ,
\end{equation}
where $R_k$ is a rotation operator that satisfies
\begin{equation} \label{eq:parallel_transport_2}
    R_k (\bm t^{k-1}) = \bm t^k, \quad  R_k (\bm t^{k-1} \times \bm t^{k}) = \bm t^{k-1} \times \bm t^{k} ,
\end{equation}
and $R_k$ is defined as the identity if $\bm t^{k-1} = \bm t^{k}$. We assign $\bm b_1^0$ to be equal to $\bm m_1^0$ to obtain a unique Bishop frame.

We consider the potential energy of $\bm \Gamma$ as the sum of its elastic energy and gravity energy, defined as
\begin{equation}
    E(\bm \Gamma) = E_{\rm bend}(\bm \Gamma) + E_{\rm twist}(\bm \Gamma) + E_{\rm gravity}(\bm \Gamma) .
\end{equation}
The following energy equations are for naturally straight and isotropic DLOs.

\textbf{Bending energy}: The bending energy is determined by the curvature of the centerline. The curvature binormal at a vertex $\bm x_k$ is defined as 
\begin{equation} \label{eq:bend_energy}
    (\kappa \bm b)_k = \frac{2 \bm e^{k-1} \times \bm e^{k}} {\| \bm e^{k-1} \|_2 \| \bm e^{k} \|_2 + \bm e^{k-1} \cdot \bm e^{k} } ,
\end{equation}
whose magnitude is $2 \tan(\phi_k / 2)$ where $\phi_k$ is the angle between $\bm e^{k-1}$ and $\bm e^{k}$. Define $l_k = \|\bm e^{k-1}\|_2 + \|\bm e^{k}\|_2$. Then, the bending energy can be defined as
\begin{equation}
   E_{\rm bend}(\bm \Gamma) 
   = \frac{1}{2} \sum_{k = 1}^{m} \lambda_{\rm b} \left( \frac{(\kappa \bm b)_k}{l_k/2} \right)^2 \frac{l_k}{2} 
   =  \lambda_{\rm b} \sum_{k = 1}^{m} \frac{((\kappa \bm b)_k)^2}{l_k} ,
\end{equation}
where $\lambda_{\rm b}$ is the bending stiffness.

\textbf{Twisting energy}: The twisting energy is defined as 
\begin{equation} \label{eq:twist_energy}
   E_{\rm twist}(\bm \Gamma) 
   = \sum_{k = 1}^{m} \lambda_{\rm t}  \frac{(\theta^k - \theta^{k-1})^2}{l_k} ,
\end{equation}
where $\lambda_{\rm t}$ is the twisting stiffness, and $\theta^k$ is the angle between the Bishop frame $\bm b_1^k$ and material frame $\bm m_1^k$, which is illustrated in Fig. \ref{fig:der_frames} .

\textbf{Gravity energy}: The gravity energy is defined as 
\begin{equation} \label{eq:gravity_energy}
   E_{\rm gravity}(\bm \Gamma) 
   = \sum_{k=1}^{m} \rho g h_k \frac{l_k}{2} = \rho \sum_{k=1}^{m} \frac{g h_k l_k}{2} ,
\end{equation}
where $\rho$ is the linear density (unit: $\rm kg/m$) of the DLO, $h_k$ is the height of $\bm x_k$, and $g$ is the gravitational acceleration.

Readers can refer to \citet{bergou2008discrete} for the gradient of each energy term and more detailed explanation. 

\subsection{DLO Jacobian model}

We use a DLO motion model proposed in our previous work \citep{yu2023global} for control, which is called the DLO Jacobian model. 
\citet{bergou2008discrete} derived that under the quasi-static assumption, the internal material frames $\bm M^1, \cdots, \bm M^{m-1}$ are uniquely determined by the centerline and boundary material frames, so the potential energy $E$ of a clamped DLO can be fully determined by the configurations of the DLO centerline $\bm x$ and robot end-effectors $\bm r$.
Then, according to the proof in \citep{yu2023global}, the motion of the feature points of an elastic DLO is related to that of the robot end-effectors as
\begin{equation} \label{eq:DLO_jacobian_model}
    \underbrace{
    \left[ \begin{array}{c}
         \dot{\bm x}_1 \\
         \vdots \\
         \dot{\bm x}_m
    \end{array} \right]
    }_{\dot{\bm x}}
    = 
    \underbrace{
    \left[ \begin{array}{c}
         \bm J_1(\bm x, \bm r)  \\
         \vdots \\
         \bm J_m(\bm x, \bm r)
    \end{array} \right]
    }_{\bm J(\bm x, \bm r)}
    \underbrace{
    \left[ \begin{array}{c}
         \bm v_l \\
         \bm \omega_l \\
         \bm v_r \\
         \bm \omega_r
    \end{array} \right]
    }_{\bm \nu} ,
\end{equation}
where $\bm v_l, \bm v_r \in \Re^3$ is the linear velocity of the left and right end, respectively; $\bm \omega_l, \bm \omega_r \in \Re^3$ is the angular velocity of the left and right end, respectively. In addition, $\bm J_k(\cdot)$ represents a mapping from the current configuration $\bm x, \bm r$ to the corresponding Jacobian matrix for the $k^{\rm th}$ feature point such that $\dot{\bm x}_k = \bm J_k(\bm x, \bm r) \bm \nu$. 

Note that this model is independent with the specific formula of the energy $E$. We use a data-driven approach to learn the Jacobian mapping through pre-training on diverse simulation data and further updating it during actual manipulation using online data.

\section{Overview of the complementary framework}

In this section, we provide an overview of the proposed complementary framework for manipulating DLOs through a combination of whole-body planning and control. 
Before explaining the rationality and strengths of this framework, we emphasis that the modeling error of DLOs is a crucial factor when designing approaches to manipulating different DLOs without fine model identification.

However, the presence of inevitable DLO modeling errors does not undermine the significance of DLO models. 
Our core concept is to use a moderately coarse DLO model to plan a path that closely approximates real-world conditions without significantly increasing the time cost. 
Subsequently, we use closed-loop control with an adaptive DLO motion model to compensate for the residual modeling errors. 
Under such a complementary framework, we can achieve both efficient global planning and accurate execution of the task.

\subsection{Global planning}

A higher quality of the planned path corresponds to less likelihood of the controller getting stuck in a local optimum during tracking.
During the planning phase, we consider all constraints of configurations based on the assumed DLO model, including: 
\begin{enumerate}
    \item The configuration of the DLO is constrained to be valid and stable, which is denoted as $\bm C_{\rm stable}(\bm \Gamma) = \bm 0$. 
    \item At any waypoint, the DLO and dual arms form a closed chain to maintain rigid grasps. The closed-chain constraint is denoted as $\bm C_{\rm chain}(\bm \Gamma, \bm q) = \bm 0$. 
    \item The DLO and arms are constrained to not collide with themselves, each other, or environment. 
    The collision constraint is denoted as $\bm C_{\rm collision}(\bm \Gamma, \bm q, \mathcal{O}) = \bm 0$. 
\end{enumerate}
Notably, it is challenging to plan paths satisfying all these constraints while staying efficient. The existing works made attempts at it in simple environments with the help of pre-build roadmaps or neural networks trained by large amounts of motion data.
In contrast, we achieve efficient planning in complex environments without time-consuming offline preparation, by using appropriate projection methods to generate constrained configurations. 

We employ the DER model for planning rather than data-driven forward predictive models of DLOs, such as \citet{mitrano2021learning}, since the latter may suffer from accumulated errors and fail to satisfy the constraints over long paths.

To generalize to various types of DLOs, the planner needs to use a DER model with appropriate parameters that approximate the specific property of the manipulated DLO. Thus, we need to efficiently identify the DER model parameters in advance. 
However, identifying an anisotropic DLO model which assigns different parameters for each discrete element is too complicated and impractical.
Thus, we use a simplified model that assumes naturally straight and isotropic DLOs in planning. Such a model involves only three scalar parameters, and we design an efficient strategy to coarsely identify them through a simple trajectory.
This simplification also helps improve the planning efficiency.

\begin{figure} [tb]
  \centering 
    \includegraphics[width=\linewidth]{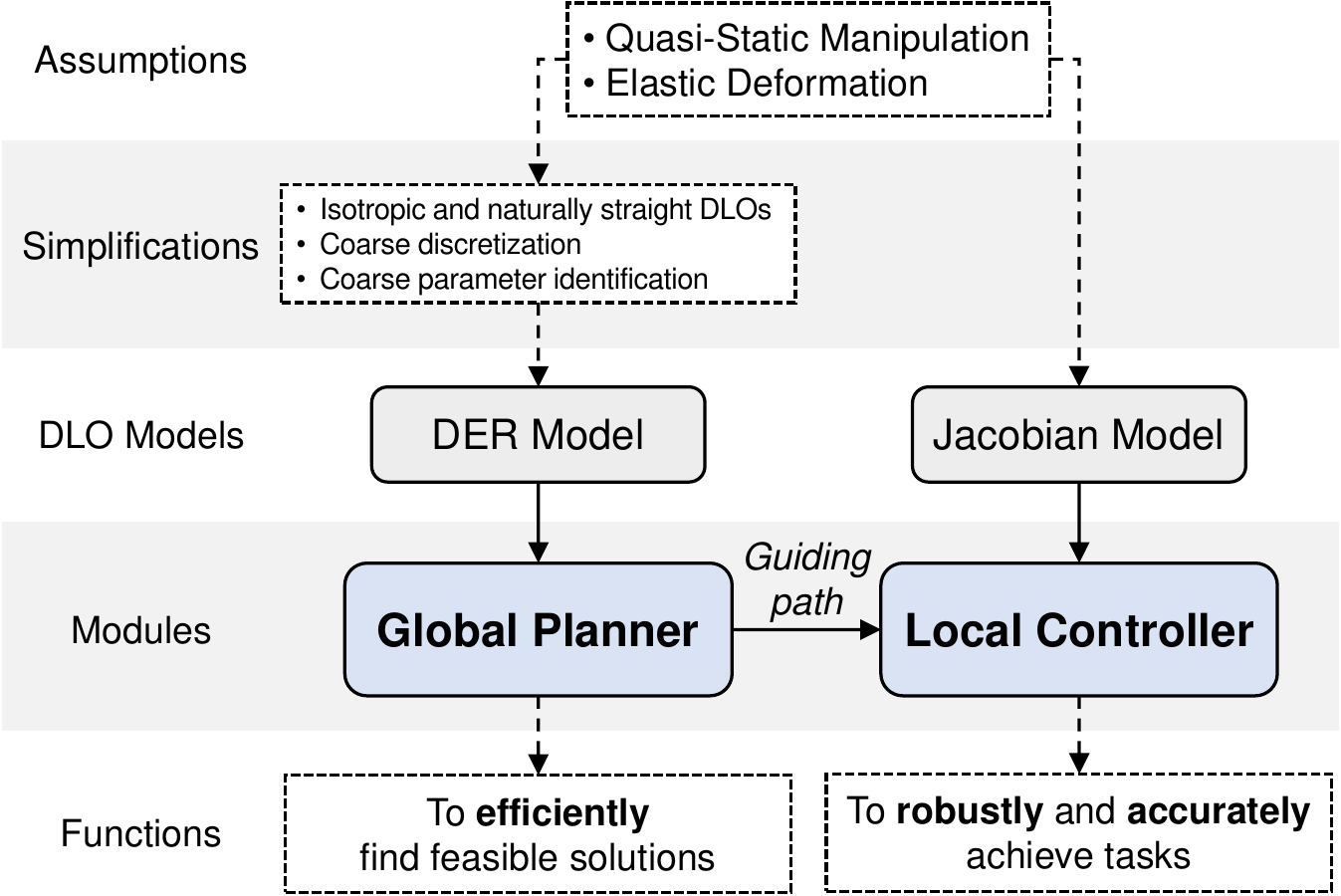} 
 \caption{Relationships between the global planner, local controller, and their corresponding DLO models in the proposed complementary framework.}
  \label{fig:models_and_modules}
\end{figure}

\subsection{Local control}

During the planning phase, the following simplifications are made in exchange for realizability and high efficiency:
\begin{enumerate}
    \item The energy formulas (\ref{eq:bend_energy}) to (\ref{eq:gravity_energy}) are specified for isotropic and naturally straight DLOs.
    \item The discretization of the DLO is coarse.
    \item The identification of the model parameters is coarse.
\end{enumerate}
Because of these simplifications, if the planned robot path is directly executed in an open-loop manner, the DLO may not move exactly as expected, potentially failing to reach the goal configuration owing to collisions or deviations.
Consequently, we use closed-loop control to compensate for the residual modeling errors and achieve robust and accurate manipulation. 

A DLO motion model is required for model-based control. We apply our previously proposed DLO Jacobian model (\ref{eq:DLO_jacobian_model}). Unlike the simplified analytical model used in planning, this data-driven model only assumes the elasticity of the DLO and quasi-static manipulation, and is constantly updated to adapt to the specific manipulated DLO using online data during actual manipulation.
As a result, the controller is more general and locally precise compared with the planner.
The complementary relationship between the planner and controller is illustrated in Fig. \ref{fig:models_and_modules}.

The tracking objective of the controller includes both the planned DLO and robot path, aiming to ensure that both the DLO and robot follow the planned path as close as possible to avoid becoming trapped in local optima. 
Given that modeling errors and real-time adjustments will cause the DLO and robot to not move exactly as expected, the controller should be able to locally avoid potential collisions.
Consequently, we formulate this problem as an MPC with hard collision constraints. Other constraints, such as the overstretch and robot DoFs constraints, can also be easily incorporated. 
The controller is the first to consider the full configurations of both the DLO and robot while enforcing collision avoidance through hard constraints.

\section{Whole-body global planning}

In our complementary framework, the global planner is first invoked to efficiently find a global collision-free path of the DLO and robot from the start to the goal configuration. 

We use a bi-directional RRT framework. Two search trees, originating from the start and goal configuration respectively, grow toward each other until they are connected. The bi-directional RRT can be used for this under-actuated problem because the moving path is assumed to be reversible according to (\ref{eq:DLO_jacobian_model}).
Each node $\mathcal{N}_i$ contains both the DLO configuration $\mathcal{N}_i.{\bm \Gamma}$ and dual arm configuration $\mathcal{N}_i.{\bm q}$. We introduce the proposed planning algorithm and its key steps such as sampling and steering in the following sections.
The used distance metrics are defined in Appendix C.

\subsection{Constraining DLO configurations}

In the RRT framework, it is necessary to guarantee that every node in the exploration trees contains a stable DLO configuration. The raw configuration space of the DLO is $4m + 1$ dimensional (3-D positions of the $m$ feature points and angles of the $m+1$ frames). However, only a subspace contains stable equilibrium configuration, which is theoretically a lower-dimensional manifold \citep{bretl2014quasi}. Consequently, randomly sampling stable configurations from the raw space with rejection strategies is impractical.

The analytical model presented by \citet{bretl2014quasi} introduces an approach to direct sampling equilibriums from a six-dimensional chart and mapping them to Cartesian-space configurations. 
Planning on this chart is feasible. However, mapping a Cartesian-space DLO configuration back to its six-dimensional parameter in the chart is not straight-forward\footnote{Such mapping could be possibly achieved by using Jacobian-based iterative methods when the DLO stiffnesses are exact and the gravity is ignored \citep{borum2014state}, at the cost of more computation.}, 
leading to inconvenience for planning and manipulation. In addition, the model ignores gravity, which can significantly affect the stable DLO configurations. Further discussion is provided in the \textit{Results} section.

Instead, we use projection methods to move a randomly sampled or steered configuration onto its neighboring constrained manifold. Under the quasi-static assumption, a stable configuration implies that it is at an equilibrium where the DLO configuration $\bm \Gamma$ locally minimizes the potential energy $E(\bm \Gamma)$ subject to boundary conditions (poses of the ends) \citep{bretl2014quasi,navarro2016automatic,yu2023global}.
Consequently, a random configuration $\bar{\bm \Gamma}$ can be projected onto the stable configuration manifold by a local minimization of the energy subject to the two end pose constraints and inextensible constraints,
with $\bar{\bm \Gamma}$ as the initial value.
Such an approach is inspired by \citet{wakamatsu2004static} and \citet{moll2006path} who conducted simulations in 2-D or open space without robot bodies. Different from them, we use a discrete DLO energy model and incorporate it into a single-query RRT planning framework to achieve efficient global planning in 3-D constrained space.

\begin{algorithm}[tb] 
\caption{:  $\bm \Gamma_{\rm stable} = \text{ProjectStableDLOConfig}(\bm \Gamma_{\rm init})$ } 
\label{algorithm:dlo_projection}
\begin{algorithmic}[1]
    \Statex \Comment{  note that $\bm \Gamma = \{ \bm x_{0}, \cdots, \bm x_{m+1}, \bm M^0, \cdots, \bm M^m \} $ } 

    \Statex \Comment{preparing the boundary conditions and constraints} 
    
    \State  $\bm{\bar{x}}_1 \leftarrow \bm{x}_{1, \rm{init}}$, \quad $\bm{\bar{x}}_m \leftarrow \bm{x}_{m, \rm{init}}$; 
    
    \State  $\bm{\bar{e}}^k \leftarrow \bm{e}^{k}_{\rm init}, \quad k = 1, \cdots, m-1 $; 

    \State  $\bm{x}_{0, \rm stable} \leftarrow \bm{x}_{0, \rm init}$, \quad $\bm{x}_{m+1, \rm stable} \leftarrow \bm{x}_{m+1, \rm init}$; 

    \State  $\bm{M}^0_{\rm stable} \leftarrow \bm{M}^0_{\rm init}$, \quad $\bm{M}^m_{\rm stable} \leftarrow \bm{M}^m_{\rm init}$; 

    \Statex \Comment{ calculating the centerline by optimization} 

    \State  Obtain $\bm x_{k, \rm stable} (k=1,\cdots,m)$ by solving the optimization problem (\ref{eq:dlo_projection}) with $\bm \Gamma_{\rm init}$ as the initial value; 

    \Statex  \Comment{calculating the corresponding material frames} 
    
    \State  $\bm B^0_{\rm stable} \leftarrow \bm{M}^0_{\rm stable}$, \quad $\theta^0_{\rm stable} \leftarrow 0$; 

    \State Obtain the Bishop frames $\bm B^k_{\rm stable} (k=1, \cdots, m)$ by the parallel transport (\ref{eq:parallel_transport_1}) and (\ref{eq:parallel_transport_2}) using $\bm x_{\rm stable}$ and $\bm B^0_{\rm stable}$; 

    \State Obtain $\theta^m_{\rm stable}$ using  $\bm{M}^m_{\rm stable}$ and $\bm B^m_{\rm stable}$; 

    \State {Obtain $\theta^k_{\rm stable} (k \!= \!1, \! \cdots \!, m \! - \! 1)$ by (\ref{eq:uniform_twist}) using $\theta^0_{\rm stable}$ and $\theta^m_{\rm stable}$;} 

    \State  {Obtain $\bm M^k_{\rm stable} (k \!= \!1, \! \cdots \!, m \! - \! 1)$ using $\bm B^k_{\rm stable}$ and $\theta^k_{\rm stable}$;} 

    \State \scalebox{0.94}{  $\bm \Gamma_{\rm stable} \leftarrow \{ \bm x_{0, \rm stable}, \! \cdots \!, \bm x_{m+1, \rm stable}, \bm M^{0}_{\rm stable}, \! \cdots \! ,  \bm M^{m}_{\rm stable} \}$; } 

    \State  return $\bm \Gamma_{\rm stable} $; 
\end{algorithmic}
\end{algorithm}

\begin{figure} [tb]
  \centering 
    \includegraphics[width=\linewidth]{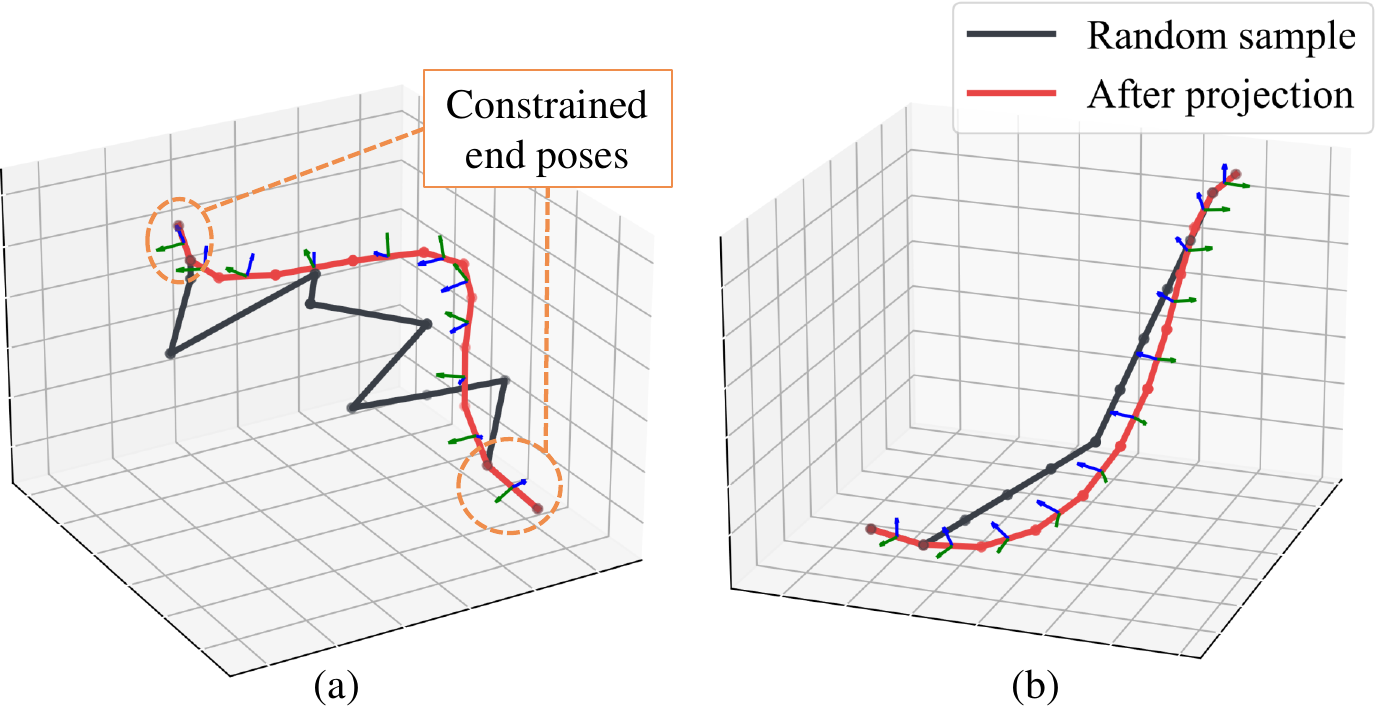} 
  \caption{Two examples of projecting randomly sampled invalid DLO configurations to stable configurations by (\ref{eq:dlo_projection}). The material frames $\bm m_1$ and $\bm m_2$ of the projected configurations are shown by the blue and green arrows, respectively.}
  \label{fig:derm_projection}
\end{figure}

Specifically, we employ the DER model. Note that only the twisting energy is related to  $\theta_k$. Thus, for a stable configuration at a local minimum of energy, we have 
\begin{equation} \label{eq:partial_twist_energy}
   \frac{\partial E_{\rm twist}(\bm \Gamma)}{\partial \theta_k} = 0 , \quad k = 1, \cdots, m-1 .
\end{equation}
Then, from (\ref{eq:twist_energy}) and (\ref{eq:partial_twist_energy}), it can be analytically derived that
\begin{equation} \label{eq:uniform_twist}
\frac{\theta^k - \theta^{k-1}}{l_k} 
= constant
= \frac{\theta^m - \theta^{0}}{\sum_{k=1}^{m}l_k}, \quad k = 1, \cdots, m
\end{equation}
at a stable configuration, which means that the DLO has uniform twist. Then, the twist energy can be formulated as 
\begin{equation}
   E_{\rm twist}(\bm \Gamma) 
   = \sum_{k = 1}^{m} \lambda_{\rm t}  \frac{(\theta^k - \theta^{k-1})^2}{l_k}
   = \lambda_{\rm t} \frac{(\theta^m - \theta^{0})^2}{\sum_{k=1}^{m}l_k} .
\end{equation}
Now, the potential energy is fully determined by the positions of the centerline and poses of the two ends. Therefore, we can omit the internal material frames $\bm M^1, \cdots, \bm M^{m-1}$ and only consider $\check{\bm \Gamma}$ instead of $\bm \Gamma$ in the minimization. Moreover, because the $\bm M^0$ and $\bm M^m$ are fixed boundary conditions, only the centerline $\bm x$ (i.e., positions of the features) are variables, which reduces the complexity of the minimization.
Note that $E_{\rm twist}$ is related to $\bm x$, since $\theta^m$ depends on the Bishop frame, and the Bishop frame is calculated by the parallel transport throughout $\bm x$ using (\ref{eq:parallel_transport_1}) and (\ref{eq:parallel_transport_2}).

Then, the local minimization problem for projecting a DLO configuration to a neighboring stable one using the DER model can be formulated as:
\begin{equation} \label{eq:dlo_projection}
\begin{aligned}
    \bm x_{\rm stable} = 
     \arg & \min_{\bm x} 
       \,   E_{\rm bend}({\bm \Gamma}) + E_{\rm twist}({\bm \Gamma}) + E_{\rm gravity}({\bm \Gamma})
    \\
    \text{s.t.} 
    \quad & \bm x_1 = \bar{\bm x}_1 
    \\
    \quad & \bm x_m = \bar{\bm x}_m
    \\
    \quad & \| \bm e^k \|_2 = \| \bar{\bm e}^k \|_2, \quad k = 1, \cdots, m-1 ,
\end{aligned}
\end{equation}
where $\bar{\bm x}_1$ and $\bar{\bm x}_m$ are the fixed positions of the grasped ends, and the lengths of all edges $\bm e_k = \bm x_{k+1} - \bm x_{k} \, (k=1,\cdots,m-1)$ are constrained to be fixed.
Then, we can apply a general nonlinear optimization solver to efficiently solve this problem. 
The resulting stable configuration $\bm \Gamma_{\rm stable}$ can be fully determined by $\bm x_{\rm stable}, \bm M^0$ and $\bm M^m$.
We denote such a process of projecting a random configuration $\bm \Gamma_{\rm init}$ to a neighbor stable configuration as $\bm \Gamma_{\rm stable} = \text{ProjectStableDLOConfig}(\bm \Gamma_{\rm init})$, which is detailedly described in Algorithm \ref{algorithm:dlo_projection}.
Fig. \ref{fig:derm_projection} visualizes two examples of such projections.

\subsection{Sampling random nodes}

In the RRT framework, a $\mathcal{N}_{\rm rand}$ is randomly sampled in each iteration to guide the exploration of the search tree. Our strategy for sampling involves the following steps. First, we sample a coarse DLO configuration using heuristic methods, such as sampling broken lines, like the random sample in Fig. \ref{fig:derm_projection}(b). 
Second, we randomly sample a robot configuration satisfying the closed-chain constraints by robot inverse kinematics (IK). If no valid IK solutions are found, we discard the DLO configuration and sample another. 
Third, the ProjectStableDLOConfig() is performed to project the sampled DLO configuration to a stable one. 
The sampling process is denoted as $\mathcal{N}_{\rm rand} = \text{RandomSampleInFullSpace}()$.

Since all nodes in the exploration trees are constrained by the steering function (described below) and sampled nodes $\mathcal{N}_{\rm rand}$ are only used as directional guidance, it is not necessary to ensure that $\mathcal{N}_{\rm rand}$ satisfy all constraints. 
In practice, we typically omit the relatively time-consuming step ProjectStableDLOConfig() in every random sampling. We also find that ignoring collisions during sampling can enhance the planning efficiency.

\subsection{Constrained steering and extending}

\begin{figure} [tb]
  \centering 
    \includegraphics[width=\linewidth]{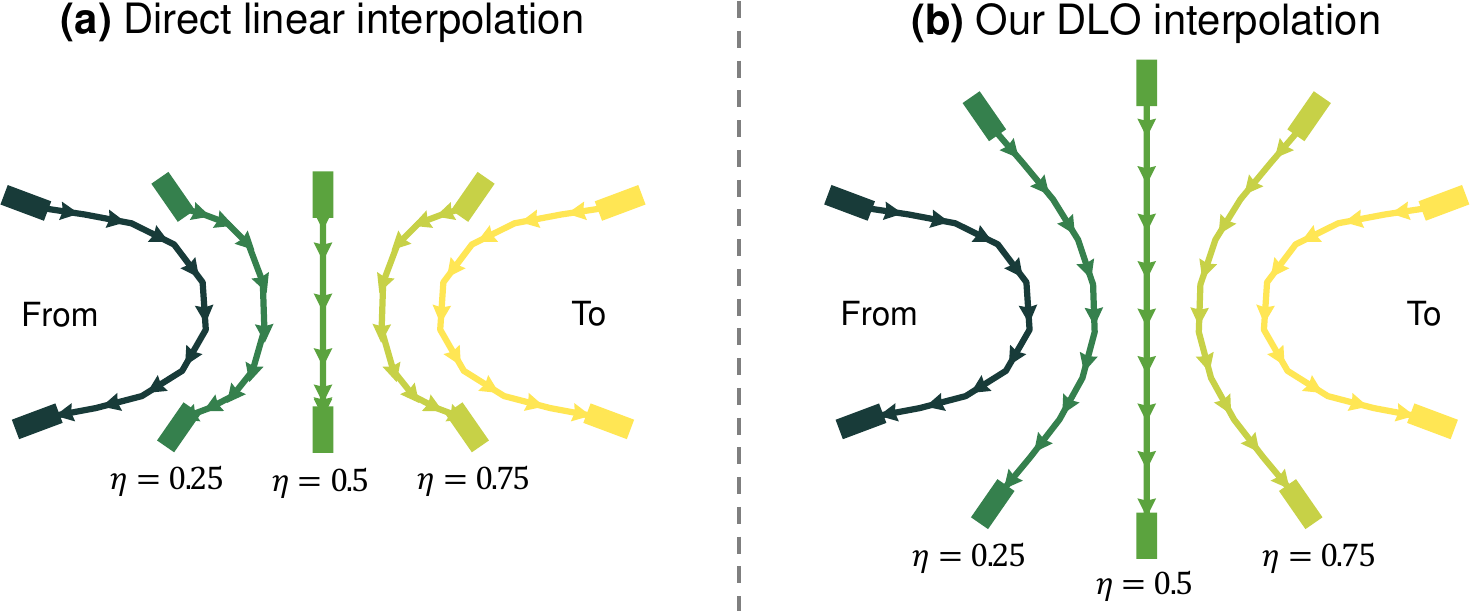} 
  \caption{Comparison between the direct linear interpolation (a) and our DLO interpolation method (b). The lines represent the DLO centerlines; the arrows indicate the directions from left to right ends; the rectangles indicate the poses of DLO ends;  
  and $\eta \in [0, 1]$ refers to the interpolation ratio.}
  \label{fig:dlo_interpolation}
\end{figure}

A key component of the RRT algorithm is the steering function
$\mathcal{N}_{\rm steer} = \text{Steer} \left( \mathcal{N}_{\rm from}, \mathcal{N}_{\rm to}, \bm \eta_{\rm step} \right)$, which generates a new node $\mathcal{N}_{\rm steer}$ by moving from $\mathcal{N}_{\rm from}$ towards $\mathcal{N}_{\rm to}$ with a small step size $\bm \eta_{\rm step}$.
The challenges of steering in this problem stem from the constraints and under-actuated nature of DLO manipulation.

In configuration-space planning, steering is usually achieved by interpolation, such as linear interpolation between two arm joint configurations. 
Thus, we first design an approach to interpolating between two DLO configurations, denoted as $\bm \Gamma_{\rm erp} = \text{DLOerp} \left( \bm \Gamma_{\rm from}, \bm \Gamma_{\rm to}, \eta_{\rm ratio} \right)$, where $\eta_{\rm ratio} \in [0,1]$ is the interpolation ratio. 
A straightforward way is to linearly interpolate the positions of the DLO vertices independently. However, it does not preserve the DLO length and will generate invalid over-compressed configurations, as shown in Fig. \ref{fig:dlo_interpolation}(a).
Instead, we first interpolate the centroid and material frames and then generate the centerline using the original edge lengths (Alogrithm \ref{algorithm:dlo_interpolation}).
First, we linearly interpolate (Lerp) the centroid of the centerline.
Next, we interpolate the material frames by the spherical linear interpolation (Slerp) between the corresponding quaternions. 
Then, using the new material frames and original edge lengths, we establish the new centerline regardless of overall translation.
Finally, we translate the new centerline to the new centroid and get the final interpolated $\bm \Gamma_{\rm erp}$.
As shown in Fig. \ref{fig:dlo_interpolation}(b), this approach can generate smooth interpolations closer to the stable configuration manifold, even if the deformation between $\bm \Gamma_{\rm from}$ and $\bm \Gamma_{\rm to}$ is large.

\begin{algorithm}[tb] 
\caption{: $\bm \Gamma_{\rm erp} = \text{DLOerp} \left( \bm \Gamma_{\rm from}, \bm \Gamma_{\rm to}, \eta_{\rm ratio} \right)$}
\label{algorithm:dlo_interpolation}
\begin{algorithmic}[1]
    \Statex \Comment{note that $\bm \Gamma = \{ \bm x_{0}, \cdots, \bm x_{m+1}, \bm M^0, \cdots, \bm M^m \} $}

    \Statex \Comment{interpolating the centroid of the centerline}
    
    \State $\bar{\bm x}_{\rm from} \leftarrow \frac{1}{m+2} \sum_{k=0}^{m+1} {\bm x}_{k, \rm from}$; 
    
    \State $\bar{\bm x}_{\rm to} \leftarrow \frac{1}{m+2} \sum_{k=0}^{m+1} {\bm x}_{k, \rm to}$;

    \State $\bar{\bm x}_{\rm erp} \leftarrow \text{Lerp}\left( \bar{\bm x}_{\rm from}, \bar{\bm x}_{\rm to}, \eta_{\rm ratio} \right)$;

    \Statex \Comment{interpolating the material frames}
    
    \For{ $k=0$ {\bf to} $m$ }
        \State ${\bm M}^k_{\rm erp} \leftarrow \text{Slerp} \left(\bm M^k_{\rm from}, \bm M^k_{\rm to}, \eta_{\rm ratio} \right)$; 
    \EndFor

    \Statex \Comment{establishing the new centerline}
    
    \State ${\bm x}_{0, \rm temp} \leftarrow \bm 0$;
    
    \For{ $k=0$ {\bf to} $m$ }
        \Statex \quad \Comment{note that $\bm M^k = \{ \bm t^k, \bm m_1^k, \bm m_2^k \}$}
        \State ${\bm x}_{(k+1), \rm temp} \leftarrow {\bm x}_{k, \rm temp} + \| \bm e^k \|_2 \cdot \bm t^k_{\rm erp}$;
    \EndFor
    
    \State $\bar{\bm x}_{\rm temp} \leftarrow \frac{1}{m+2} \sum_{k=0}^{m+1} \bm x_{k, \rm temp}$;

    \For{ $k=0$ {\bf to} $(m+1)$ }
        \State $\bm x_{k, \rm erp} \leftarrow \bm x_{k, \rm temp} - \bar{\bm x}_{\rm temp} + \bar{\bm x}_{\rm erp}$;
    \EndFor

    \State $\bm \Gamma_{\rm erp} \leftarrow \{ \bm x_{0, \rm erp}, \cdots, \bm x_{m+1, \rm erp}, \bm M^{0}_{\rm erp}, \cdots, \bm M^{m}_{\rm erp} \} $;
    \State return $\bm \Gamma_{\rm erp} $;
\end{algorithmic}
\end{algorithm}

\begin{figure*} [tb]
  \centering 
    \includegraphics[width=\textwidth]{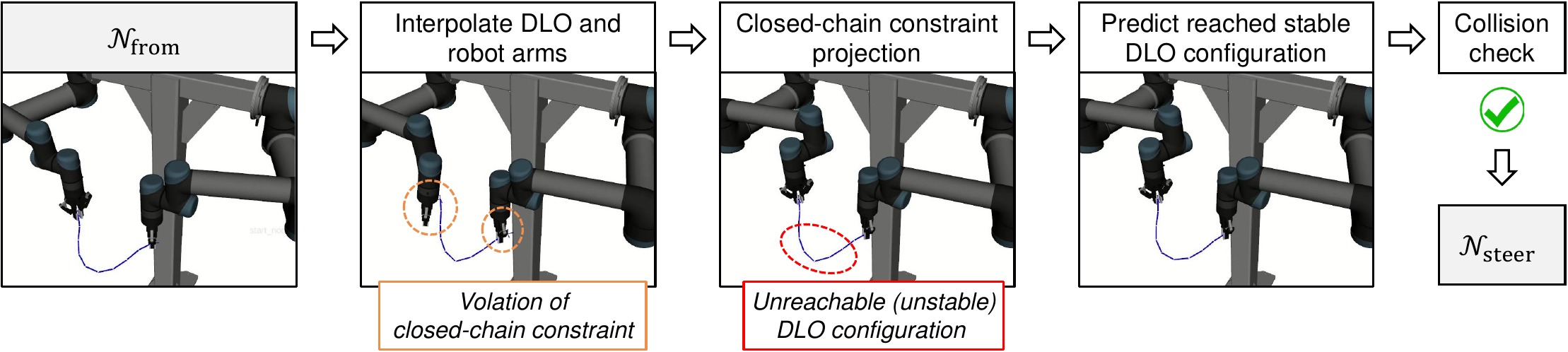} 
  \caption{Illustration of the proposed constrained steering function (Algorithm \ref{algorithm:steering}) in the planning algorithm.}
  \label{fig:steering}
\end{figure*}

\begin{algorithm}[tb] 
\caption{: $\bm q_{\rm new} = \text{ProjectClosedChainConfig} \left(\bm q, \bm \Gamma \right)$}
\label{algorithm:closed_chain_projection}
\begin{algorithmic}[1]
    \Statex \Comment{note that $\bm q = [ \bm q_l; \bm q_r ]$}
    \State $\bm p_l, \bm p_r \leftarrow \text{GetTwoDLOEndPoses}(\bm \Gamma)$;
    \State $i \leftarrow 0$;
    \While{$i < \text{max\_iter}$}
        \State $\bm e_l \leftarrow \bm p_l - \text{LeftArmForwardKinematics}(\bm q_l)$;
        \State $\bm e_r \leftarrow \bm p_r - \text{RightArmForwardKinematics}(\bm q_r)$;
        \If{$\|\bm e_l\|_2 \leq \text{err\_thres}$ \textbf{and} $\|\bm e_r\|_2 \leq \text{err\_thres}$}
            \State $\bm q_{\rm new} \leftarrow \bm q$;
            \State \textbf{return} $\bm q_{\rm new}$;
        \EndIf
        \If{$\|\bm e_l\|_2 > \text{err\_thres}$}
            \State $\bm J^{\rm arm}_l \leftarrow \text{LeftArmJacobian}(\bm q_l)$;
            \State $\bm q_l \leftarrow \bm q_l + (\bm J^{\rm arm}_l)^{\dagger} \bm e_l$;
        \EndIf
        \If{$\|\bm e_r\|_2 \geq \text{err\_thres}$}
            \State $\bm J^{\rm arm}_r \leftarrow \text{RightArmJacobian}(\bm q_r)$;
            \State $\bm q_r \leftarrow \bm q_r + (\bm J^{\rm arm}_r)^{\dagger} \bm e_r$;
        \EndIf
        \State $i \leftarrow i + 1$;
    \EndWhile
    \State \textbf{return} NULL;
\end{algorithmic}
\end{algorithm}

\begin{algorithm}[tb] 
\caption{\hspace{-1mm} : $\mathcal{N}_{\rm steer} \hspace{-1mm} = \hspace{-1mm} \text{ConstrainedSteer} \left( \mathcal{N}_{\rm from}, \mathcal{N}_{\rm to}, \bm \eta_{\rm step} \right)$}
\label{algorithm:steering}
\begin{algorithmic}[1]
   \State \label{line:CalcInterpolationRatio} $\eta_{\rm ratio} \leftarrow \text{CalcInterpolationRatio}(\bm \eta_{\rm step}, \mathcal{N}_{\rm from}, \mathcal{N}_{\rm to})$;
   \State $\bm \Gamma_{\rm erp} \leftarrow \text{DLOerp} \left( \mathcal{N}_{\rm from}.{\bm \Gamma}, \mathcal{N}_{\rm to}.{\bm \Gamma}, \eta_{\rm ratio} \right)$;
   \State $\bm q_{\rm erp} \leftarrow \text{Lerp}\left( \mathcal{N}_{\rm from}.{\bm q}, \mathcal{N}_{\rm to}.{\bm q}, \eta_{\rm ratio} \right)$;
   
   \State $\bm q_{\rm steer} \leftarrow \text{ProjectClosedChainConfig} \left(\bm q_{\rm erp}, \bm \Gamma_{\rm erp} \right)$; \label{line:project_closed_chain_config}
   
   \State $\bm \Gamma_{\rm steer} \leftarrow  \text{ForwardPred}(\bm \Gamma_{\rm from}, \text{TwoEndPoses}(\bm \Gamma_{\rm erp}))$;
   \State $\mathcal{N}_{\rm steer} \leftarrow \{ \bm \Gamma_{\rm steer}, \bm q_{\rm steer}\}$;
    \If{$\bm q_{\rm steer} \neq \text{NULL}$ and $\text{Collision}(\mathcal{N}_{\rm steer}, \mathcal{O}) = \text{false}$}
        \State \textbf{return} $\mathcal{N}_{\rm steer}$;
    \Else
        \State \textbf{return} NULL;
    \EndIf
\end{algorithmic}
\end{algorithm}

As shown in Fig. \ref{fig:steering}, after separately interpolating the configuration of the DLO and dual arm, the stable DLO constraint $\bm C_{\rm stable}(\bm \Gamma) = \bm 0$ and closed-chain constraint $\bm C_{\rm chain}(\bm \Gamma, \bm q_l, \bm q_r) = \bm 0$ are violated.
To satisfy the closed-chain constraint, a straightforward way is to set the $\bm q_{\rm steer}$ as a pair of robot IK solutions of the two DLO end poses \citep{sintov2020motion,yu2023acoarse}. However, it does not utilize the target robot configuration $\mathcal{N}_{\rm to}.\bm q$ to guide the search direction. Considering the non-uniqueness of IK solutions, this may yield a final reached node whose DLO configuration is identical to that of $\mathcal{N}_{\rm to}$ but the robot configuration is different.
Instead, inspired by \citet{berenson2011task}, we use a method to project the interpolated robot configuration to a near one that satisfies the constraint, denoted as $\bm q_{\rm new} = \text{ProjectClosedChainConfig}\left(\bm q, \bm \Gamma \right)$. 
The projection (Algorithm \ref{algorithm:closed_chain_projection}) uses the pseudo-inverse of the arm Jacobian $(\bm J^{\rm arm})^{\dagger}$ to iteratively move the dual-arm robot to reach the two DLO ends by its end-effectors.
This method achieves more effective constrained search guided by $\mathcal{N}_{\rm to}$, especially for redundant robot arms.

We may apply $\bm \Gamma_{\rm steer} = \text{ProjectStableDLOConfig}(\bm \Gamma_{\rm erp})$ to get a steered DLO configuration satisfying the stable DLO constraint. 
However, the calculation of $\bm \Gamma_{\rm erp}$ 
assumes the DLO configurations are full-actuated, whereas manipulating DLOs is actually under-actuated.
To address it, we use the DER model as an approximate kinodynamic model to predict the DLO configuration after robot movements. Specifically, we set the initial value of ProjectStableDLOConfig() as $\bm \Gamma_{\rm from}$, where the end poses of $\bm \Gamma_{\rm from}$ are replaced by those of $\bm \Gamma_{\rm erp}$. The forward prediction is denoted as
\begin{equation}
\begin{aligned}
     \bm \Gamma_{\rm steer} 
    & = \text{ForwardPred}(\bm \Gamma_{\rm from}, \text{TwoEndPoses}(\bm \Gamma_{\rm erp}))
    \\
     = & \text{ProjectStableDLOConfig}(\bm x_{(2, \rm from)}, \cdots , \bm x_{(m-1, \rm from)}  
    \\
    & \quad \quad \quad \quad , \bm x_{(1, \rm erp)}, \bm x_{(m, \rm erp)}, \bm M^0_{\rm erp}, \bm M^{m}_{\rm erp}) .
\end{aligned}
\end{equation}
From another perspective, the purpose of DLOerp() is to efficiently calculate suboptimal constrained robot motion to bring the DLO from $\bm \Gamma_{\rm from}$ towards $\bm \Gamma_{\rm to}$, as the calculation of exactly optimal solutions is too computationally expensive.

The whole process of our constrained steering function $\mathcal{N}_{\rm steer} = \text{ConstrainedSteer} \left( \mathcal{N}_{\rm from}, \mathcal{N}_{\rm to}, \bm \eta_{\rm step} \right)$ is summarized in Algorithm \ref{algorithm:steering} and illustrated in Fig. \ref{fig:steering}.
Note that the interpolation ratio $\eta_{\rm ratio} \in [0, 1]$ is calculated according to the maximum step size $\bm \eta_{\rm step}$ (Line \ref{line:CalcInterpolationRatio}) that consists of the maximum translation step size (unit: $\rm m$), rotation step size (unit: $\rm rad$), and robot joint position step size (unit: $\rm rad$).

\begin{algorithm}[tb] 
\caption{: $\mathcal{N}_{\rm reach} = \text{Extend}(\mathcal{T}, \mathcal{N}_{\rm from}, \mathcal{N}_{\rm to})$}
\label{algorithm:extend}
\begin{algorithmic}[1]
    \Statex \Comment{dist() is defined in (\ref{eq:dist_between_nodes}); Dist() is defined in (\ref{eq:vectorial_dist_between_nodes});}
    
    \State $\mathcal{N}_{\rm s} \leftarrow \mathcal{N}_{\rm from}$; $\mathcal{N}_{\rm last} \leftarrow \mathcal{N}_{\rm from}$;
    
    \While {true}
        \If{$\mathcal{N}_{\rm s} = \mathcal{N}_{\rm to}$}
            \State \textbf{return} $\mathcal{N}_{\rm s}$;
        \EndIf

         \If {$\text{dist}(\mathcal{N}_{\rm s}, \mathcal{N}_{\rm to}) \geq                  \text{dist}(\mathcal{N}_{\rm last}, \mathcal{N}_{\rm to})$} \label{line:judgment_for_extend}
            \State \textbf{return} $\mathcal{N}_{\rm last}$; 
        \EndIf

        \State $\mathcal{N}_{\rm last} \leftarrow \mathcal{N}_{\rm s}$;

        \State $\mathcal{N}_{\rm s} \leftarrow \text{ConstrainedSteer}(\mathcal{N}_{\rm s}, \mathcal{N}_{\rm to}, \bm \eta_{\rm step})$;

        \If{$\mathcal{N}_{\rm s} \neq \text{NULL}$ \textbf{and}
            \Statex \quad \quad \quad  \quad
            $\text{Collision}(\mathcal{N}_{\rm last}, \mathcal{N}_{\rm s}, \mathcal{O}) =            \text{false}$ \textbf{and}
            \Statex \quad \quad \quad \quad 
            $\text{Dist}(\mathcal{N}_{\rm last},                 \mathcal{N}_{\rm s})   \prec  2 \bm \eta_{\rm step}$} \label{line:Dist_less_than_step_size} 
            
            \State $\mathcal{T}.\text{AddVertex}(\mathcal{N}_{\rm s})$;
            \State $\mathcal{T}.\text{AddEdge}(\mathcal{N}_{\rm last}, \mathcal{N}_{\rm s})$;

        \Else
            \State \textbf{return} $\mathcal{N}_{\rm last}$;
        \EndIf
        
    \EndWhile
\end{algorithmic}
\end{algorithm}

Based on the one-step steering function, we define the extending function (Algorithm \ref{algorithm:extend}), which extends the exporation tree from $\mathcal{N}_{\rm from}$ towards $\mathcal{N}_{\rm to} $ as far as possible by iteratively using the steering function and returns the finally reached node $\mathcal{N}_{\rm reach}$. 
Given the nature of constraint projection, similar to \citep{berenson2011task}, a constraint is added in Line \ref{line:judgment_for_extend} to guarantee that the newly generated $\mathcal{N}_{\rm s}$ is closer to the target $\mathcal{N}_{\rm to}$. Another constraint is added in Line \ref{line:Dist_less_than_step_size} to ensure that the distance between $\mathcal{N}_{\rm last}$ and $\mathcal{N}_{\rm s}$ is smaller than the maximum step size allowed.
Then, the new node $\mathcal{N}_{\rm s}$ and edge from $\mathcal{N}_{\rm last}$ to $\mathcal{N}_{\rm s}$ are added to the tree if the steering is successful and the edge is collision-free.

\begin{algorithm}[tb] 
\caption{: $\mathcal{P} = \text{Planning}(\bm \Gamma_{\rm start}, \bm q_{\rm start}, \bm \Gamma_{\rm goal}, \bm q_{\rm goal}, \mathcal{O})$}
\label{algorithm:complete_planning}
\begin{algorithmic}[1]
    \State $\mathcal{T}_A \leftarrow \emptyset, \mathcal{T}_B \leftarrow \emptyset, i \leftarrow 0$;
    \State $\mathcal{N}_{\rm start} \leftarrow \{ \bm \Gamma_{\rm start}, \bm q_{\rm start} \}$;
    \State $\mathcal{T}_A.\text{AddVertex}(\mathcal{N}_{\rm start})$.

    \If{ $\bm q_{\rm goal}  \neq \text{NULL}$ }
        \State $\mathcal{N}_{\rm goal} \leftarrow \{ \bm \Gamma_{\rm goal},    
            \bm q_{\rm goal} \}$;
        \State $\mathcal{T}_B.\text{AddVertex}(\mathcal{N}_{\rm goal})$;
    \Else
        \State $\text{AddRoot}(\mathcal{T}_{B}, \bm \Gamma_{\rm goal}, n_{\rm sg})$; \label{line:goal_sampling}
    \EndIf

    \Statex \Comment{starting iterations}

    \While {$i < \text{max\_iter}$}
        \If {$\text{rand}(0,1) < p_{\rm sg}$} \label{line:goal_sampling_2}
            \State $\mathcal{T}_{\rm goal} = \text{GetBackwardTree}(\mathcal{T}_A, \mathcal{T}_B)$;
            \State $\text{AddRoot}(\mathcal{T}_{\rm goal}, \bm \Gamma_{\rm goal}, 1)$;
        \EndIf \label{line:goal_sampling_3}

        \If {$\text{rand}(0,1) < p_{\rm ts}$} \label{line:task_space_exploring_1}
            \State $\mathcal{N}_{\rm rand} \leftarrow \text{RandomSampleInTaskSpace}()$; \label{line:task_space_exploring_2}
        \Else
            \State $\mathcal{N}_{\rm rand} \leftarrow \text{RandomSampleInFullSpace}()$;
        \EndIf 

        \State $\mathcal{N}_{\rm near}^{A} \leftarrow \text{NearestNeighbor}(\mathcal{T}_A, \mathcal{N}_{\rm rand}) $;

        \State $\mathcal{N}_{\rm reach}^{A} \leftarrow \text{Extend}(\mathcal{T}_A, \mathcal{N}_{\rm near}^{A}, \mathcal{N}_{\rm rand}) $;

        \State $\mathcal{N}_{\rm near}^{B} \leftarrow \text{NearestNeighbor}(\mathcal{T}_B, \mathcal{N}_{\rm reach}^{A}) $;

        \State $\mathcal{N}_{\rm reach}^{B} \leftarrow \text{Extend}(\mathcal{T}_B, \mathcal{N}_{\rm near}^{B}, \mathcal{N}_{\rm reach}^{A})$;

        \Statex \quad \Comment{whether a feasible path is found}

        \If {$\mathcal{N}_{\rm reach}^{A} = \mathcal{N}_{\rm reach}^{B}$}
            \State $\mathcal{P} \leftarrow \text{ExtractPath}(\mathcal{T}_A, \mathcal{N}_{\rm reach}^{A}, \mathcal{T}_B, \mathcal{N}_{\rm reach}^{B})$;
            \State $\mathcal{P} \leftarrow \text{ShortenPath}(\mathcal{P})$;
            \State \textbf{return} $\mathcal{P}$;
        \Else
            \Statex \quad \quad \Comment{encouraging the smaller tree to explore}
            \If{$|\mathcal{T}_A| > |\mathcal{T}_B|$} 
                \State $\text{Swap}(\mathcal{T}_A, \mathcal{T}_B)$;
            \EndIf
            \State $i \leftarrow i + 1$;
        \EndIf
        
    \EndWhile
    
    \State\textbf{return} fail;
    
\end{algorithmic}
\end{algorithm}

\subsection{Overall planning algorithm}

The overall planning algorithm is presented as Algorithm \ref{algorithm:complete_planning}. The structure of the algorithm shares similarity with the constrained bi-directional RRT planner for manipulator planning in \citep{berenson2011task}.

First, we initialize the forward and backward exploration trees with the start and goal configuration. 
We use a goal sampling strategy (Lines \ref{line:goal_sampling}, \ref{line:goal_sampling_2}-\ref{line:goal_sampling_3}) when only the goal DLO configuration is specified while the goal robot configuration is not. This strategy is described in Appendix C.

Then, the trees start growing.
We introduce an exploration strategy (Lines \ref{line:task_space_exploring_1}-\ref{line:task_space_exploring_2}) named \textit{assistant task-space guided exploration} to further accelerate exploration, as described in Appendix C.
In each iteration, the algorithm samples a random node $\mathcal{N}_{\rm rand}$. Then, one of the trees grows towards $\mathcal{N}_{\rm rand}$ from its nearest node $\mathcal{N}_{\rm near}^{A}$, using the Extend() function. The Extend() moves as close to $\mathcal{N}_{\rm rand}$ as possible until the constraints are violated, ultimately reaching $\mathcal{N}_{\rm reach}^{A}$.
The other tree then grows greedily towards $\mathcal{N}_{\rm reach}^{A}$. If it reaches $\mathcal{N}_{\rm reach}^{A}$, the two trees are connected and a feasible path is found. Otherwise, the next iteration begins, with the smaller tree being set as $\mathcal{T}_A$ to encourage balanced growth of the two trees.

Finally, the found feasible path is shortened using ShortenPath(), which is a constrained version of short-cut methods like Algorithm 4 in \citep{berenson2011task}, in which the Extend() function is based on Algorithm \ref{algorithm:extend} in this article.

\begin{figure} [tb]
  \centering 
    \includegraphics[width=\linewidth]{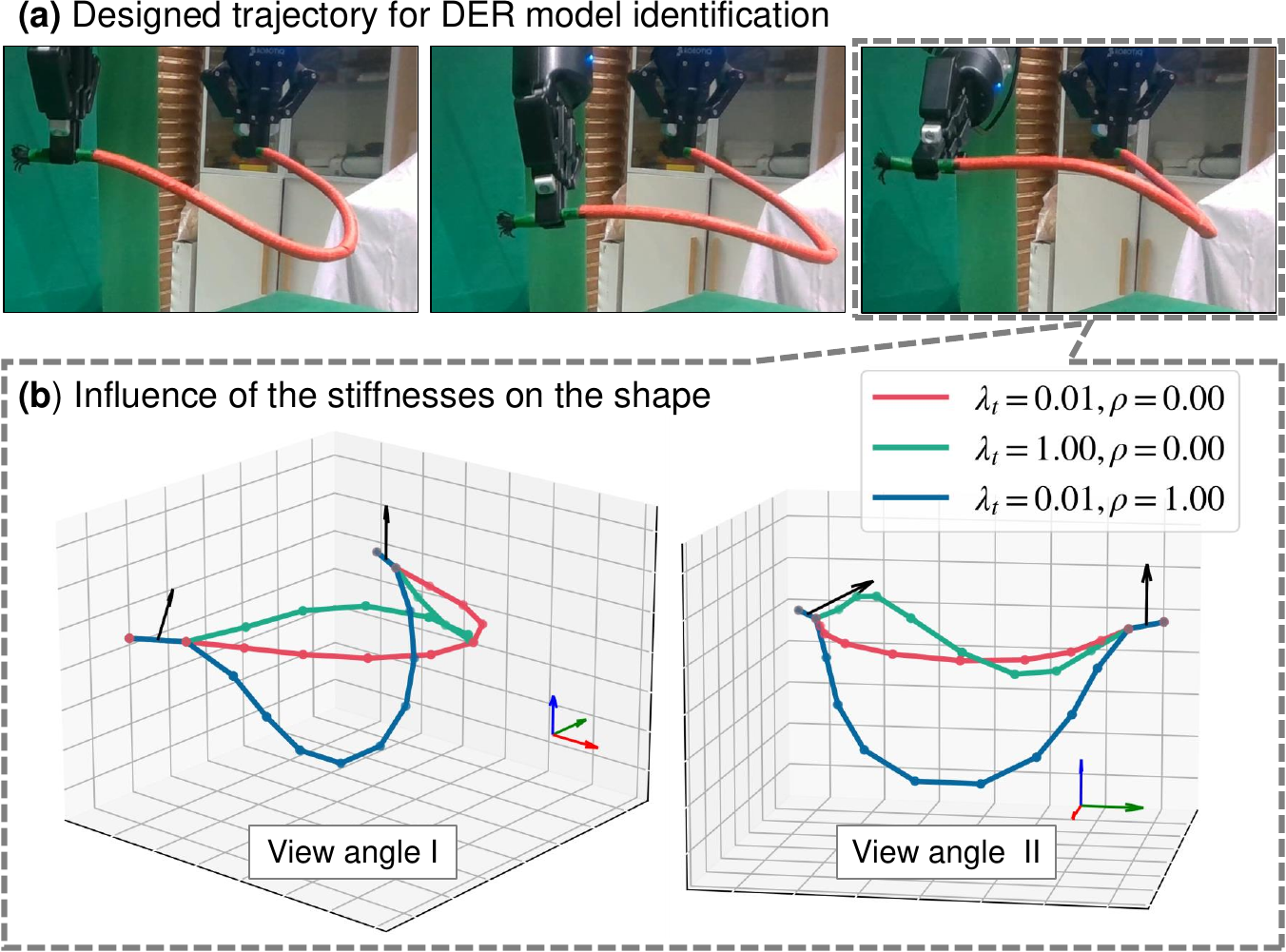} 
  \caption{(a) Designed robot trajectory to collect data for efficiently identify the DER model parameters. (b) The projected stable DLO configurations by (\ref{eq:dlo_projection}) using different DER model parameters when the DLO end poses are constrained as the final poses of the designed trajectory, where the black arrows refer to the Z-axis of the end orientations.}
  \label{fig:derm_identification}
\end{figure}

\subsection{Identification for the DER model parameters} 

The simplified DER model consists of only three scalar parameters: bending stiffness $\lambda_{\rm b}$, twisting stiffness $\lambda_{\rm t}$, and density $\rho$. We design an efficient approach to coarsely identify these parameters in advance for better planning results.
Note that only the relative scale between these parameters affects the results of optimization (\ref{eq:dlo_projection}), so in practice we fix $\lambda_{\rm b}$ and set $\lambda_{\rm t}$ and $\rho$ as variables.
We first collect shapes of the manipulated DLO, and then apply the particle swarm optimization (PSO) method \citep{kennedy1995particle} to minimize the differences between the observed and projected DLO configurations. 

The variable is represented as $\Theta = [\log{ \lambda_{\rm t}}; \log{\rho}]$. The collected $n$ DLO configurations are denoted as $\bm \Gamma_i (i=1,\cdots,n)$. Then, the cost function of the PSO is defined as
\begin{equation}
    L_{\rm pso}(\Theta) 
    =
    \sum_{i=1}^{n} \text{dist}^{\rm p}_{\infty}(\bm \Gamma_{i}, \text{ProjectStableDLOConfig}_{\Theta}(\bm \Gamma_{i})) .
\end{equation}

As shown in Fig. \ref{fig:derm_identification}, we design a simple trajectory to collect the data, during which the twisting and gravity significantly affect the DLO shape, enabling more accurate identification. 
Notably, the identified model is still a coarse and imprecise model, as the used DER model has been simplified.

\section{Closed-loop manipulation}

The planned path is executed in a closed-loop manner to compensate for the simplifications made during planning.
We design an MPC to track the planned path as guidance while adjusting the robot motion based on the real-time feedback to ensure that the actual path of both the DLO and robot is collision-free, and the final desired DLO configuration is precisely reached.

\subsection{MPC design}

The ultimate goal of the controller is to move the DLO to the final desired configuration. However, owing to the ``local'' property of control, it cannot accomplish complex tasks in constrained environments on its own. Therefore, the planned global path is utilized as the tracking objective of the controller. We interpolate the planned path and assign a time stamp to each waypoint to obtain a desired DLO trajectory $\{ \bm \Gamma_{\rm d}^1, \cdots, \bm \Gamma_{\rm d}^N \}$ and corresponding desired robot trajectory $\{ \bm q_{\rm d}^1, \cdots, \bm q_{\rm d}^N \}$, where $N$ is the length of the trajectory and the time interval between waypoints is $\delta t$.

The designed controller must
\begin{enumerate}
    \item track both the planned DLO path and robot path;
    \item ensure that no collision or overstretch occurs during manipulation in constrained environments;
    \item obtain solutions fast to ensure real-time performance.
\end{enumerate}
To achieve these objectives, we formulate the controller as an MPC, which optimizes the control inputs during a finite time horizon $T$ from the current time $t$ to minimize the cost function while satisfying the constraints.
 The variables of the optimization problem include the control inputs $\bm u[0 : T-1]$ and states $\bm x[0 : T]$ and $\bm q[0 : T]$. Note that $\bm x$ is the centerline of $\bm \Gamma$. Frames $\bm M^0$ and $\bm M^{m}$ are not set as variables since they are determined by the robot forward kinematics.


\subsubsection{Objective}
Denoting the tracking objective of the finite-time-horizon MPC at step $t$ as the $(i_{\rm d} + 1)^{\rm th}$ waypoint to $(i_{\rm d} + T)^{\rm th}$ waypoint of the planned DLO and robot trajectory, we define the cost function as 
\begin{equation} \label{eq:mpc_objective}
\begin{aligned}
    J &= \frac{\beta_{x}}{2} \sum_{i = 1}^{T}   \tilde{\bm x}[i]^\transpose \bm W_{x} \tilde{\bm x}[i]  
    + \frac{\beta_{q}}{2} \sum_{i = 1}^{T}  \tilde{\bm q}[i]^\transpose \bm W_{q} \tilde{\bm q}[i] 
    \\
    & + \frac{\beta_{u}}{2} \sum_{i = 0}^{T-1} \bm u[i]^\transpose \bm W_u \bm u[i]
    + \frac{\beta_{a}}{2} \sum_{i = 0}^{T-1} \bm a[i]^\transpose \bm W_a \bm a[i] ,
\end{aligned}
\end{equation}
where 
\begin{equation} \label{eq:dlo_tracking_error}
    \tilde{\bm x}[i] = {\bm x}[i] - {\bm x}_{\rm d}^{i_{\rm d} +i},
\end{equation}
\begin{equation}
    \tilde{\bm q}[i] = {\bm q}[i] - {\bm q}_{\rm d}^{i_{\rm d} +i},
\end{equation}
\begin{equation}
    {\bm a}[i] = \left({\bm u}[i] - {\bm u}[i-1] \right) / \delta t .
\end{equation}
The term $\tilde{\bm x}$ and $\tilde{\bm q}$ denotes the error of the DLO centerlines and robot joint angles, respectively; and $\bm a$ represents the acceleration of the robot.
In the cost function, the first and second term is for tracking the desired DLO and robot trajectory, respectively; the third term is for minimizing the actuation effort; and the fourth term is for minimizing the changes of the actuation effort to ensure smooth execution. Note that $\bm u[-1]$ is the executed control input at the last step. The scalar parameters $\beta_x, \beta_q, \beta_u, \beta_a$ are the weighting coefficients for combining these cost terms, and $\bm W_{x}, \bm W_{q}, \bm W_{u}, \bm W_{a}$ are for weighting the different dimensions of each corresponding cost term.


\subsubsection{State prediction}
The MPC requires a transition model of the system for state prediction. As we assume the manipulation to be quasi-static and the control to be kinematical, the transition model of the dual-arm robot can be expressed as 
\begin{equation}
    \bm q[i+1] = \bm q[i] + \bm u[i] \delta t .
\end{equation}
As for the transition model of the DLO, we use the DLO Jacobian model (\ref{eq:DLO_jacobian_model}) and discretize it as 
\begin{equation}
    \bm x[i+1]  \approx \bm x[i] + \bm J^{\rm dlo}(\bm x[i], \bm r[i]) \, \bm \nu[i] \, \delta t .
\end{equation}
According to the robot kinematics, we have 
\begin{equation}
    \underbrace{
    \left[ \begin{array}{c}
         \bm v_l \\
         \bm \omega_l \\
         \bm v_r \\
         \bm \omega_r
    \end{array} \right]
    }_{\bm \nu}
    =
    \underbrace{
    \left[ \begin{array}{cc}
         \bm J^{\rm arm}_l(\bm q_l) & \bm 0 \\
         \bm 0 &  \bm J^{\rm arm}_r(\bm q_r)
    \end{array} \right]
    }_{\bm J^{\rm arm}(\bm q)}
    \underbrace{
    \left[ \begin{array}{c}
         \dot{\bm q}_{l}  \\
         \dot{\bm q}_{r}
    \end{array} \right]
    }_{\dot{\bm q}} ,
\end{equation}
where $\bm J^{\rm arm}_l$ and $\bm J^{\rm arm}_r$ is the Jacobian matrix of the left and right arm, respectively. Then, it is obtained that 
\begin{equation}
    \bm x[i+1]  \approx \bm x[i] + \bm J^{\rm dlo}(\bm x[i], \bm r[i]) \bm J^{\rm arm}(\bm q[i]) \, \bm u[i] \, \delta t .
\end{equation}
We employ our previously proposed offline-online data-driven method \citep{yu2023global} to obtain an estimated $\hat{\bm J}^{\rm dlo}$, which is further updated online using data collected during actual manipulations.

\subsubsection{Constraints}
1) Avoiding obstacles: 
The MPC must ensure that the robot arm bodies and DLO do not collide with obstacles. 
Although a few existing DLO controllers consider obstacle avoidance \citep{berenson2013manipulation,ruan2018accounting,mcconachie2018estimating}, they focus only on obstacle avoidance of grippers without that of arm bodies and deformable objects.
Our previous work \citep{yu2023acoarse} attempted on it by using artificial potential methods as soft constraints. However, it is difficult to specify appropriate potential functions and weighting coefficients, especially for narrow spaces. To address the issues, we model them as hard constraints in this work. 
We do not introduce any assumption on the static environment, such as the convexity of obstacles, to enable our method to work in any environment. The cost is that the calculation of distances to non-convex obstacles is too time-consuming to be included in optimization iterations. Our solution is to pre-compute a signed distance field (SDF) of the environment before planning and manipulation. 
We denote the minimum distance of a point $\bm p$ to obstacles as $d_{\rm o}(\bm p)$, which can be efficiently obtained online by looking up the SDF table and using 3-D linear interpolation.
We model the collision shape of the dual-arm robot as a series of spheres attached to the links. 
The centers of spheres are denoted as $\bm \xi_j (j=1,\cdots,n_{\rm s})$, and the corresponding radii are denoted as $s_{{\rm a},j}$.
We also use spheres to represent the DLO, whose centers $\check{\bm x}_k (k=1,\cdots,m_{\rm e})$ are obtained by linear interpolation between the feature points, and radii are set as $s_{\rm d}$.
Then, the hard constraints for obstacle avoidance is expressed as 
\begin{equation}
    d_{\rm o}(\bm \xi_j) - s_{{\rm a},j} \geq \epsilon_{\rm d}, \quad j = 1, \cdots, n_{\rm s} ,
\end{equation}
\begin{equation}
    d_{\rm o}(\check{\bm x}_k) - s_{\rm d} \geq \epsilon_{\rm d}, \quad k = 1, \cdots, m_{\rm e} ,
\end{equation}
where $\epsilon_{\rm d}$ is the minimum allowed distance to obstacles.

2) Avoiding overstretch: we introduce a constraint for preventing the DLO from being overstretched during manipulations. Since the potential overstretch by obstacles is avoided by the obstacle avoidance constraint, we consider only uncollided situations, in which the distance between the two DLO ends is constrained as
\begin{equation}
    \| \bm x_{m} - \bm x_{1} \|_2 \leq L - \epsilon_{\rm stretch} ,
\end{equation}
where $L$ is the DLO length, and $\epsilon_{\rm stretch}$ is a safety threshold.

\subsubsection{Optimization problem} 
\, 
Given the desired trajectory $\{ \bm x_{\rm d}^{i_{\rm d} + 1}, \cdots, \bm x_{\rm d}^{i_{\rm d} + T} \}$ and $\{ \bm q_{\rm d}^{i_{\rm d} + 1}, \cdots, \bm q_{\rm d}^{i_{\rm d} + T} \}$
as well as the current DLO centerline $\bar{\bm x}$, robot configuration $\bar{\bm q}$, and last control input $\bar{\bm u}$, the optimization problem is defined as 
\begin{equation} \label{eq:mpc}
\begin{aligned}
    & \min_{\tiny \bm U, \bm X, \bm Q}  \quad 
     J
    \\
    \text{s.t.} \quad
    & \bm q[i+1] = \bm q[i] + \bm u[i] \delta t
    , \quad \forall i \in [0, T-1]
    \\
    & \bm x[i+1]  = \bm x[i] + \hat{\bm J}^{\rm dlo}(\bm x[i], \bm r[i]) \bm J^{\rm arm}(\bm q[i]) \, \bm u[i] \, \delta t,
    \\
    & \quad\quad\quad\quad\quad\quad\quad\quad\quad\quad\quad \forall i \in [0,T-1]
    \\
    & d_{\rm o}(\bm \xi_j[i]) - s_{{\rm a},j} \geq \epsilon_{\rm d}, \quad \forall j \in [1, n_{\rm s}], \forall i \in [1, T]
    \\
    & d_{\rm o}(\check{\bm x}_k[i]) - s_{\rm d} \geq \epsilon_{\rm d}, \quad \forall k \in [1, m_{\rm e}], \forall i \in [1, T]
    \\
    & \| \bm x_{m}[i] - \bm x_{1}[i] \|_2 \leq L - \epsilon_{\rm stretch}, \quad \forall i \in [1, T]
    \\
    & \bm u[i] \preceq \bm u_{\rm max}, \quad \forall i \in [0, T-1]
    \\
    & \bm x[0] = \bar{\bm x}
    \\
    & \bm q[0] = \bar{\bm q} 
    ,
\end{aligned}
\end{equation}
where $J$ has been defined in (\ref{eq:mpc_objective}) with $\bm u[-1] = \bar{\bm u}$. The optimization variables include $\bm U = \bm u[0 : T - 1]$, $\bm X = \bm x[0 : T]$, and $\bm Q = \bm q[0 : T]$. 
The term $\bm u_{\rm max}$ denotes the maximum robot joint velocities allowed.

In practice, we include the sphere centers of the collision shapes $\bm \Xi = \bm \xi[1 : T]$ and $\check{\bm X} = \check{\bm x}[1 : T]$ in the optimization variables. Correspondingly, we add robot forward kinematics constraints $\bm \xi_j = \text{FK}_j(\bm q)$ and DLO edge constraints $\check{\bm x}_k = \text{Lerp}_k(\bm x)$ to the constraints. We find that this approach yields faster solving than directly substituting these constraints into the obstacle avoidance constraints.

We solve this nonlinear problem using a general nonlinear optimization solver.
The optimized $\bm u[0]$ is sent to the robot for execution at time $t$. When moving to the next step $t+1$, the optimization problem is solved again, in which the optimization results of the last step are used as the initial values to accelerate the solving.

In practice, to improve the solving efficiency, we do not incorporate hard constraints for avoiding self-collisions or collisions between the robot and DLO in the MPC (note that they are avoided in planning), as they require real-time distance calculations and increase the dimension of constraints. 
In our experiments, we find that such collisions rarely occur when both the planned DLO and robot paths are tracked by the MPC.
A possible solution to ensure strict collision avoidance is to halt the robot and invoke replanning when such collisions are about to happen in actual manipulations.

\begin{figure*} [tb]
  \centering 
    \includegraphics[width=\textwidth]{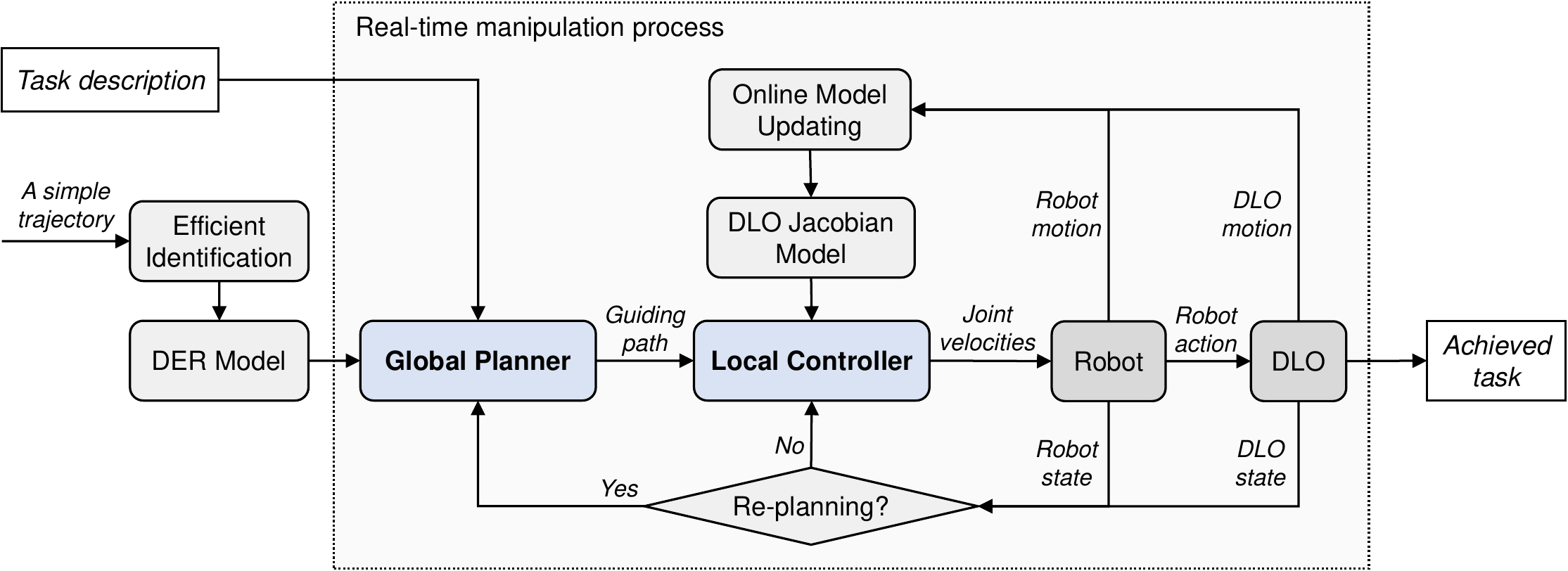} 
  \caption{Overview of the proposed manipulation framework with whole-body global planning and closed-loop executions.}
  \label{fig:manipulation_process}
\end{figure*}

\subsection{Manipulation process}

Owing to the modeling errors, the actual moving path will be different from the planned path. Although the MPC is designed to track the planned path and compensate for the errors, there are some situations that the local controller cannot handle. For example, the local controller may get stuck when there are obstacles exist between the real-time configuration and planned corresponding waypoint, which may occur in complex environments with slender obstacles. In addition, the quasi-static assumption may be violated in some cases, i.e., the DLO may rapidly transition to another shape with a considerably lower energy while the robot executes minor movements. 

To address these extreme situations, we introduce a re-planning strategy. During manipulation, if the controller gets stuck (i.e., the computed control input remains close to zero in the presence of large tracking errors),
we halt the robot and invoke the re-planning module. In addition, if we detect that the DLO shape changes too rapidly, we stop the robot, wait for the DLO to stabilize, and then invoke the re-planning module.
Note that this strategy is reserved for addressing corner cases. 
In practice, we find that almost no cases require re-planning when our controller is used. 

The overall manipulation process is illustrated in Fig. \ref{fig:manipulation_process}. Our method continuously monitors the real-time states of the system and uses the local controller and re-planning strategy to compensate for the modeling simplifications, thereby ensuring the task completion.

\begin{table*}[bp]
\centering
\caption{Key hyper-parameters used in the simulation and the real-world experiments.}
\label{tab:sim_real_hyperparameters}
\begin{tabular}{c|c|c|l} 
\toprule
\multirow{2}{*}{Parameter} & \multicolumn{2}{c|}{Value} & \multicolumn{1}{c}{\multirow{2}{*}{Description}} \\ 
\cline{2-3}
 & Sim. & Real. & \multicolumn{1}{c}{} \\ 
\hline
$m$ & \multicolumn{2}{c|}{10} & Number of DLO feature points \\
$p_{\rm ts}$ & \multicolumn{2}{c|}{0.5} & Probability of using the task-space guided search in each iteration (planner) \\
$n_{\rm sg}$ & \multicolumn{2}{c|}{50} & Number of samples for goal robot configurations before the iteration process (planner) \\
$p_{\rm sg}$ & \multicolumn{2}{c|}{0.1} & Probability of sampling new goal robot configurations in one iteration (planner) \\
$\beta_x$ & \multicolumn{2}{c|}{10.0} & Weight of the cost of DLO tracking errors (controller) \\
$\beta_q$ & \multicolumn{2}{c|}{1.0} & Weight of the cost of robot tracking errors (controller) \\
$\beta_u$ & \multicolumn{2}{c|}{0.1} & Weight of the cost of actuation efforts (controller) \\
$\beta_a$ & 0.1 & 0.5 & Weight of the cost of changes in actuation efforts (controller) \\
$\epsilon_{\rm d}$ & 0.01 & 0.015 & Minimum allowed distance to obstacles (m) (planner \& controller) \\
$\epsilon_{\rm stretch}$ & \multicolumn{2}{c|}{0.01} & Safety threshold in the constraints for avoiding overstretch (m) (controller) \\
$1/\delta t$ & \multicolumn{2}{c|}{5} & Control frequency (Hz) (controller) \\
$T$ & 3 & 5 & Length of the MPC horizon (controller)  \\
\bottomrule
\end{tabular}
\end{table*}

\section{Results}

\subsection{Implementation details}

All algorithms are implemented in C++. The optimization (\ref{eq:dlo_projection}) for stable DLO configuration projection is performed using the Ceres solver \citep{Agarwal_Ceres_Solver_2022}. Although the Ceres is mainly designed for unconstrained nonlinear problems, we find it to be considerably faster than other general nonlinear optimization solvers. Therefore, we select it and implement the constraints as penalty terms in the cost function. 
The optimization (\ref{eq:mpc}) of the MPC is performed using the Ipopt solver \citep{wachter2006implementation} with the Ifopt interface \citep{ifopt}.
We use some functions of the MoveIt! framework \citep{moveit} for convenience, including the establishment of planning scenes, calculation of robot kinematics, collision check, and calculation of distance field. 
In planning, we sample random DLO configurations similar to the sample shown in Fig. \ref{fig:derm_projection}(b), whose end orientations are restricted to be vertically upward along the Z-axis.
In control, we directly use the DLO Jacobian model previously pre-trained offline using 60k data of 10 simulated DLOs \citep{yu2023global}. 

The key hyper-parameters used in the simulations and real-world experiments are listed in Table \ref{tab:sim_real_hyperparameters}. Readers can refer to our released code for other parameters.

\begin{figure*} [tb]
  \centering 
    \includegraphics[width=\textwidth]{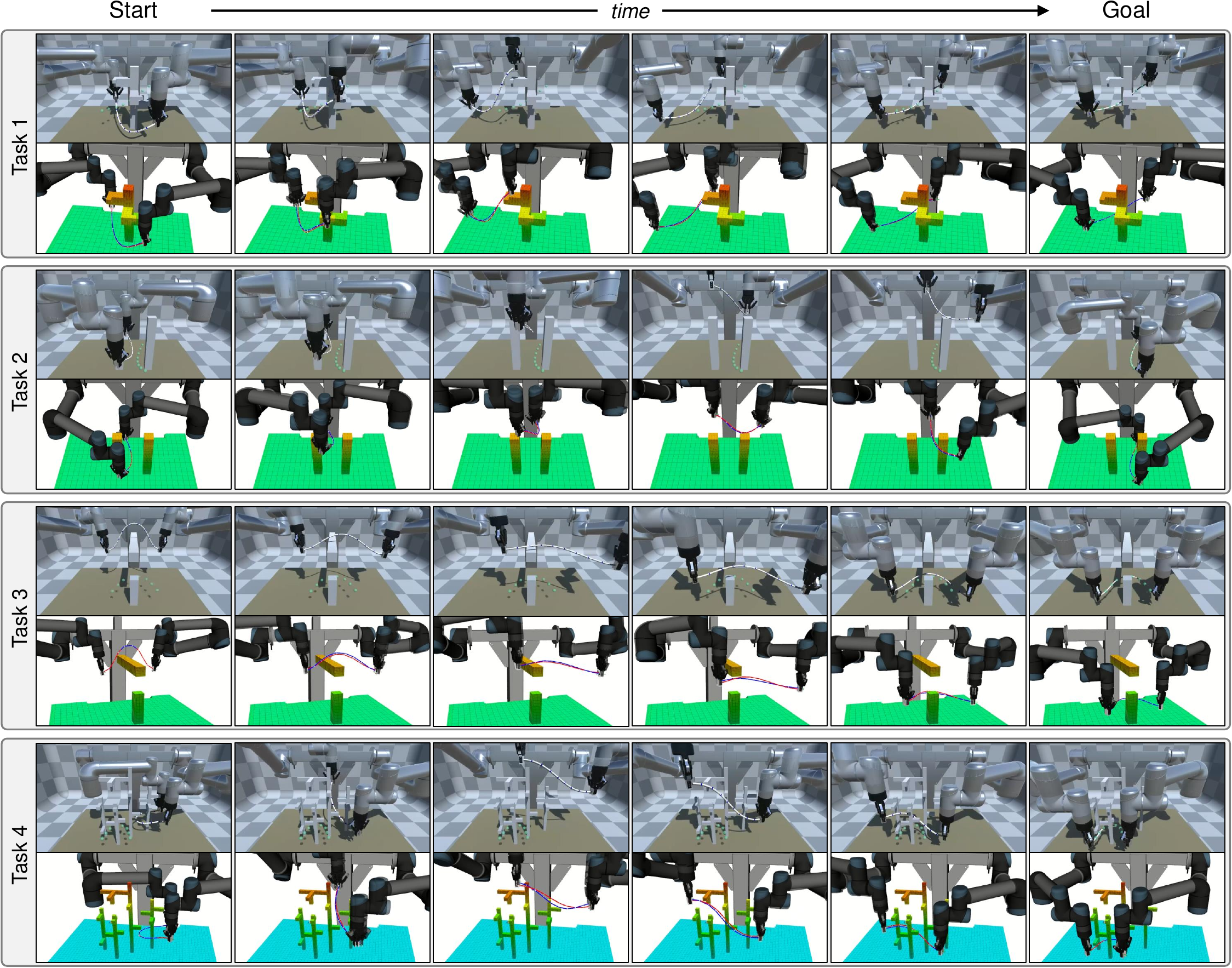} 
  \caption{Four designed tasks in the simulation and the manipulation processes using the proposed method. The start and goal configurations are shown in the pictures in the first and last column, respectively. For each task, the pictures in the first row are taken from the Unity simulator, where the translucent green points indicate the goal positions of the DLO feature points. The pictures in the second row are visualizations by the Rviz, in which the blue lines and translucent robots indicate the planned waypoints, and the red lines and non-translucent robots indicate the real-time configurations.}
  \label{fig:sim_tasks}
\end{figure*}

\subsection{Evaluation metrics}

Following metrics are used for performance evaluation.
The metrics for planning: 
\begin{enumerate}
    \item \textit{Success rate}: a planning is regarded successful if it finds a feasible path within 50,000 RRT iterations;
    \item \textit{Time for finding a feasible path / smoothing}: the time cost for finding a feasible path / smoothing the path; 
    \item \textit{Path length}: the moving distance of the DLO along the path, which is specified as the average moving distance of the DLO feature points.
\end{enumerate}
The metrics for manipulation:
\begin{enumerate}
    \item \textit{Final task error}: the Euclidean distance between the goal DLO configuration and reached configuration in a manipulation, i.e., $\| 
\bm x_{\rm reached} - \bm x_{\rm goal} \|_2$; note that here $\bm x = [\bm x_1; \cdots; \bm x_m]$, which contains all feature points;
    \item \textit{Success rate}: a manipulation is regarded successful if the final task error is less than 5 cm within a maximum time of 180 s and no overstretch occurs;
    \item \textit{Collision time}: the duration (seconds) of collisions with obstacles during a manipulation;
\end{enumerate}
The statistical results are reported in the format of [mean value $\pm$ standard deviation] in the following tables.
For the planning time for finding a feasible path, we report the statistical results of the shortest 80\% of trials to exclude extreme cases. We also report the time cost of all trials in Appendix D.

\subsection{Simulation studies}

The simulation environment is constructed in the Unity \citep{juliani2018unity} with the Obi \citep{obi} for simulating DLOs and the Unity Robotics Hub \citep{unity_robotics_hub} for integration with the ROS \citep{quigley2009ros}.
The DLO is rigidly grasped by a dual-UR5 robot.
The simulator operates as a black box. 
The simulation and algorithms are run on a desktop with an Intel i7-10700 CPU and a 32-GB RAM.

We design four tasks with different obstacles, start/goal configurations, and DLO properties for exhaustive quantitative testing, as shown in Fig. \ref{fig:sim_tasks}. 
Tasks 1 to 3 are relatively simple (yet challenging for existing methods), while Task 4 is considerably more complex.

Before manipulations, the four different DLOs used in the four tasks are coarsely identified to get the corresponding DER model parameters. 
The lengths of the four DLOs used in Tasks 1 to 4 are 0.5, 0.4, 0.6, and 0.5 m, respectively. The identified relative twist stiffnesses ($\lambda_t / \lambda_b$) are 1.1, 1.4, 1.1, and 1.3, respectively; and the identified relative density ($\rho / \lambda_b$) are 46, 43, 0.011, and 4.7, respectively.

\begin{table*}
\centering
\caption{Performance of the proposed global planner in the simulated tasks.}
\label{tab:sim_planner_overall_performance}
\begin{tabular}{c|c|ccccc} 
\toprule
Task & \begin{tabular}[c]{@{}c@{}}Goal robot \\configuration\end{tabular} & Success rate & \begin{tabular}[c]{@{}c@{}}Time for finding \\a feasible path (s)\end{tabular} & \begin{tabular}[c]{@{}c@{}}Feasible path \\length (m)\end{tabular} & \begin{tabular}[c]{@{}c@{}}Time for \\smoothing (s)\end{tabular} & \begin{tabular}[c]{@{}c@{}}Smoothed path \\length (m)\end{tabular} \\ 
\hline
\multirow{2}{*}{1} & w & 100/100 & 1.92~$\pm$ 0.66 & 1.94~$\pm$ 0.39 & 2.33~$\pm$~0.40 & 1.04~$\pm$ 0.20 \\
 & w/o & 100/100 & 1.97~$\pm$ 0.75 & 1.89~$\pm$ 0.39 & 2.29~$\pm$ 0.36 & 1.05~$\pm$ 0.22 \\ 
\hline
\multirow{2}{*}{2} & w & 100/100 & 1.43~$\pm$~0.56 & 1.71~$\pm$ 0.36 & 2.73~$\pm$ 0.34 & 0.86~$\pm$~0.12 \\
 & w/o & 100/100 & 1.38~$\pm$ 1.26 & 1.73~$\pm$ 0.33 & 2.86~$\pm$ 0.45 & 0.86~$\pm$ 0.13 \\ 
\hline
\multirow{2}{*}{3} & w & 100/100 & 3.18~$\pm$ 1.61 & 1.59~$\pm$~0.39 & 2.78~$\pm$~0.45 & 0.79~$\pm$ 0.17 \\
 & w/o & 100/100 & 2.92~$\pm$~2.32 & 1.61~$\pm$ 0.40 & 2.81~$\pm$ 0.40 & 0.78~$\pm$ 0.16 \\ 
\hline
\multirow{2}{*}{4} & w & 100/100 & 8.22~$\pm$ 3.34 & 2.28~$\pm$ 0.55 & 2.61~$\pm$ 0.44 & 1.33~$\pm$ 0.27 \\
 & w/o & 100/100 & 7.94~$\pm$ 3.82 & 2.26~$\pm$ 0.47 & 2.67~$\pm$ 0.47 & 1.33~$\pm$ 0.23 \\
\bottomrule
\end{tabular}
\end{table*}

\begin{table*}
\centering
\caption{Performance of the proposed closed-loop manipulation framework in the simulated tasks.}
\label{tab:sim_manipulation_overall_performance}
\begin{tabular}{c|ccccc} 
\toprule
Task & Success rate & Final task error (mm) & Collision time (s) & Execution time (s) & Total replanning times \\ 
\hline
1 & 100/100 & 0.10~$\pm$ 0.03 & 0.0 $\pm$ 0.0 & 47.93~$\pm$ 9.09 & 0 \\
2 & 100/100 & 0.20 $\pm$ 0.08 & 0.0 $\pm$ 0.0 & 41.02~$\pm$ 7.40 & 0 \\
3 & 100/100 & 0.44 $\pm$ 0.11 & 0.02 $\pm$ 0.15 & 37.45~$\pm$ 6.92 & 0 \\
4 & 100/100 & 0.13 $\pm$ 0.11 & 0.03 $\pm$ 0.15 & 62.51 $\pm$ 12.64 & 1 \\
\bottomrule
\end{tabular}
\end{table*}

\subsubsection{Overall performance of the proposed method}

First, we validate the performance of the proposed global planner, in which we separately test the planning algorithm for situations in which the goal robot configuration is known and unknown. 
For each task, we run the planner 100 times with different random seeds, and the results are summarized in Table \ref{tab:sim_planner_overall_performance}.
The results demonstrate that 
1) our global planner is highly robust, as the planning success rate is  $100\%$ for all 800 trials; 
2) the planner is highly efficient, as the average time for finding a feasible path is about 1 to 3 s for Tasks 1 to 3 and about 10 s for the challenging Task 4; 
and 3) the planning algorithm can effectively handle tasks without specified goal robot configurations, and the performance is similar to the cases with known goal robot configurations. 
The time cost of each key function and distribution of planning time are reported in Appendix D.

\begin{figure} [tb]
  \centering 
    \includegraphics[width=\linewidth]{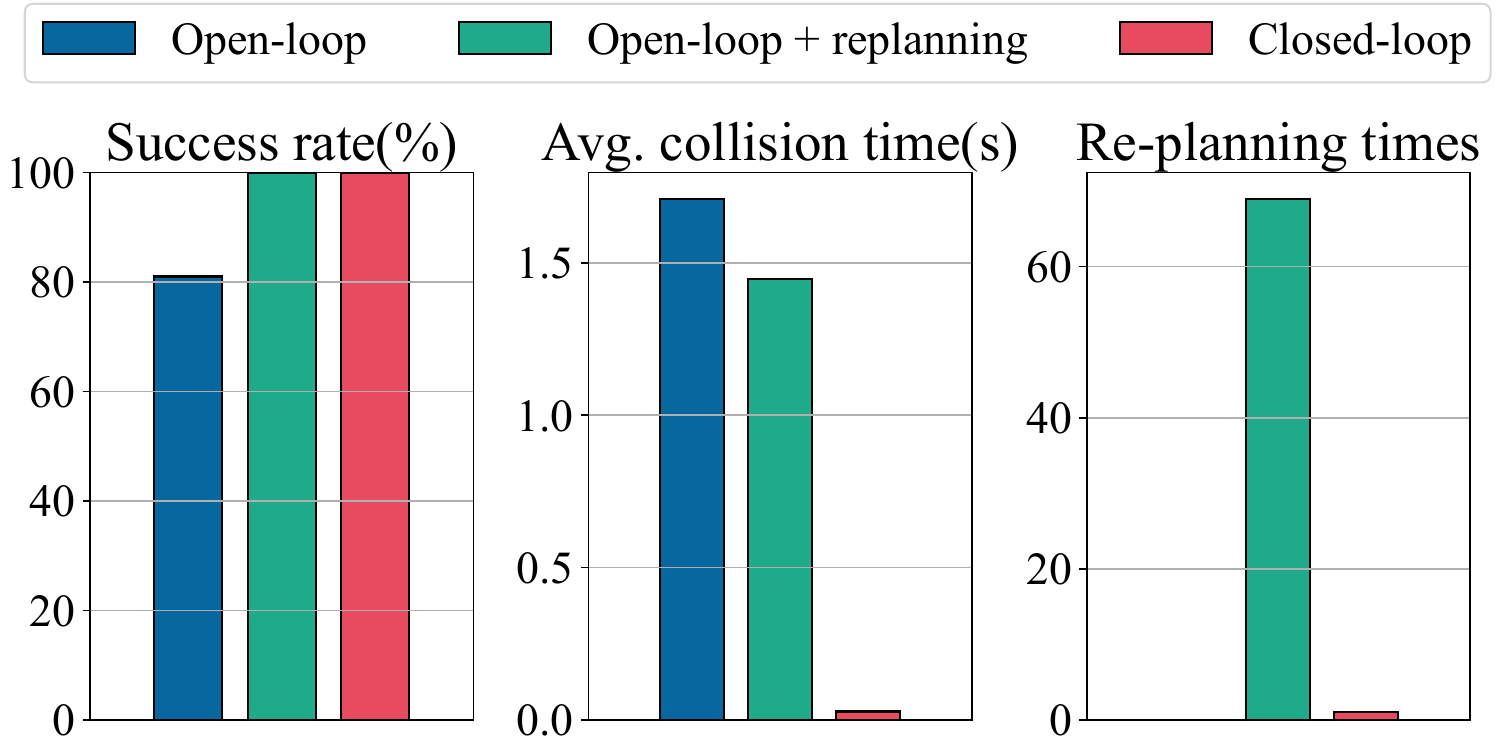} 
  \caption{Comparison between the open-loop, open-loop + replanning, and closed-loop manipulation schemes in the simulated Task 4. The reported collision time and replanning times correspond to the successful cases.}
  \label{fig:sim_closed_open_loop_compare}
\end{figure}

Then, we validate the performance of the proposed manipulation scheme. For each task, we execute 100 different planned paths using the proposed closed-loop manipulation method. 
The manipulation results, summarized in Table \ref{tab:sim_manipulation_overall_performance}, indicate that 1) our method can robustly and precisely accomplish such global manipulation tasks, as all 400 tests are successful and the final task errors are less than 0.5 mm; 
2) our method can effectively avoid collision when using imprecise DLO models, as collision is minimal with an average collision time of less than 0.05 s; 
and 3) the replanning module is invoked only once among all 400 tests, indicating that the extreme situations that cannot be handled by the local controller rarely occur.

\subsubsection{Closed-loop vs. open-loop}

We compare the closed-loop manipulation scheme with an open-loop manner, in which only the planned robot paths are tracked without considering the actual DLO states. 
In the simpler Tasks 1 to 3, we observe that the performance of the open-loop manner is comparable to that of the closed-loop manner, with a success rate $\geq 99\%$. However, it leads to more collisions (with durations of 0.13, 0.15, and 1.59 s in Task 1, 2, and 3, respectively).
Detailed results of Task 4 are presented in Fig. \ref{fig:sim_closed_open_loop_compare}, where the differences are more significant owing to the increased complexity of Task 4. We also show the manipulation results of the open-loop + replanning scheme, in which replanning is invoked if obstacles are detected between the current configuration and corresponding planned waypoint.
The results indicate that 
1) the global planner is reliable, as the open-loop execution achieves a 81\% success rate for such a complicated task;
2) the replanning strategy improves the robustness of the open-loop manner (to 100\% success rate) but at the expense of higher planning cost (totally 65 times of replanning) and also longer execution time.
3) the proposed closed-loop manner also improves the robustness (to 100\% success rate) while invoking the replanning module only once; and
4) the local controller can avoid most potential collisions resulting from imprecise DLO models, with the collision duration being reduced from 1.71 to 0.03 s.
These results confirm the effectiveness of both the standalone global planner and the closed-loop manipulation scheme.

\subsubsection{More ablation study}
We conduct a series of ablation study about the components of our approach, including the stable DLO configuration constraint, closed-chain constraint projection, under-actuation property, assistant task-space guided exploration, and weights of the cost terms in the MPC.
The detailed results are presented in Appendix D.

\begin{table*}
\centering
\setlength{\extrarowheight}{0pt}
\addtolength{\extrarowheight}{\aboverulesep}
\addtolength{\extrarowheight}{\belowrulesep}
\setlength{\aboverulesep}{0pt}
\setlength{\belowrulesep}{0pt}
\caption{Comparison between using the DER model and \citet{bretl2014quasi} for planning in the simulated tasks.}
\label{tab:sim_comparison_with_bretl2014}
\scalebox{0.91}{
\begin{threeparttable}[b]
\begin{tabular}{c|c|ccc|ccc} 
\toprule
Task & Method & \begin{tabular}[c]{@{}c@{}}Planning\\success rate $\uparrow$\end{tabular} & \begin{tabular}[c]{@{}c@{}}Planning \\time (s)\tnote{a} ~$\downarrow$\end{tabular} & \begin{tabular}[c]{@{}c@{}}Path \\length (m)\tnote{b} 
 ~$\downarrow$\end{tabular} & \begin{tabular}[c]{@{}c@{}}Open-loop manipulation\\success rate~$\uparrow$\end{tabular} & \begin{tabular}[c]{@{}c@{}}Collision \\time (s)\tnote{c} ~$\downarrow$\end{tabular} & \begin{tabular}[c]{@{}c@{}}Execution \\time (s)~$\downarrow$\end{tabular} \\ 
\hline
\multirow{2}{*}{1} & Bretl et al. & 100/100 & 13.21 $\pm$ 6.58 & 2.02 $\pm$ 0.58 & 62/100 & 2.58~$\pm$ 3.44 & 119.77 $\pm$~23.5 \\
 & {\cellcolor[rgb]{0.902,0.902,0.902}}Ours & {\cellcolor[rgb]{0.902,0.902,0.902}}100/100 & {\cellcolor[rgb]{0.902,0.902,0.902}}1.92 $\pm$ 0.75 & {\cellcolor[rgb]{0.902,0.902,0.902}}1.04 $\pm$ 0.20 & {\cellcolor[rgb]{0.902,0.902,0.902}}100/100 & {\cellcolor[rgb]{0.902,0.902,0.902}}0.13~$\pm$ 0.60 & {\cellcolor[rgb]{0.902,0.902,0.902}}52.84 $\pm$~9.14 \\ 
\hline
\multirow{2}{*}{2} & Bretl et al. & 100/100 & 7.00 $\pm$ 4.64 & 1.22~$\pm$ 0.30 & 92/100 & 4.37~$\pm$ 3.53 & 76.34~$\pm$ 14.79 \\
 & {\cellcolor[rgb]{0.902,0.902,0.902}}Ours & {\cellcolor[rgb]{0.902,0.902,0.902}}100/100 & {\cellcolor[rgb]{0.902,0.902,0.902}}1.43 $\pm$ 0.56 & {\cellcolor[rgb]{0.902,0.902,0.902}}0.86 $\pm$ 0.12 & {\cellcolor[rgb]{0.902,0.902,0.902}}99/100 & {\cellcolor[rgb]{0.902,0.902,0.902}}0.15~$\pm$ 0.60 & {\cellcolor[rgb]{0.902,0.902,0.902}}46.00 $\pm$ 7.30 \\ 
\hline
\multirow{2}{*}{3} & Bretl et al. & 100/100 & 21.01 $\pm$ 12.72 & 1.29~$\pm$ 0.42 & 100/100 & 0.26~$\pm$ 0.89 & 101.53~$\pm$ ~20.09 \\
 & {\cellcolor[rgb]{0.902,0.902,0.902}}Ours & {\cellcolor[rgb]{0.902,0.902,0.902}}100/100 & {\cellcolor[rgb]{0.902,0.902,0.902}}3.18 $\pm$ 1.61 & {\cellcolor[rgb]{0.902,0.902,0.902}}0.79 $\pm$ 0.17 & {\cellcolor[rgb]{0.902,0.902,0.902}}100/100 & {\cellcolor[rgb]{0.902,0.902,0.902}}1.59~$\pm$ 2.02 & {\cellcolor[rgb]{0.902,0.902,0.902}}42.14 $\pm$ 6.83 \\ 
\hline
\multirow{2}{*}{4} & Bretl et al. & 93/100 & 46.02 $\pm$ 43.88 & 1.81~$\pm$ 0.31 & 74/93 & 1.81~$\pm$ 2.29 & 125.31~$\pm$ 18.42 \\
 & {\cellcolor[rgb]{0.902,0.902,0.902}}Ours & {\cellcolor[rgb]{0.902,0.902,0.902}}100/100 & {\cellcolor[rgb]{0.902,0.902,0.902}}8.22 $\pm$ 3.34 & {\cellcolor[rgb]{0.902,0.902,0.902}}1.33 $\pm$ 0.27 & {\cellcolor[rgb]{0.902,0.902,0.902}}81/100 & {\cellcolor[rgb]{0.902,0.902,0.902}}1.71~$\pm$ 2.34 & {\cellcolor[rgb]{0.902,0.902,0.902}}65.98 $\pm$ 11.81 \\
\bottomrule
\end{tabular}
\begin{tablenotes}
     \item[a] Mean value ± standard deviation of the time (s) for finding a feasible path over the successful planning cases.
     \item[b] Mean value ± standard deviation of the path length (m) after smoothing over the successful planning cases.
     \item[c] Mean value ± standard deviation of the collision time (s) over the successful manipulation cases.
\end{tablenotes}
\end{threeparttable}
}
\end{table*}

\subsection{Comparison with the existing methods}

After presenting our methodology and demonstrating its overall performance, we would like to further provide a detailed comparison between our method and the representative existing methods.

\subsubsection{DLO model}

We compare our used DER model with the analytical model proposed by \citet{bretl2014quasi} for planning. Their analytical model can directly sample equilibriums of a one-end-fixed DLO from a six-dimensional chart (which is called $\mathcal{A}$), by solving multiple differential equations to check the stability of the sample and map the parameterized sample in the chart to the DLO configuration space in $SE(3)$ (which is called $\mathcal{C}$).
However, using this model for planning has some inconvenience.
First, it ignores gravity which may significantly affect the stable DLO configurations. 
Second, this model does not support straight-forward mapping of a DLO configuration from $\mathcal{C}$ to $\mathcal{A}$. However, in practice, the start and goal DLO configurations are given in $\mathcal{C}$, so it is necessary to map the start and goal DLO configurations from $\mathcal{C}$ to $\mathcal{A}$ before doing planning in $\mathcal{A}$. 
Third, the identification of DLO parameters (e.g., stiffness) from actual observations in $\mathcal{C}$ is not easy to achieve, since the gravity effects are not considered, and the parameterized DLO configurations in $\mathcal{A}$ of the observations are unknown.
In addition, it is non-trivial to constrain the motion of the free DLO end-tip when planning in $\mathcal{A}$. Thus, the planned DLO paths usually involve huge changes of the end poses, which may be unreachable by the robot arms.
In contrast, our used DLO model is fully defined in $\mathcal{C}$. It projects a DLO configuration in $\mathcal{C}$ to a neighboring stable configuration with the same end poses. Unlike \citet{bretl2014quasi} which first assigns the stable DLO configuration and then extracts the end poses from it, we first assign the end poses and then obtain the stable DLO configuration (with an initial value). Therefore, the movements of the robot end-effectors in our planned paths are shorter. Furthermore, our model takes gravity into account, which highly improves the planning accuracy.

\begin{figure} [tb]
  \centering 
    \includegraphics[width=\linewidth]{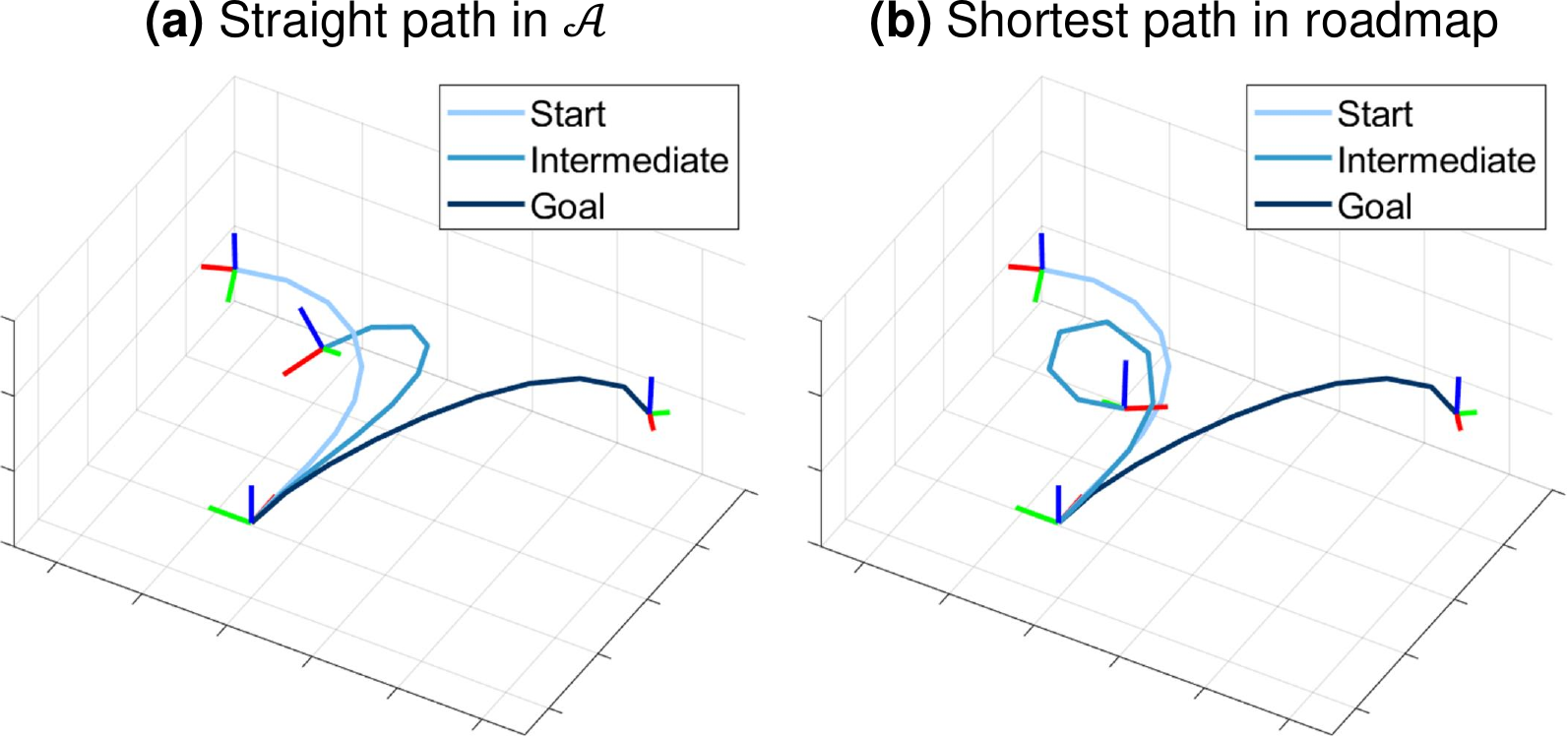} 
  \caption{Different paths from the start to goal DLO configuration using the model in \citet{bretl2014quasi}, shown in $\mathcal{C}$. (a) The straight path in $\mathcal{A}$, where the intermediate configuration refers to the midpoint of the path in $\mathcal{A}$. (b) The searched shortest path in a built roadmap, which contains three vertices of the roadmap.}
  \label{fig:sim_sintov_drawback}
\end{figure}

We quantitatively compare the performances of planning using their model and our used DER model in the simulated tasks. Specifically, we replace the DER model with their model in our planning framework. Regarding the planning algorithm, the only difference is that the sampling or steering of a tree node first generates a six-dimensional parameterized DLO configuration in $\mathcal{A}$ and a pose of the fixed end, then maps them to $\mathcal{C}$, and finally calculates the corresponding robot configuration. We manually assign the DLO stiffnesses and the start and goal DLO configurations in $\mathcal{A}$. 
The performance of planning and open-loop execution is listed in Table \ref{tab:sim_comparison_with_bretl2014}. The results show that
1) the planning success rate is close to ours, demonstrating that our planning algorithm can be used with different DLO models; 
2) the planning time cost is higher, since the exploration in $\mathcal{A}$ neglects the reachability of the robot arms;
3) the planned paths are longer, since the Euclidean distance in $\mathcal{A}$ may not well reflect the distance in $\mathcal{C}$, which is also illustrated in Fig. \ref{fig:sim_sintov_drawback}(a);
4) due to the neglect of gravity during planning, the success rate and collision time of open-loop manipulation are worse than ours in Task 1 and 2, where the gravity effects are significant; 
and 5) the execution time is longer, since the planned paths usually involve large robot motions.

The advantage of their model is that it is more accurate in zero-gravity scenarios, since it is a continuous model while our simplified DER model involves coarse discretization. The higher accuracy of their model results in more precise open-loop executions in Task 3, where the gravity effect is negligible. However, our method enables more efficient planning, shorter robot paths, and handling of gravity.

\begin{table*}[tb]
\centering
\setlength{\extrarowheight}{0pt}
\addtolength{\extrarowheight}{\aboverulesep}
\addtolength{\extrarowheight}{\belowrulesep}
\setlength{\aboverulesep}{0pt}
\setlength{\belowrulesep}{0pt}
\caption{Comparison between our planning + control framework and \citet{mcconachie2020manipulating} in the simulated tasks.}
\label{tab:sim_comparison_with_mcconachie2020}
\scalebox{0.9}{
\begin{threeparttable}[b]
\begin{tabular}{c|c|cccc} 
\toprule
Task & Method & Planning success rate $\uparrow$ & Manipulation success rate $\uparrow$ & Total replanning time\tnote{a} $\downarrow$ & Collision time (s)\tnote{b} $\downarrow$ \\ 
\hline
\multirow{2}{*}{1} & McConachie et al. & 78/100 & 66/78 & 10 & 12.01 $\pm$ 17.36 \\
 & {\cellcolor[rgb]{0.902,0.902,0.902}}Ours & {\cellcolor[rgb]{0.902,0.902,0.902}}100/100 & {\cellcolor[rgb]{0.902,0.902,0.902}}100/100 & {\cellcolor[rgb]{0.902,0.902,0.902}}0 & {\cellcolor[rgb]{0.902,0.902,0.902}}0.0 $\pm$ 0.0 \\ 
\hline
\multirow{2}{*}{2} & McConachie et al. & 100/100 & 80/100 & 53 & 14.41 $\pm$ 20.26 \\
 & {\cellcolor[rgb]{0.902,0.902,0.902}}Ours & {\cellcolor[rgb]{0.902,0.902,0.902}}100/100 & {\cellcolor[rgb]{0.902,0.902,0.902}}100/100 & {\cellcolor[rgb]{0.902,0.902,0.902}}0 & {\cellcolor[rgb]{0.902,0.902,0.902}}0.0 $\pm$ 0.0 \\ 
\hline
\multirow{2}{*}{3} & McConachie et al. & 91/100 & 87/91 & 57 & 10.77 $\pm$ 13.82 \\
 & {\cellcolor[rgb]{0.902,0.902,0.902}}Ours & {\cellcolor[rgb]{0.902,0.902,0.902}}100/100 & {\cellcolor[rgb]{0.902,0.902,0.902}}100/100 & {\cellcolor[rgb]{0.902,0.902,0.902}}0 & {\cellcolor[rgb]{0.902,0.902,0.902}}0.02 $\pm$ 0.15 \\ 
\hline
\multirow{2}{*}{4} & McConachie et al. & 0/100 & - & - & - \\
 & {\cellcolor[rgb]{0.902,0.902,0.902}}Ours & {\cellcolor[rgb]{0.902,0.902,0.902}}100/100 & {\cellcolor[rgb]{0.902,0.902,0.902}}100/100 & {\cellcolor[rgb]{0.902,0.902,0.902}}1 & {\cellcolor[rgb]{0.902,0.902,0.902}}0.03 $\pm$ 0.15 \\
\bottomrule
\end{tabular}
\begin{tablenotes}
     \item[a] Total replanning times in the successful manipulation cases. In each failed case, replanning is invoked repeatedly until the maximum manipulation time is exceeded.
     \item[b] Mean value ± standard deviation of the collision time (s) over the successful manipulation cases.
\end{tablenotes}
\end{threeparttable}
}
\end{table*}

\subsubsection{Planning algorithm}

\citet{sintov2020motion} also investigated the motion planning considering both the DLO and dual arms.
They pre-built a roadmap of stable DLO configurations using the DLO model \citep{bretl2014quasi}, aiming to reduce the computation of solving the differential equations of the model during online planning. 
In our implementation, we find that these equations can be efficiently solved within 2 ms using the ODEINT C++ library on our CPU. Therefore, building a roadmap ahead benefits little. On the other hand, it has some drawbacks. 
First, building a roadmap requires sampling and connecting a large amount of stable DLO configurations, which is computationally expensive. 
Second, the built roadmap for one DLO cannot be used for another DLO with different properties. It is impractical to re-building the roadmap every time manipulating a new DLO with different parameters.
Third, the planning performance on this high-dimensional problem highly depends on the size of the roadmap. A small roadmap leads to winding paths and planning failures, while a large roadmap requires a large amount of computation time and storage memory.

We provide a case study to demonstrate the disadvantages of planning with roadmaps. 
Fig. \ref{fig:sim_sintov_drawback}(a) shows the straight path in $\mathcal{A}$ from the start to goal DLO configuration (the same as those in the simulated Task 1 but with one end fixed and no gravity). The intermediate configuration refers to the midpoint of the path in $\mathcal{A}$; however, it does not appear to be a midpoint between the start and goal in $\mathcal{C}$, indicating that the Euclidean distance in $\mathcal{A}$ is not a perfect distance metric for DLO planning.
Fig. \ref{fig:sim_sintov_drawback}(b) shows the searched shortest path from the start to goal configuration in a built roadmap that contains 100 vertices (as suggested in \citet{sintov2020motion}), each of which is connected to its 10 nearest neighbors. The shortest path contains three vertices of the roadmap; however, the intermediate configuration appears to be an unnecessary waypoint. This is because the start and goal vertices are not directly connected in the roadmap, so the path has to go through an intermediate vertex.
In $\mathcal{A}$, the distance between the start and goal, start and intermediate, and intermediate and goal vertices is 19.8, 18.1, and 5.4, respectively. 
This case indicates that using an insufficiently large roadmap may not allow high-quality DLO planning, even for such a simple scenario.

\subsubsection{Combination of planning and control}

We compare our method with another framework of combining planning and control \citep{mcconachie2020manipulating} for manipulating deformable objects. 
Their aim and combination approach are different from ours. First, their planner uses a more simplified model named ``virtual elastic band'' focusing on only the overstretching of deformable objects by grippers or obstacles. In this way, the DLO may be over-compressed or hooked by obstacles.
Second, their planned gripper path is executed in an open-loop manner without local adjustment based on real-time feedback, as their controller is only activated when the deformable object is close to the final desired configuration.
Third, collisions between the DLO and obstacles are allowed in their approach.

We compare the methods in the simulated tasks. For each task, their global planner is run 100 times and all successfully planned paths are executed using their manipulation framework. 
The planning and manipulation results are summarized in Table \ref{tab:sim_comparison_with_mcconachie2020}. In the simpler Tasks 1 to 3, their global planner finds feasible paths with a success rate of 89.7\%. We find it is because the planner spends excessive time in searching configurations with contacts, which is avoided in our tasks. 
As their planner considerably simplifies the representations of DLO states, DLO states highly deviating from the expected states often occur during actual manipulations. Thus, their manipulation scheme relies on the real-time deadlock prediction to detect unplanned DLO overstretch and uses the replanning strategy to recover.
We find that some manipulation cases fail because of executing too many times of replanning and finally exceeding the maximum manipulation time (180 s), resulting in a manipulation success rate of 86.6\%. 
In addition, collision is allowed in their method while not in our problem formulation. 
In the more challenging Task 4, their method fails to find a feasible path while our method achieves a 100\% success rate for both planning and manipulation.

The advantages of their approach are that it does not require identification of the DLO model before planning and can also be applied to the manipulation of planar deformable objects (e.g., cloth) in some scenarios.

\begin{figure} [tb]
  \centering 
  \includegraphics[width=0.47\textwidth]{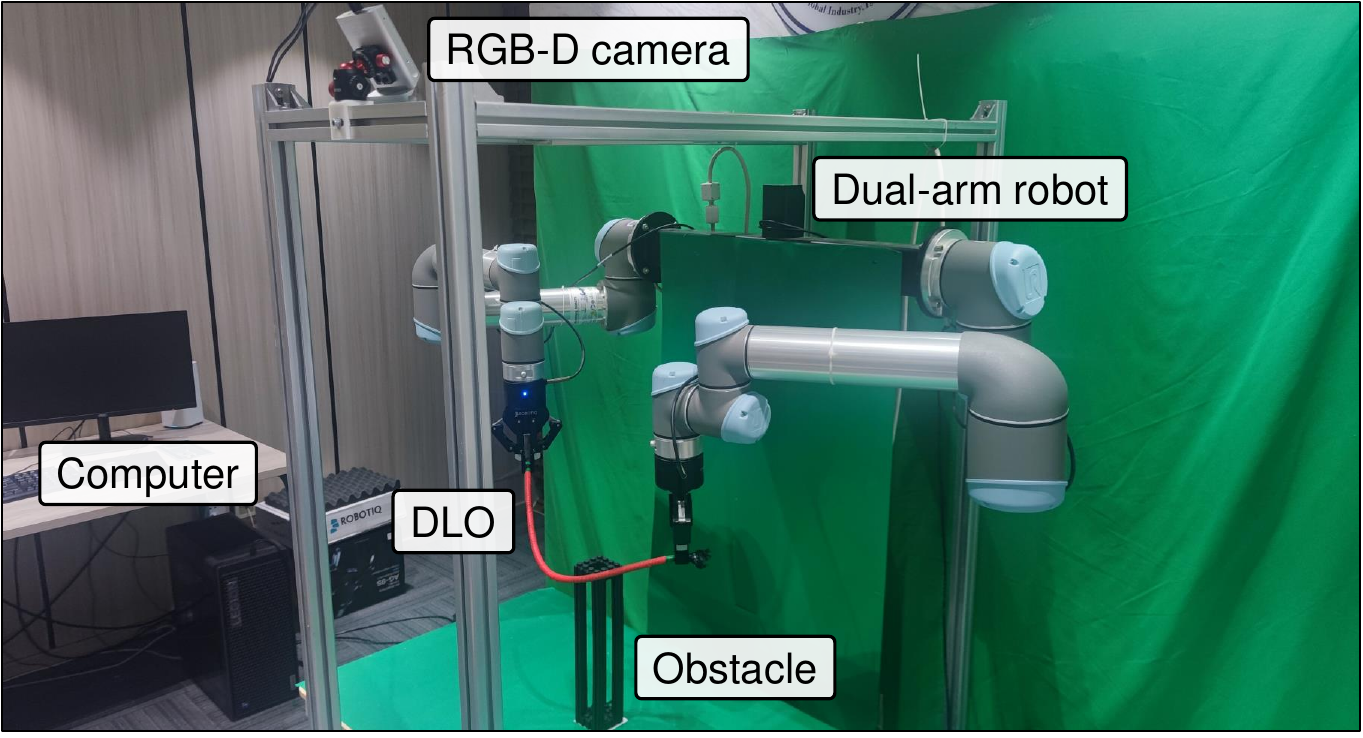} 
  \caption{Setup of the real-world experiments. The DLO is rigidly grasped by two grippers installed on the dual-arm robot. The top-view RGB-D camera is used to detect the DLO.}
  \label{fig:real_experimental_setup}
\end{figure}

\begin{figure*} [bp]
  \centering 
    \includegraphics[width=\textwidth]{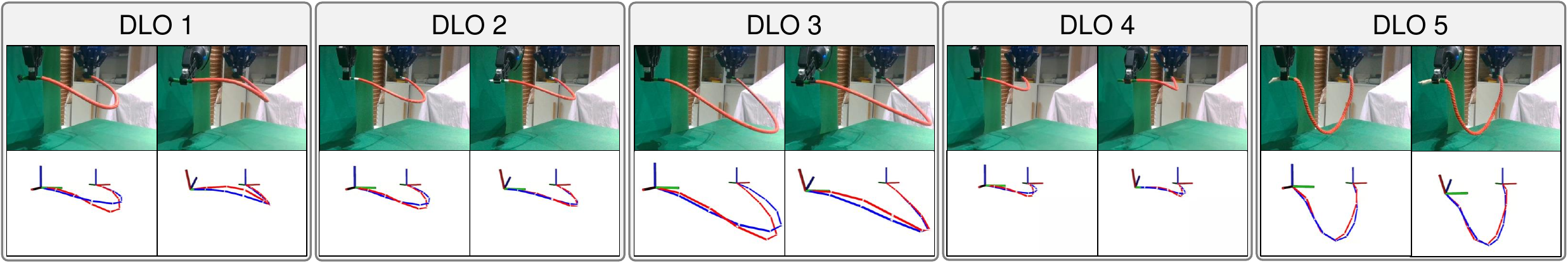} 
  \caption{Five DLOs used in the real-world experiments. For each DLO, the photos in the first row show the start and end configurations of the designed trajectory for the DER model identification. The figures in the second row are the corresponding visualizations in the Rviz, where the red lines indicate the actual observed DLO configuration and the blue lines indicate the projected configuration by (\ref{eq:dlo_projection}) using the identified DER model parameters.}
  \label{fig:real_dlos}
\end{figure*}

\begin{table*}[bp]
\centering
\caption{Parameters of the five DLOs used in the real-world experiments.}
\label{tab:real_DLO_params}
\begin{threeparttable}[b]
\begin{tabular}{c|cccc|cc} 
\toprule
\multirow{2}{*}{DLO} & \multirow{2}{*}{Type} & \multirow{2}{*}{Length~} & \multirow{2}{*}{Diameter} & \multirow{2}{*}{Stiffness\tnote{a}} & \multicolumn{2}{c}{Identified DER parameters} \\ 
\cline{6-7}
 &  &  &  &  & Relative bend stiffness\tnote{b} & Relative twist stiffness\tnote{c} \\ 
\hline
1 & TPU elastic & 0.49 m & 10 mm & $\bigstar\bigstar\bigstar\bigstar$ & 0.089 $\sim$ 0.10 & 0.36 $\sim$ 0.64 \\
2 & Electric wire & 0.46 m & 7 mm & $\bigstar\bigstar\bigstar$ & 0.059 $\sim$ 0.071 & 0.093 $\sim$ 0.12 \\
3 & HDMI cable & 0.72 m & 9 mm & $\bigstar\bigstar\bigstar\bigstar$ & 0.14 $\sim$ 0.16 & 0.35 $\sim$ 0.47 \\
4 & Nylon rope & 0.32 m & 8 mm & $\bigstar\bigstar$ & 0.016 $\sim$ 0.019 & 0.046 $\sim$ 0.072 \\
5 & Hemp rope & 0.54 m & 11 mm & $\bigstar$ & 0.0060 $\sim$ 0.071 & $< 0.0001$ \\
\bottomrule
\end{tabular}
\begin{tablenotes}
     \item[a] Qualitative stiffness estimated by humans.
     \item[b, c] Relative bend stiffness = bend stiffness / density; relative twist stiffness = twist stiffness / density.
\end{tablenotes}
\end{threeparttable}
\end{table*}

\subsection{Real-world experiments}

We conduct a series of real-world experiments involving various DLOs to demonstrate the applicability, generalizability, and robustness of the proposed method in the real world.

\subsubsection{Experimental setup}

As shown in Fig. \ref{fig:real_experimental_setup}, in the experiments, the two ends of the DLOs are rigidly grasped by two Robotiq 2F-85 grippers installed on a dual-UR5 robot. 
The DLOs are observed using a top-view RGB-D camera (Azure Kinect DK).
The positions of the DLO feature points $\bm x$ are estimated using a marker-free DLO detection method \citep{kichi2023dloftb}, and the DLO end poses $\bm M^0, \bm M^m$ are obtained by the robot forward kinematics. 
We segment DLOs from images by color, and also apply the tricks in \citep{bidzinksi2023toward} to get better estimations of the DLO segments near the grippers. 
The positions and geometries of obstacles are manually measured in advance.
Communication between the devices is based on the ROS. 
All algorithms are run on a desktop with an Intel i9-13900K CPU and a 32-GB RAM.

\begin{figure*} [tb] 
  \centering 
    \includegraphics[width=\textwidth]{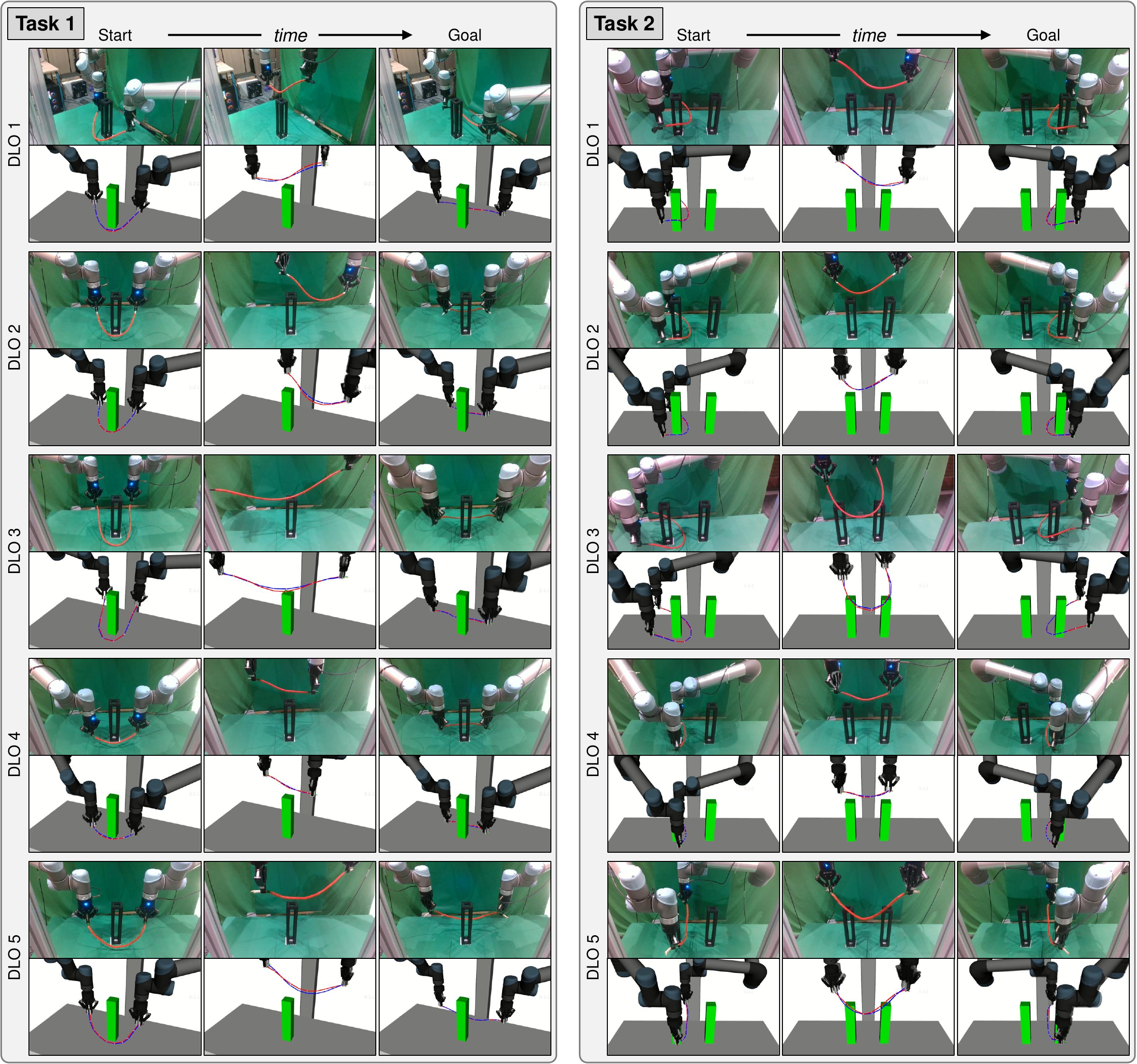} 
  \caption{Tasks 1 and 2 in the real-world experiments using the five DLOs, and corresponding manipulation processes using the proposed method. For each DLO in each task, the start and goal configurations are shown in the pictures in the first and last column, respectively. The pictures in the first row are captured by another camera (not used for tracking DLOs). The pictures in the second row are visualizations by the Rviz, in which the blue lines and translucent robots indicate the planned waypoints, and the red lines and non-translucent robots indicate the real-time configurations.}
  \label{fig:real_task_1_2}
\end{figure*}

\subsubsection{Five used DLOs and DER model identification}

To validate the generalizability of the proposed method, we use five different DLOs with lengths ranging from 0.32 m to 0.72 m, diameters ranging from 7 mm to 11 mm, and material stiffnesses ranging from stiff (TPU elastic) to soft (hemp rope), as shown in Fig. \ref{fig:real_dlos}. The parameters are listed in Table \ref{tab:real_DLO_params}. None of these DLOs are strictly naturally straight.

Before manipulation, the DER parameters of the DLOs are coarsely identified. Fig. \ref{fig:real_dlos} shows the configurations of DLOs in the designed trajectory for identification as well as the identification results. 
It can be seen that 
1) the DLOs exhibit different shapes under the effect of gravity and twist, reflecting the different properties of these DLOs and the effectiveness of the designed trajectory;
2) the DER models after identification well approximate the real DLOs with acceptable errors.  However, errors are inevitable because of the simplifications made to improve planning efficiency, such as ignoring anisotropic or naturally nonstraight properties.

In the following experiments of three tasks, we carry out the identification process once before each task. 
Table \ref{tab:real_DLO_params} lists the minimum and maximum of the DER model parameters among the three identifications. First, the results indicate that the identification can effectively distinguish the different DLO properties, as the ranking of the identified DLO stiffnesses is consistent with that estimated by humans.
Second, the identified parameters exhibit slight variations across different tests (< 2 times) owing to sensing noises and potential plastic deformation after long-time manipulations.
However, the following experiments demonstrate the robustness of our closed-loop manipulation framework to these identification errors.

\begin{figure*} [bp]
  \centering 
    \includegraphics[width=\textwidth]{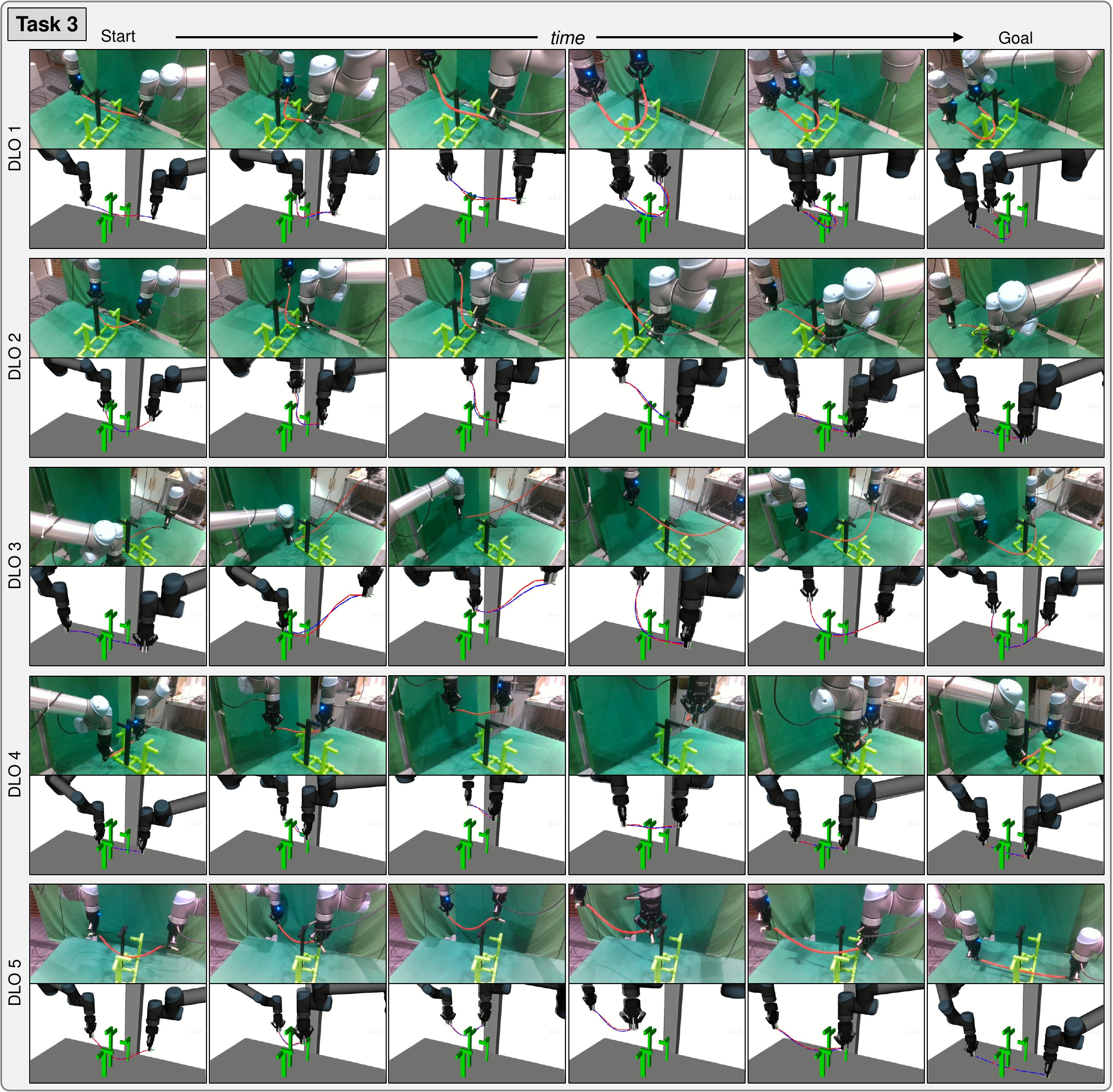} 
  \caption{Task 3 in the real-world experiments using the five DLOs, and corresponding manipulation processes using the proposed method. For each DLO, the start and goal configurations are shown in the pictures in the first and last column, respectively. The pictures in the first row are captured by another camera (not used for tracking DLOs). The pictures in the second row are visualizations by the Rviz, in which the blue lines and translucent robots indicate the planned waypoints, and the red lines and non-translucent robots indicate the real-time configurations.}
  \label{fig:real_task_3}
\end{figure*}

\subsubsection{Three designed tasks for evaluation}

We design three 3-D tasks for evaluation, as shown in Figs. \ref{fig:real_task_1_2} and \ref{fig:real_task_3}. 
In Task 1, the robot needs to manipulate the DLO from a top-view lower-semicircle shape to an upper-semicircle shape while bypassing a cuboid obstacle.
In Task 2, the robot needs to rotate the DLO by 180 degrees while avoiding two cuboid obstacles. Note that the start and goal configurations of different DLOs are slightly different based on the DLO properties.
Task 3 is much more challenging, where the robot must manipulate the DLO through narrow passages between complicated obstacles while reshaping the DLO. Such a complex task has not been attempted in real-world scenarios in previous studies. Using the same obstacles, we design different start and goal configurations for different DLOs. For DLO 1, the start is a straight line, and the goal is a right semicircle. For DLO 2, the start is a lower semicircle, and the goal is a upper semicircle. For DLO 3, the start is a upper semicircle, and the goal is a lower semicircle. 
For DLO 4, both the start and goal are a straight lines. For DLO 5, the start is a sagging shape and the goal is a straight line.

\begin{table*}[bp]
\centering
\setlength{\extrarowheight}{0pt}
\addtolength{\extrarowheight}{\aboverulesep}
\addtolength{\extrarowheight}{\belowrulesep}
\setlength{\aboverulesep}{0pt}
\setlength{\belowrulesep}{0pt}
\caption{Performance of the proposed manipulation method in the real-world experiments.}
\label{tab:real_manipulation_overall_performance}
\begin{tabular}{c|c|c|cccc} 
\toprule
Task & \begin{tabular}[c]{@{}c@{}}Planning time (s)\end{tabular} & \begin{tabular}[c]{@{}c@{}}Manipulation \\mode\end{tabular} & Success rate & \begin{tabular}[c]{@{}c@{}}Final task \\error (mm)\end{tabular} & Collision time (s) & Execution time (s) \\ 
\hline
\multirow{2}{*}{1} & \multirow{2}{*}{0.90 $\pm$~0.83} & Open-loop & 43/45 & 16.15 $\pm$ 9.68 & 0.51 $\pm$ 0.78 & \multirow{2}{*}{41.31 $\pm$ 2.85} \\
 &  & {\cellcolor[rgb]{0.902,0.902,0.902}}Closed-loop & {\cellcolor[rgb]{0.902,0.902,0.902}}45/45 & {\cellcolor[rgb]{0.902,0.902,0.902}}9.44 $\pm$ 5.03 & {\cellcolor[rgb]{0.902,0.902,0.902}}0.0 $\pm$ 0.0 &  \\ 
\hline
\multirow{2}{*}{2} & \multirow{2}{*}{0.83 $\pm$ 0.46} & Open-loop & 45/45 & 21.77 $\pm$ 10.06 & 1.58 $\pm$ 4.18 & \multirow{2}{*}{42.79 $\pm$ 4.41} \\
 &  & {\cellcolor[rgb]{0.902,0.902,0.902}}Closed-loop & {\cellcolor[rgb]{0.902,0.902,0.902}}45/45 & {\cellcolor[rgb]{0.902,0.902,0.902}}10.38 $\pm$ 4.34 & {\cellcolor[rgb]{0.902,0.902,0.902}}0.0 $\pm$ 0.0 &  \\ 
\hline
\multirow{2}{*}{3} & \multirow{2}{*}{3.82 $\pm$ 3.04} & Open-loop & 39/45 & 14.46 $\pm$ 8.07 & 2.22 $\pm$ 2.31 & \multirow{2}{*}{53.12 $\pm$ 6.07} \\
 &  & {\cellcolor[rgb]{0.902,0.902,0.902}}Closed-loop & {\cellcolor[rgb]{0.902,0.902,0.902}}45/45 & {\cellcolor[rgb]{0.902,0.902,0.902}}9.52 $\pm$ 4.45 & {\cellcolor[rgb]{0.902,0.902,0.902}}0.004 $\pm$ 0.029 &  \\
\bottomrule
\end{tabular}
\end{table*}

\begin{figure*} [tb]
  \centering 
    \includegraphics[width=\textwidth]{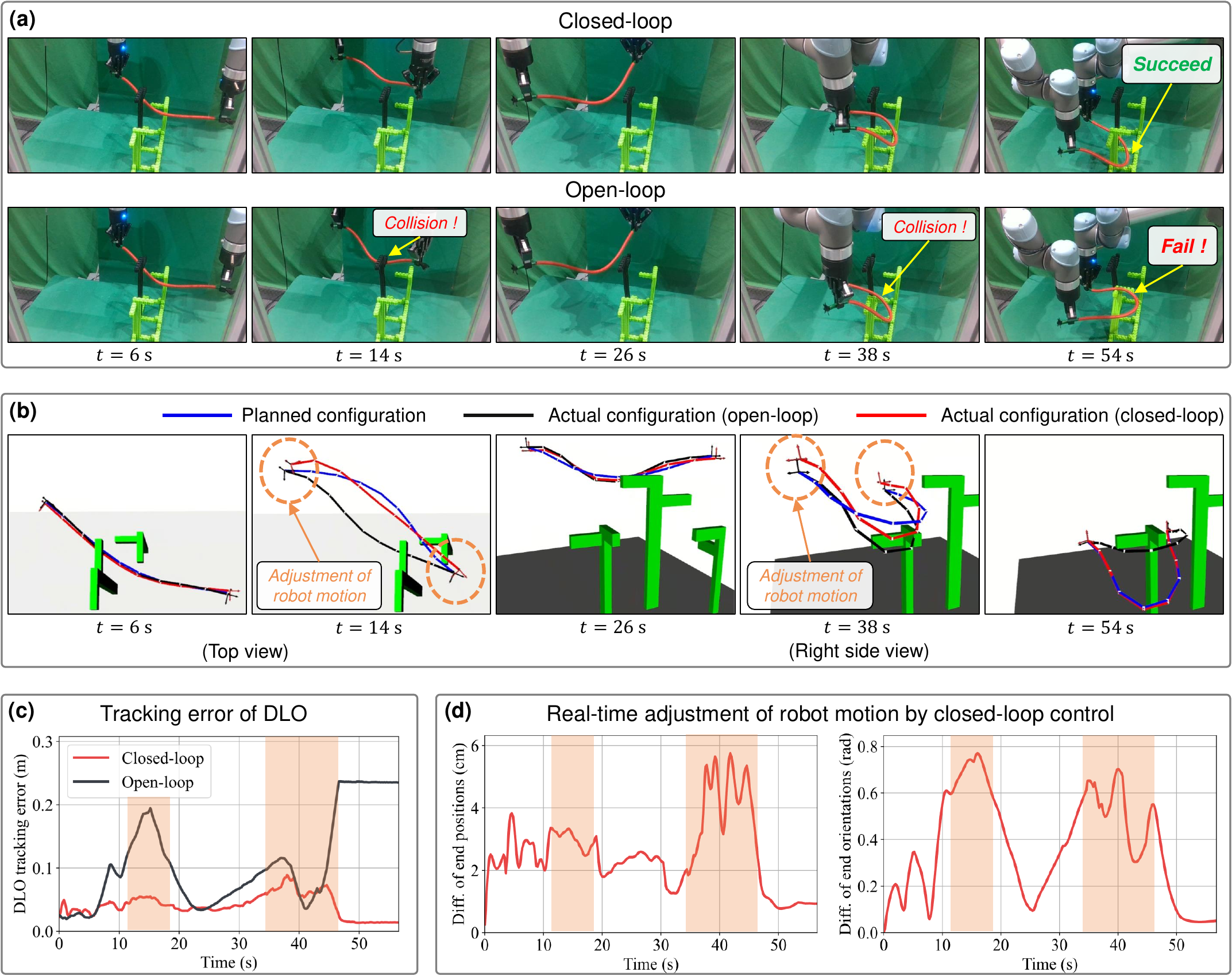} 
  \caption{A case for illustrating the comparison between the open-loop and proposed closed-loop manner using the same planned paths. 
  (a) Snapshots of the manipulation processes. 
  (b) Visualization of the actual DLO configurations during the open-loop and closed-loop manipulations and the corresponding planned DLO configurations, where the lines indicate the DLO centerlines, and the axes indicate the robot gripper poses. 
  (c) Tracking error of DLO (i.e., $\tilde{\bm x}$ in (\ref{eq:dlo_tracking_error})) during the open-loop and closed-loop manipulations.
  (d) Differences between the robot gripper poses during the open-loop and closed-loop manipulations, indicating the real-time adjustment of the robot motion by the closed-loop control.
  The orange circles in (b) and boxes in (c)(d) are for highlighting the comparison.
  }
  \label{fig:real_closed_open_compare}
\end{figure*}

\subsubsection{Manipulation performance}

For each DLO in each task, we plan three paths using different random seeds and execute each path three times using the closed-loop and open-loop manner, respectively. The results are summarized in Table \ref{tab:real_manipulation_overall_performance}, which indicate that 
1) the planning is efficient, as the average time for finding a feasible path is only 3.82 s in the most challenging Task 3;
2) the proposed closed-loop manipulation framework is robust, as the manipulation success rate of the closed-loop manner is 135/135, while that of the open-loop manner is 127/135;
3) the closed-loop manner improves the final task precision (reducing the average error over the three tasks from 17.46 to 9.78 mm), as the real DLOs exhibit elastoplastic deformation during manipulations and may not reach exactly the same configuration between different open-loop executions;
4) the closed-loop manner effectively avoids unexpected collisions, as collision occurs only once (0.2 s) during all closed-loop manipulations, while the average collision time of open-loop manipulations in Task 3 is 2.22 s.
Additionally, no replanning is invoked in any of the manipulations. The average execution time of Task 3 is 53.12 s.

For better illustration, in Fig. \ref{fig:real_closed_open_compare} we visualize a manipulation case where the open-loop scheme fails while the closed-loop scheme succeeds, indicating how the local controller compensates for the planning errors. The snapshots in Fig. \ref{fig:real_closed_open_compare}(a) shows that during the open-loop manipulation, the DLO collides with the obstacles twice (when $t \approx 14 ~ \text{s}$ and $t \approx 38 ~ \text{s}$) and ultimately fails to reach the goal configuration. The visualization in Fig. \ref{fig:real_closed_open_compare}(b) and measured DLO tracking error in Fig. \ref{fig:real_closed_open_compare}(c) show that those collisions occur because the DLO does not move as planned when the planned robot path is open-loop executed. 
We observe that the planning errors are significant in two situations. 
The first situation is when the DLO must be shaped through a singular configuration (i.e., a straight-line shape) to another side, e.g., from $t \approx 6 ~\text{s}$ to 14 s. It requires very high precision or real-time feedback. 
The second situation is when the DLO is naturally nonstraight or exhibits elastoplastic deformation, which is common for real DLOs. We observe that the DLOs usually have plastic deformation along the direction of gravity after long-time manipulations, so the DLO internal body is lower than planned, e.g., at $t \approx 38$ s. 
To compensate for the planning errors, the controller adjusts the robot motion according to the real-time feedback, which is clearly shown in Figs. \ref{fig:real_closed_open_compare}(b) and \ref{fig:real_closed_open_compare}(d). At $t \approx 14$ s and $t \approx 38$ s, the controller significantly adjusts the robot gripper poses to track the planned DLO path more closely and avoid the obstacles.

These experimental results demonstrate the effectiveness and necessity of the closed-loop manipulation framework in the real world where the properties of DLOs are much more complex and difficult to accurately model.

\begin{figure*} [tb]
  \centering 
    \includegraphics[width=\textwidth]{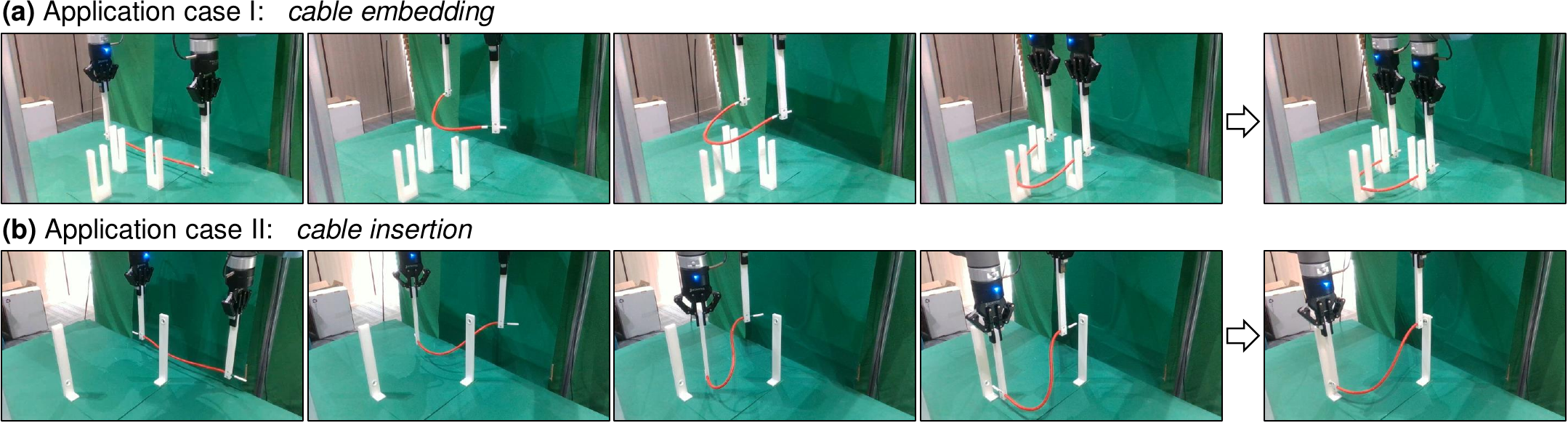} 
  \caption{Two cases of potential applications: (a) moving and embedding the DLO into the grooves; (b) moving and inserting the DLO ends into the holes. We use the 3-D printed long grippers to improve the robot's dexterity in narrow spaces.
  }
  \label{fig:real_application_cases}
\end{figure*}

\subsubsection{Cases of potential applications}

We show two cases of potential applications of the proposed method in Fig. \ref{fig:real_application_cases}. 
We additionally use two 3-D printed long grippers to improve the dexterity of the robot in narrow spaces. 
In each case, the whole manipulation is planned and executed by our method in a single complete path (figures in the first four columns), except for the final step (figure in the last column) which includes interactions with the environment.

The first case is moving and embedding the DLO into the grooves. In this case, the DLO should be first moved above the grooves while avoiding obstacles, and then accurately embedded into the narrow and deep grooves, during which the internal shapes of the DLO are more important. 
The second case is moving and inserting the DLO ends into the holes. In this case, the DLO should be accurately moved to the pre-installation configuration while avoiding obstacles, during which the poses of the DLO ends are more important.
In general, the proposed method can be applied to a wide range of applications that involve moving DLOs in constrained environments with various obstacles.

\section{Conclusion and discussion}

\subsection{Conclusion}

\textbf{Overall}: This article proposes a complementary framework that integrates whole-body planning and control for global collision-free manipulation of DLOs in constrained environments by a dual-arm robot.
The proposed framework efficiently and accurately accomplishes this high-dimensional and highly constrained task, and it also effectively generalizes to various DLOs, even when accurate DLO models are unavailable.
The core of the framework is the combination of global planning and local control: the global planner efficiently finds a feasible path based on a coarse DLO model; then, the local controller tracks this guiding path while adjusting the robot motion according to the real-time feedback to compensate for planning errors. 
Unlike existing works, we consider the full configurations of both the DLO and robot bodies and all essential constraints in the whole closed-loop manipulation process from planning to control.

\textbf{Details}: The framework uses a discrete representation of DLO configurations. Based on the quasi-static and elastic deformation assumptions, we apply two DLO models: the simplified DER model for planning and the DLO Jacobian model for control. 
We use an RRT framework for planning, in which the DER model is used for projecting DLO configurations to satisfy the stable DLO constraint, and the arm Jacobian-based projection method is used to satisfy the closed-chain constraint.
Furthermore, we propose a new DLO interpolation method for steering, and consider the under-actuation of the system by using the DER model as a kinodynamic model. Additionally, we design a simple DER model parameter identification method to get better planning results for various DLOs.
In the closed-loop manipulation framework, we design a nonlinear MPC to tracking the planned guiding path, in which we employ the DLO Jacobian model for state prediction and incorporate hard constraints for avoiding collisions and overstretch. We additionally introduce a replanning strategy to handle extreme situations.
Note that the proposed framework can also be easily applied in single-arm or 2-D manipulation scenarios.

\textbf{Performance}: We conduct simulations and real-world experiments to demonstrate that our method can efficiently, robustly, and accurately accomplish tasks that the existing works cannot realize. 
Even in the most complicated task, our planner achieves a 100\%  planning success rate and an average planning time cost of less than 15 s. Our closed-loop manipulation manner achieves a 100\% manipulation success rate, almost no collision, and an average execution time of less than 65 s. 
Our approach successfully accomplishes all 135 tests in the three real-world tasks involving five DLOs with different properties.

\subsection{Discussion}

\textbf{About problem assumptions}: 
1) Our problem formulation assumes quasi-static manipulation and elastic DLO deformation. 
Consequently, our method is designed for slow manipulation of elastic DLOs without dynamic effects. 
However, these assumptions may not be satisfied in some situations. 
For example, the DLO may quickly transition to another shape with much lower energy if the current energy becomes too high. To address this, we introduce the replanning strategy. 
In practice, we find that such situations rarely occur.
2) Additionally, the real DLOs are typically not ideally elastic but elastoplastic, i.e., involving some plastic deformation after manipulation. Our closed-loop manipulation framework is designed to compensate for planning errors, including accommodating such changes, and the experimental results show the robustness of our approach in handling such deviations.

\textbf{About collisions}: 
Our problem formulation requires collision-free manipulation. We enforce this condition because 
1) collisions may damage the objects in the environment (such as knocking over a cup of water), even though the manipulation is not obstructed; and
2) the effect of collisions on the movement of DLOs is difficult to predict, so avoiding collisions will improve the accuracy and robustness. 
Our approach is designed for general moving and shaping of DLOs. However, certain applications may require interactions between the DLO and environment, such as cable assembly. In such cases, a possible pipeline is to first use our general approach to move the DLO to a configuration near the assembly position, and then use task-specific methods to accomplish the final step. 
How to realize general DLO manipulation with environmental contacts is a topic worthy of future research.

\textbf{Potential future improvements}: 
{\spaceskip=0.18em\relax 
Several aspects of our approach may be further explored and improved in the future:
}
\begin{enumerate}
    \item How to achieve optimal global planning, such as minimizing the deformation of DLOs during manipulation? We have tried using the RRT* framework, but it is too inefficient for this high-dimensional problem.
    \item How to reduce the time cost of DLO stable configuration projection? We find that it is the most time-consuming step in our planning process.
    \item How to identify the DER models for anisotropic or naturally nonstraight DLOs without sacrificing the efficiency? Better planning results with less simplifications are always helpful.
    \item How to balance the tracking of planned DLO paths and robot paths? We currently use a fixed-weighted sum of the two tracking errors in the MPC cost function but get good enough performances. 
    It is possible to design an adaptive weight adjustment strategy.
    \item How to improve the solving efficiency of the high-dimensional MPC that includes collision constraints for both DLOs and robots?
    \item How to continually adjust the planned path in real time according to the actual states and then provide the controller with more accurate tracking objectives? 
\end{enumerate}

\textbf{Failure cases}:
Most of the failure cases of our approach is due to perception:
1) when most part of the DLO is occluded by obstacles or arm bodies, the perception algorithm fails; 
2) the depth values estimated by the Azure Kinect DK are much more noisy when the DLO is close to the background; 
and 3) there are some spatial alignment errors between the color map and depth map of the Azure Kinect DK, which is a widely reported problem.
To deal with them in the real-world experiments, we carefully design the tasks and restrict the moving range of the DLOs and arms to avoid large occlusions. Moreover, we assign a larger $\beta_a$ (weight of the cost of changes in robot joint velocities) and a longer horizon $T$ of the MPC for smoother control to compensate for the perception noises. 
After these, our approach achieves a 100\% manipulation success rate in the three real-world tasks.

In addition, the ability of our local controller to compensate for planning errors is not unlimited. Our approach may fail if the planning errors are excessively large as well as the environment is highly constrained, e.g., manipulating the stiff HDMI cable in the complicated real-world Task 3 based on the planned path for the soft hemp rope. This highlights the necessity of the coarse identification before planning.


\section{Declaration of conflicting interests}

The author(s) declared no potential conflicts of interest with respect to the research, authorship, or publication of this article.

\section{Funding}

This work was supported in part by the Science and Technology Innovation 2030-Key Project under Grant 2021ZD0201404, in part by National Natural Science Foundation of China under Grant 623B2059, U21A20517, and 52075290, and in part by the Institute for Guo Qiang, Tsinghua University.



\bibliographystyle{SageH} 
\bibliography{ref}

\section{Appendix A. Index to multimedia extensions}

The supplementary video is available on the project website \url{https://mingrui-yu.github.io/DLO_planning_2}

\section{Appendix B. Related work about path planning and control for robot arms}

We outline the popular methods about planning and control for robot arms that is closely related to the proposed approach for DLO manipulation in this article.

\textbf{Path planning}: Global planning for robot arms is a high-dimensional planning problem with a complex configuration space. For such problems, sampling-based algorithms are more popular since they are efficient and easy to implement. The multi-query PRMs \citep{amato1996randomized} and single-query RRTs \citep{lavalle1998rapidly} are the most frequently used sampling-based methods, and many of their variants have been developed and widely applied. The bi-directional RRT separately grows two search trees from the start and goal, thereby greatly improving the efficiency \citep{kuffner2000rrtconnect}. 
In cases where the planning goal is specified only in the task space, i.e., the goal pose of the end-effector, some approaches have incorporated greedy task-space search into RRTs without explicit inverse kinematics (IK) \citep{bertram2006anintergrated, vande2007randomized}. Sampling multiple IKs for the goal poses have also been shown to be effective \citep{vahrenkamp2009humanoid}, which allows the usage of more efficient bi-directional RRTs. 
Constraints may also be involved into planning, such as pose constraints for the end-effector. \citet{yao2007path} proposed an algorithm where exploration was guided entirely by task-space samples, and local movements of the end-effector were constrained. To guarantee the probabilistic completeness, constrained planning algorithms in the joint space have been studied \citep{stilman2007task, stilman2010global, berenson2009manipulation, berenson2011task}, where the pose constraints are satisfied by projecting an invalid configuration onto the constraint manifold using the arm Jacobian.
Closed-chain kinematic constraints are formed when dual arms manipulate an object. Several strategies have been designed specifically for sampling configurations that satisfy these closed-chain constraints \citep{cortes2005sampling}, while the general pose-constrained planning algorithms can also handle these tasks \citep{berenson2011task}.

\textbf{Control}: Motion control of manipulators with real-time obstacle avoidance is another topic closely related to this work. 
Most of the relevant methods are based on the artificial potentials \citep{khatib1986real} to calculate repulsive forces. In order to reduce the computation for real-time calculation, these methods typically consider only the current state and simplify the collision shape of the arm to a series of points \citep{khatib1986real, zhang2020realtime}, line segments \citep{lacevic2013safety}, spheres \citep{balan2006realtime}, or the temporarily closest point to obstacles \citep{maciejewski1985obstacle, zhang2021obstacle}.
Recently, \citet{bhardwaj2022storm} explored the use of sampling-based MPCs with a data-driven collision checking module approximated by neural networks, relying on the parallel processing capabilities of high-performance GPUs.

\section{Appendix C. More about planning}
\subsection{Distance metrics}

We define the $\mathcal{L}_{\infty}$ distance between the positions of two DLO configurations as
\begin{equation}
    \text{dist}^{\rm p}_{\infty}(\bm \Gamma_{A}, \bm \Gamma_{B}) = \max_k \| \bm x_{A, k} - \bm x_{B, k} \|_2 ,
\end{equation}
and the $\mathcal{L}_{\infty}$ distance between the orientations of two DLO configurations as 
\begin{equation}
    \text{dist}^{\rm o}_{\infty}(\bm \Gamma_{A}, \bm \Gamma_{B}) = \max_k \left( \text{angle}(\bm M^k_A, \bm M^k_B) \right) .
\end{equation}
We define the $\mathcal{L}_{\infty}$ distance between two robot configurations in the configuration space as
\begin{equation}
    \text{dist}^{\rm c}_{\infty}(\bm q_{A}, \bm q_{B}) =  \max_{i} | q_{A, i} - q_{B, i} | ,
\end{equation}
and the $\mathcal{L}_{\infty}$ distance between two robot configurations in the workspace as
\begin{equation}
    \text{dist}^{\rm w}_{\infty}(\bm q_{A}, \bm q_{B}) = \max_j \|\bm \xi_{j}(\bm q_{A}) - \bm \xi_{j}(\bm q_{B}) \|_2 ,
\end{equation}
where $\bm \xi_{j}$ is the forward kinematics of the $j^{\rm th}$ dual-arm link.

As an RRT node contains both the DLO and robot configurations, we define the scalar distance between $\mathcal{N}_{A}$ and $\mathcal{N}_{B}$ as
\begin{equation} \label{eq:dist_between_nodes}
\begin{aligned}
    & \text{dist}(\mathcal{N}_{A}, \mathcal{N}_{B}) 
    = 
    \\
    & \quad \max 
    \left\{ 
    \text{dist}^{\rm p}_{\infty}(\mathcal{N}_{A}.{\bm \Gamma}, \mathcal{N}_{B}.{\bm \Gamma}),
    \text{dist}^{\rm w}_{\infty}(\mathcal{N}_{A}.{\bm q}, \mathcal{N}_{B}.{\bm q})
    \right\}  .
\end{aligned}
\end{equation}
This definition enables a unified combination of the distance metrics for the DLO and robot, which is used in the nearest node search.
In addition, we define a vectorial distance between $\mathcal{N}_{A}$ and $\mathcal{N}_{B}$ as 
\begin{equation} \label{eq:vectorial_dist_between_nodes}
    \text{Dist}(\mathcal{N}_{A}, \mathcal{N}_{B}) 
    =
    \left[ \begin{array}{c}
         \text{dist}^{\rm p}_{\infty}(\mathcal{N}_{A}.{\bm \Gamma}, \mathcal{N}_{B}.{\bm \Gamma}) \\
         \text{dist}^{\rm o}_{\infty}(\mathcal{N}_{A}.{\bm \Gamma}, \mathcal{N}_{B}.{\bm \Gamma}) \\
         \text{dist}^{\rm c}_{\infty}(\mathcal{N}_{A}.{\bm q}, \mathcal{N}_{B}.{\bm q})
    \end{array} \right] .
\end{equation}

\subsection{Unspecified goal robot configuration}
In some applications, only the goal DLO configuration is specified, while the corresponding goal robot configuration is unknown. In order to use bi-directional RRT to accelerate the planning, we use a goal sampling strategy, inspired by \citet{vahrenkamp2009humanoid} and \citet{berenson2011task}.
Deterministically choosing one of the robot IK solutions as the goal robot configuration is inappropriate, because it may be unreachable owing to the obstacles or rigidly grasped DLO. 
Instead, we randomly sample multiple IK solutions and then add multiple goal nodes to the goal tree $\mathcal{T}_{\rm goal}$. If any of the goal nodes can be reached from the start node, the planning succeeds. 
In addition, we continue to sample a new goal node in each RRT iteration with some probability $p_{\rm sg}$. We denote the process of adding new root nodes to a tree $\mathcal{T}$ as $\text{AddRoot}(\mathcal{T}, \bm \Gamma, n_{\rm sample})$, where $\bm \Gamma$ is the root DLO configuration and $n_{\rm sample}$ is the maximum number of samples (see Algorithm \ref{algorithm:root_sampling}). Note that the $\text{AddRoot}()$ can also be applied for the start configuration.

\begin{algorithm}[tb] 
\caption{: $\text{AddRoot}(\mathcal{T}, \bm \Gamma, n_{\rm sample})$}
\label{algorithm:root_sampling}
\begin{algorithmic}[1]
    \State $\bm p_l, \bm p_r \leftarrow \text{GetTwoDLOEndPoses}(\bm \Gamma)$;
   \For {$i=1$ \textbf{to} $n_{\rm sample}$}
        \State $\bm q_l \leftarrow \text{LeftArmRandomInverseKinematics}(\bm p_l)$;
        \State $\bm q_r \leftarrow \text{RightArmRandomInverseKinematics}(\bm p_r)$;
        \State $\mathcal{N}_{\rm sample} \leftarrow \{ \bm \Gamma, \bm q_l, \bm q_r \}$;  

        \Statex \quad \Comment{discarding previously sampled IKs}
        \If {$\min_{\mathcal{N} \in \mathcal{T}} \text{dist}(\mathcal{N}, \mathcal{N}_{\rm sample}) > \epsilon_{ar}$}
            \State $\mathcal{T}.\text{AddVertex}(\mathcal{N}_{\rm sample})$;
        \EndIf
   \EndFor
\end{algorithmic}
\end{algorithm}

\subsection{Assistant task-space guided exploration}

When sampling $\mathcal{N}_{\rm rand}$, a random robot IK solution corresponding to the sampled DLO configuration is assigned. Although this encourages exploration in the robot configuration space, the sampled robot configuration is likely to be far from the task requirements, providing ineffective guidance and slowing down the planning.
To address this problem, we introduce an exploring strategy which we call \textit{assistant task-space guided exploration}, where we define the \textit{task space} as the DLO configuration space.
The core concept is that, in some RRT iterations, we find the nearest node to the randomly sampled node based solely on the distance of DLO configurations, and then the robot is greedily moved to follow the exploration of the DLO.

To implement this strategy, when using the task-space guided exploration, we make the following adjustments to the random sampling, nearest node search, and extending: 
\begin{enumerate}
    \item When sampling a random node, we only sample a DLO configuration, which is denoted as $\mathcal{N}_{\rm rand} = \text{RandomSampleInTaskSpace}()$.
    \item When searching for the nearest node to $\mathcal{N}_{\rm rand}$, the distance metric considers only the DLO. That is: $\text{dist}(\mathcal{N}_{A}, \mathcal{N}_{B}) = \text{dist}^{\rm p}_{\infty}(\mathcal{N}_{A}.{\bm \Gamma}, \mathcal{N}_{B}.{\bm \Gamma})$.
    \item During the steering, no target robot configuration $\mathcal{N}_{\rm to}.{\bm q}$ is used. Unlike Line \ref{line:project_closed_chain_config} in Algorithm \ref{algorithm:steering}, the steered robot configuration is calculated using the last robot configuration instead of the interpolated robot configuration as the initial value for projection:  
    $\bm q_{\rm steer}   \hspace{-1mm}  \leftarrow   \hspace{-1mm} \text{ProjectClosedChainConfig} \left(\mathcal{N}_{\rm from}.{\bm q}, \bm \Gamma_{\rm erp} \right)$.
    
    \item During the extending, the distance function in Line \ref{line:judgment_for_extend} of Algorithm \ref{algorithm:extend} takes into account only the DLO.
\end{enumerate}

The difference between the task-space guided exploration and original full-space guided exploration is only in the exploration direction and approach. Every node in the trees still contains both the DLO and robot configurations and satisfies all constraints.

For the convenience of expression, in Algorithm \ref{algorithm:complete_planning} we use the same notation for the operations of the two exploration approaches, whose internal implementations are different according to whether $\mathcal{N}_{\rm rand}$ contains a robot configuration.
We use the task-space guided exploration in one RRT iteration with a probability $p_{\rm ts}$. Otherwise, we use the original full-space guided exploration. In such a framework, the task-space guided exploration enables greedy search which improves the planning efficiency, and the full-space guided exploration ensures the effective exploration and connection of robot configurations, which is especially necessary for redundant robots.

\section{Appendix D. Additional results}

\begin{figure} [tb]
  \centering 
    \includegraphics[width=\linewidth]{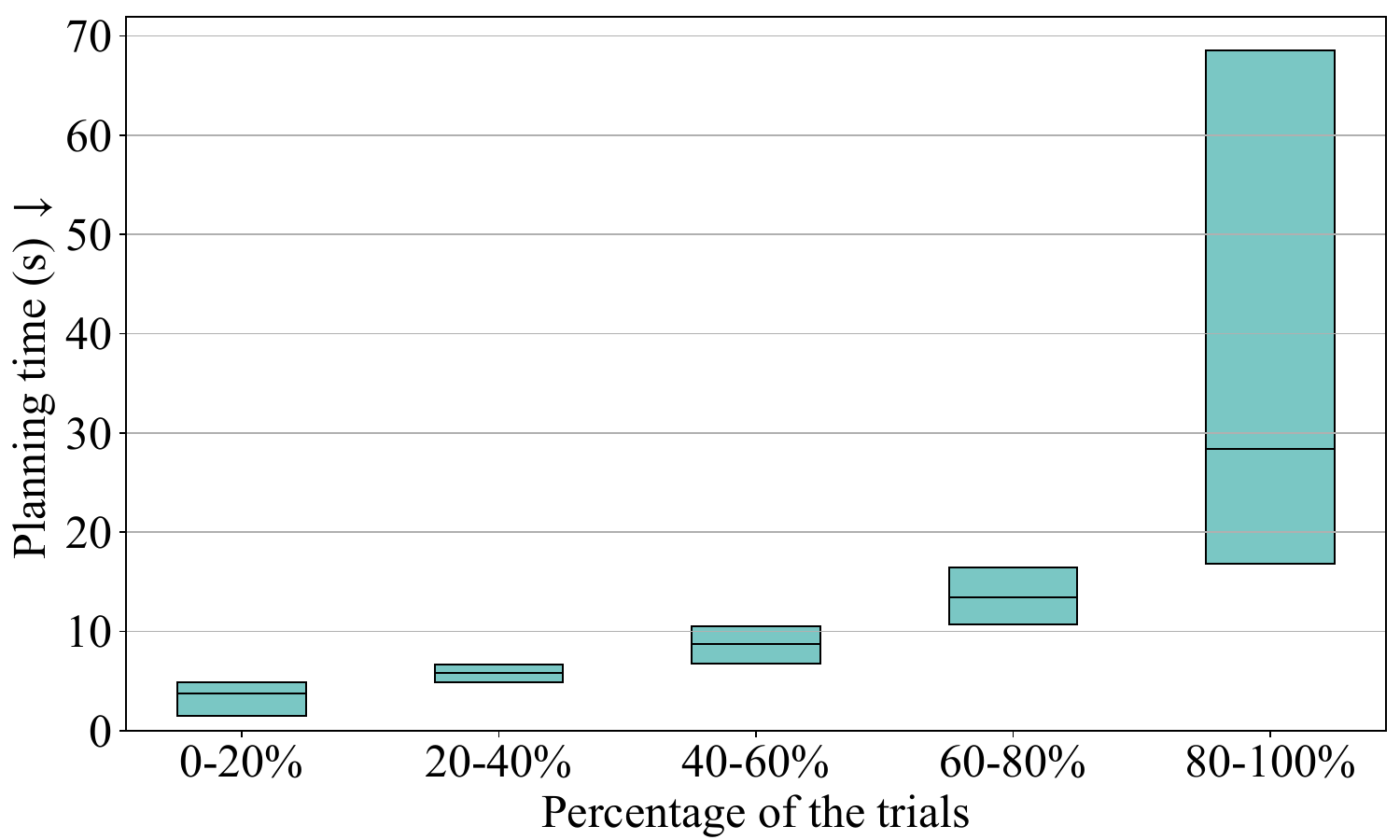}
  \caption{Distribution of the planning time for finding a feasible path among 100 planning trials in the simulated Task 4. The trials are sorted in increasing order of the time cost. The bottom, middle, and top line of each bar refers to the minimum, mean, and maximum value, respectively.}
  \label{fig:planning_time_distribution}
\end{figure}

\begin{table}
\centering
\caption{Time cost of key functions in the planning.}
\label{tab:time_cost_each_function}
\scalebox{0.9}{
\begin{tabular}{c|cc} 
\toprule
Function & \begin{tabular}[c]{@{}c@{}}Proportion of \\total planning time\end{tabular} & \begin{tabular}[c]{@{}c@{}}Time cost \\per call (ms)\end{tabular} \\ 
\hline
\small ProjectStableDLOConfig() & 93.3\% & 3.9 \\
\small ProjectClosedChainConfig() & 1.2\% & 0.051\\
\small CollisionCheck() & 3.4\% & 0.076 \\
\bottomrule
\end{tabular}
}
\end{table}

\textbf{Planning time distribution}: 
Figure \ref{fig:planning_time_distribution} shows the distribution of the time for finding a feasible path among 100 planning trials in the simulated Task 4. As the planning algorithm is based on random sampling, the time cost of each trial may be significantly different.
The results show that 40\% of the trials cost less than 6.7 s and 80\% of the trials cost less than 16.5 s. Some extreme cases may cost tens or even hundreds of seconds. In practice, an early stop and restart strategy can be used \citep{wedge2008heavy}.

We also report the time costs of the key functions in these 100 trials in Table \ref{tab:time_cost_each_function}, including the stable DLO constraint projection, closed-chain constraint projection, and collision check. The stable DLO constraint projection accounts for 93.3\% of the total planning time, with each projection costing 3.9 ms on average. Thus, it is worth studying how to constrain and predict DLO configurations more efficiently in future works.

\begin{figure} [tb]
  \centering 
    \includegraphics[width=\linewidth]{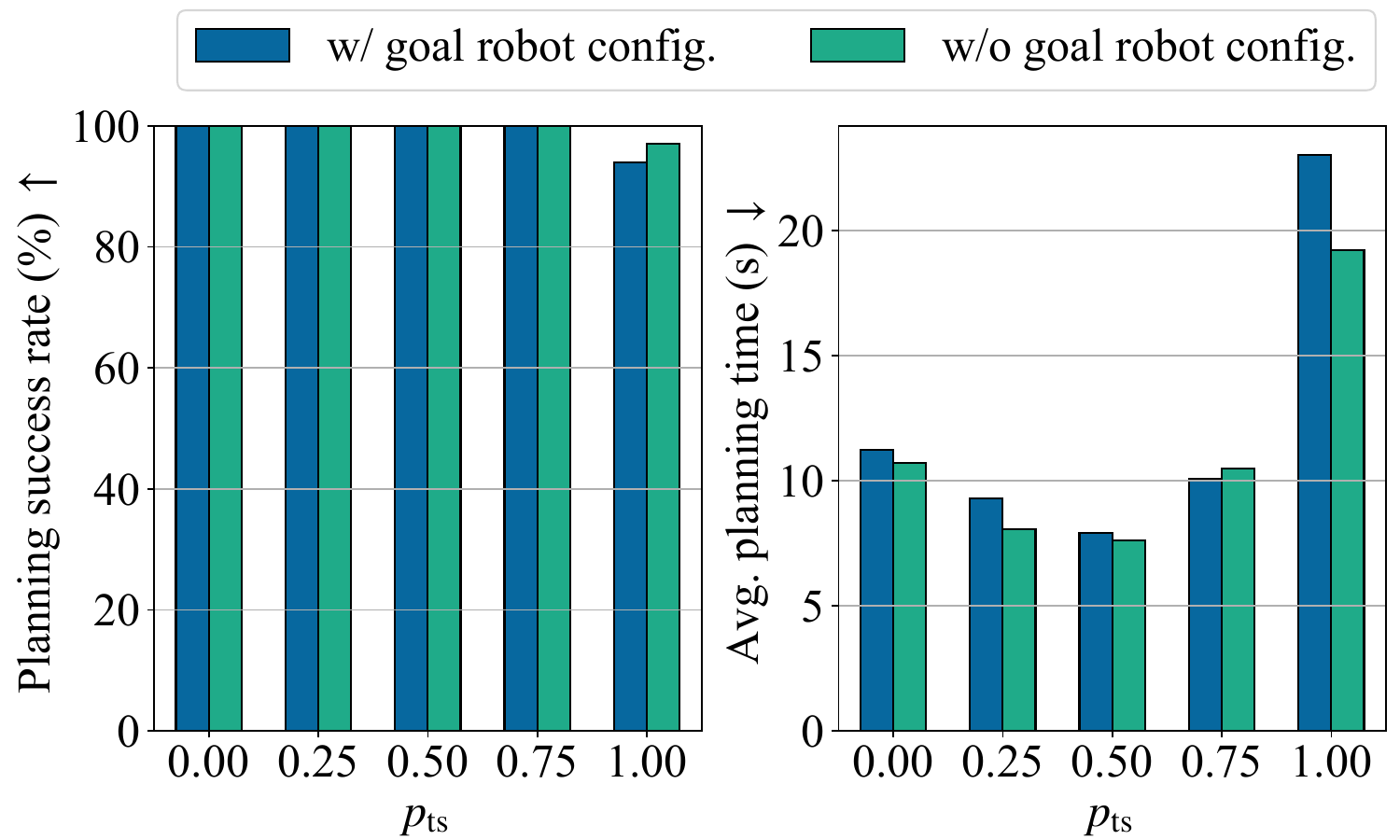} 
  \caption{Effect of the \textit{assistant task-space guided search} using different $p_{\rm ts}$ in the simulated Task 4. The average planning time refers to the average time for finding a feasible path.}
  \label{fig:sim_planner_ablation}
\end{figure}

\begin{table*}[bp]
\centering
\caption{Ablation study of the planner in the simulated Task 4.}
\label{tab:sim_planner_more_ablation}
\begin{tabular}{c|cccc} 
\toprule
Planner type & \begin{tabular}[c]{@{}c@{}}Planning\\success rate $\uparrow$~\end{tabular} & \begin{tabular}[c]{@{}c@{}}Planning \\time (s) $\downarrow$\end{tabular} & \begin{tabular}[c]{@{}c@{}}Open-loop 
 manipulation\\success rate $\uparrow$\end{tabular} & \begin{tabular}[c]{@{}c@{}}Closed-loop manipulation\\success rate $\uparrow$ \\$\|$ replanning times~$\downarrow$\end{tabular} \\ 
\hline
Ours & 100/100 & 8.22 $\pm$ 3.34 & 81/100 & 100/100 $\|$~1 \\
Not consider under-actuation & 100/100 & 5.91 $\pm$ 2.53 & 80/100 & 100/100 $\|$~1 \\
Not satisfy stable DLO constraint & 100/100 & 2.50 $\pm$ 0.97 & 55/100 & 97/100 $\|$~4 \\
\bottomrule
\end{tabular}
\vspace{-3mm}
\end{table*}

\textbf{Effect of the assistant task-space guided exploration}: 
Figure \ref{fig:sim_planner_ablation} shows the performance of the assistant task-space guided exploration.  
When the task-space exploration is applied with some probability, the planning efficiency improves (about 30\%) without reducing the robustness (100\% success rate).
We choose $p_{\rm ts} = 0.5$ for all other tests.

\textbf{Effect of the closed-chain constraint projection}:
In the steering step, to satisfy the closed-chain constraint, one straightforward way is to set the $\bm q_{\rm steer}$ as a pair of robot IK solutions of the two DLO end poses, e.g., the IK solutions closest to $\bm q_{\rm from}$. However, this strategy cannot achieve sufficient exploration in the robot configuration space. Our approach is similar to it when $p_{\rm ts} = 1.0$, in which no robot configuration is randomly sampled and no $\bm q_{\rm to}$ is used in the steering of the exploration tree. The results in Fig. \ref{fig:sim_planner_ablation} also indicate that this approach has lower planning success rate and higher time cost since the exploration may get stuck in local optima.
In contrast, the proposed steering approach based on robot configuration interpolation and closed-chain constraint projection encourages the exploration of robot configurations, making planning more effective.

\textbf{Effect of under-actuation in planning}:
We compare the results obtained with and without considering the under-actuation in planning, as shown in Line 1 and 2 in Table \ref{tab:sim_planner_more_ablation}. 
\textit{Not consider under-actuation} means that we directly use the interpolated DLO configuration as the initial value of ProjectStableDLOConfig() in the steering step instead of predicting the reached DLO configuration by ForwardPred(). 
The results show that 
1) the planner is more efficient when under-actuation is not considered, since the initial value used in ProjectStableDLOConfig() (i.e., $\bm \Gamma_{\rm erp}$) is closer to a stable configuration, leading to less time cost of per projection.
2) The manipulation results with and without consideration of under-actuation are comparable. 
We believe the reason is that, in quasi-static manipulation without contact, the stable DLO shapes in local regions can mainly be determined by the end poses since they remain in stable equilibriums, making the effect of under-actuation in stable steering not significant.
As a result, in practice, we may simplify the system to a constrained full-actuated system to further improve the planning efficiency.

\textbf{Effect of constraining DLO configurations in planning}:
We conduct an ablation study to demonstrate the necessity of constraining DLO configurations in planning. The results are shown in Line 1 and 3 in Table \ref{tab:sim_planner_more_ablation}. In the tests of \textit{Not satisfy stable DLO constraint}, we use full-actuated steering and early-exiting ProjectStableDLOConfig() with a function tolerance (see Ceres document) 100$\times$ larger than that of our standard planner. Thus, the steered DLO configuration is not strictly at a local minimum of energy.
The results show that 
1) the open-loop manipulation success rate significantly drops (from 81\% to 55\%), implying that the quality of planned paths are much worse;
and 2) our closed-loop manipulation framework cannot fully compensate for such errors, as the manipulation success rate is 97\%.

\begin{figure} [tb]
  \centering 
    \includegraphics[width=\linewidth]{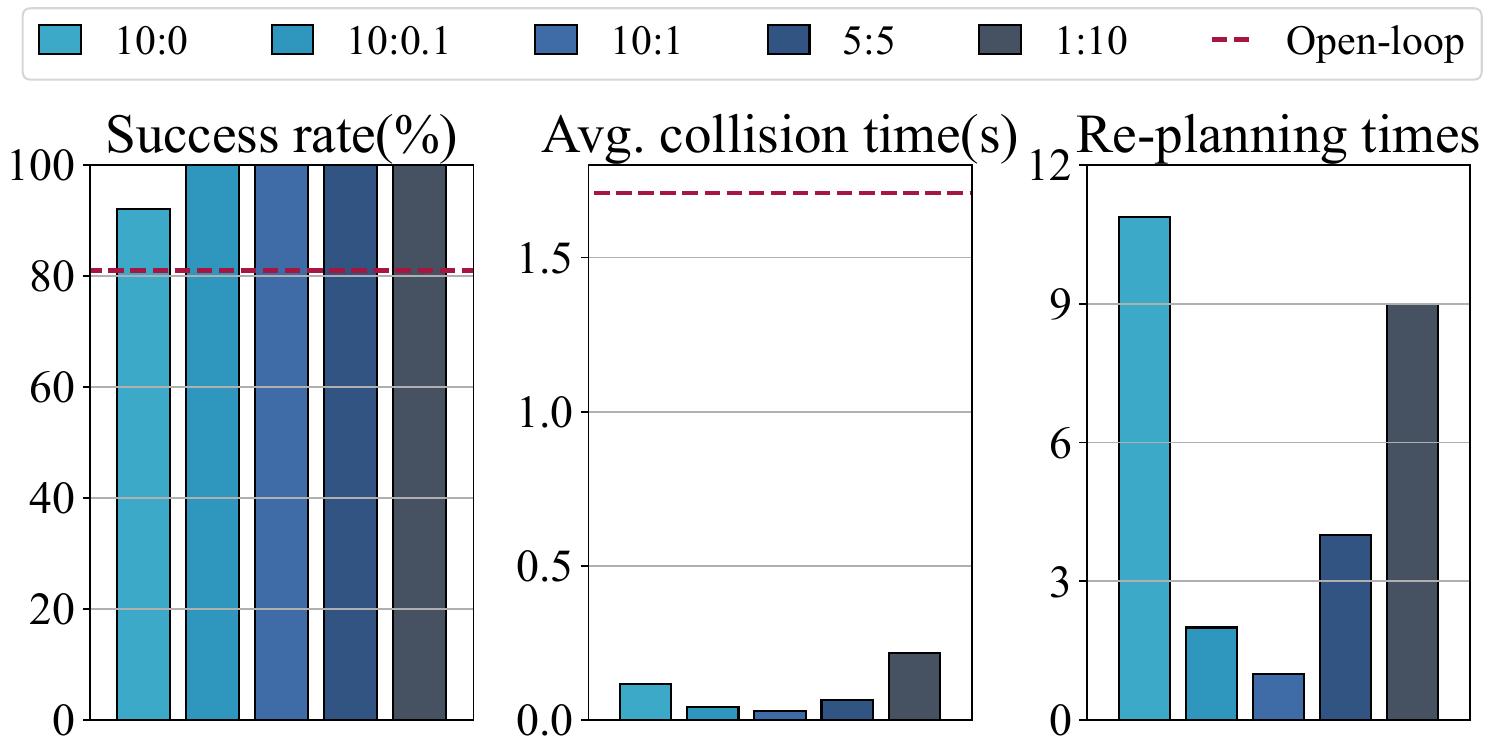} 
  \caption{Effect of the weights of the cost terms for DLO tracking error and robot tracking error in the MPC. The manipulation results in the Simulated Task 4 are presented. The labels are in the format of [$\beta_x : \beta_q$]. The performance of the open-loop manner (red lines) is also shown for reference.
  }
\label{fig:sim_mpc_weight_ablation}
\end{figure}

\textbf{Effect of weights of the cost terms in the MPC}:
Our MPC is designed to track both the planned DLO path and robot path, which is achieved by using the weighted sum of the DLO tracking error and robot tracking error in the cost function. 
Owing to DLO modeling errors, the planned DLO path and robot path cannot be precisely tracked simultaneously, so our MPC aims to balance the two objectives. 
We experimentally study the effect of the weights of these two cost terms ($\beta_x$ for DLO and $\beta_q$ for robot). We test five different combinations of weights: $\beta_x / \beta_q = 10/0, 10/0.1, 10/1, 5/5$ and $1/10$.
Fig. \ref{fig:sim_mpc_weight_ablation} shows the quantitative manipulation results in the simulated Task 4. 
First, when $\beta_x / \beta_q = 10/0$, the planned robot path is discarded and only the planned DLO path is tracked. in this case, the robot easily deviates to unexpected configurations close to obstacles or of low manipulability, where the MPC cannot find local solutions and even the replanning cannot find a new path. 
Second, the MPCs with $\beta_x / \beta_q = 10/0.1, 10/1$ and $5/5$ perform similarly well, as the success rate of each is 100\% and the replanning is invoked less than five times.
Third, when the weight of the robot tracking error is too large, the planned DLO path is almost ignored, so the DLO is more likely to be manipulated to unexpected shapes too close to obstacles where the MPC may get stuck.

\begin{figure} [tb]
  \centering 
    \includegraphics[width=\linewidth]{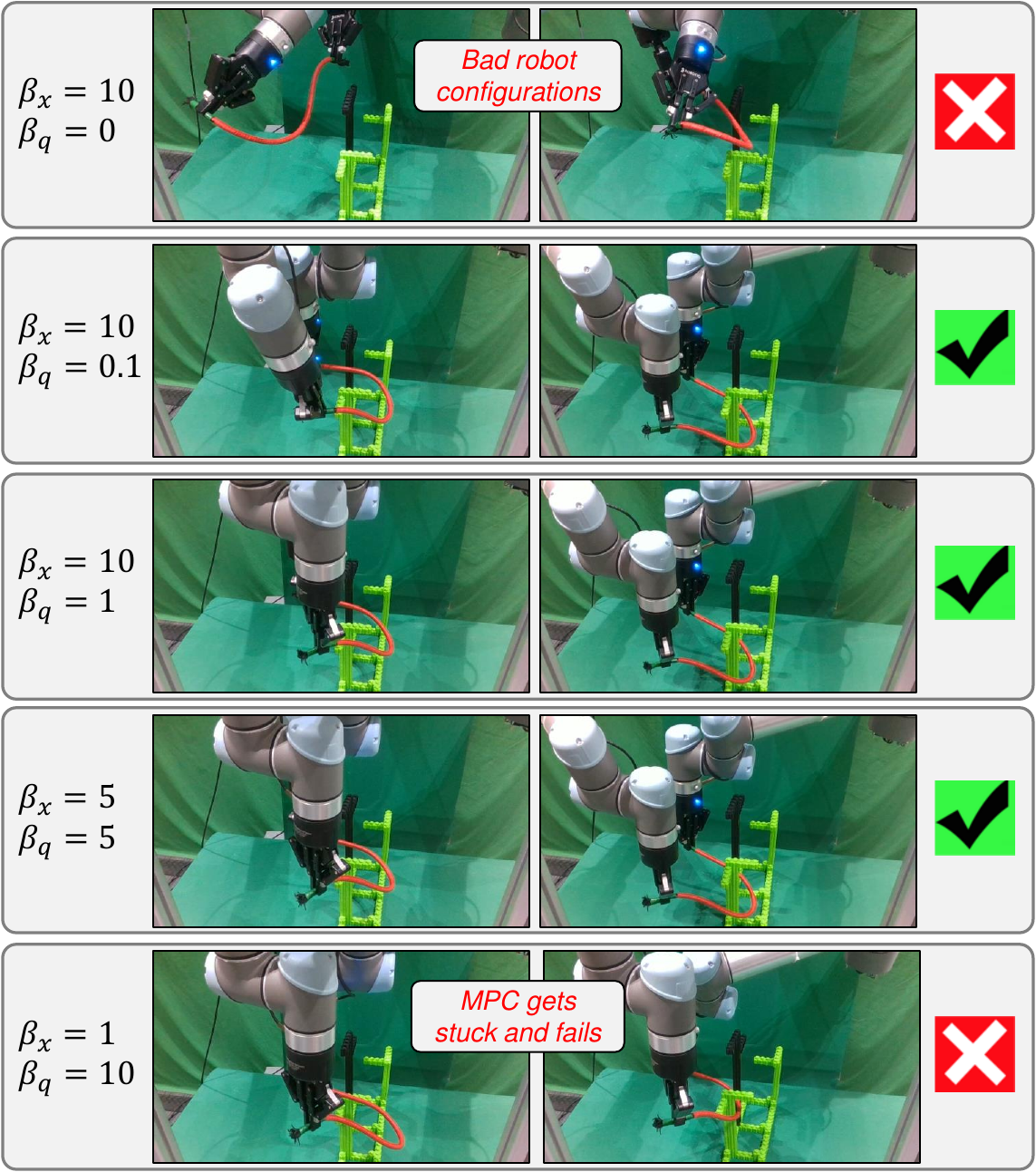} 
  \caption{Real-world experimental tests of the MPCs with different weights of DLO tracking error and robot tracking error.
  }
  \label{fig:real_mpc_weight_ablation}
\end{figure}

We also test the MPCs with these combinations of weights in real-world experiments where the DLO modeling error is more significant, using the same planned path as that shown in Fig. \ref{fig:real_closed_open_compare}. 
The manipulation processes shown in Fig. \ref{fig:real_mpc_weight_ablation} indicate that
1) when $\beta_x/\beta_q = 10/0$, the robot configurations during the execution are far from the planned path and of low manipulability, resulting in self-collision;
2) when $\beta_x/\beta_q = 10/0.1$, $10/1$, and $5/5$, the tests are successful, in which the robot motion adjustments are larger when $\beta_x/\beta_q$ is larger;
and 3) when $\beta_x/\beta_q = 1/10$, owing to the DLO modeling errors, the DLO is moved to an unexpected configuration with obstacles present between it and the planned corresponding waypoint. In this case, the MPC fails to find an acceptable solution within the maximum solving time allowed, and the system finally becomes out of control.
These results demonstrate the necessity of tracking both the DLO path and robot path by the MPC with appropriate weights. We choose $\beta_x=10$ and $\beta_q=1$ for all other tests.

\end{document}